\documentclass[10pt,twocolumn,letterpaper]{article}

\usepackage[pagenumbers]{cvpr} 

\usepackage{multirow}
\usepackage{graphicx}
\usepackage{amsmath}
\usepackage{physics}
\usepackage{caption}
\usepackage{float}
\usepackage{stfloats}
\usepackage{algorithm}       %
\usepackage{algpseudocode}   %
\definecolor{best}{rgb}{0.96, 0.57, 0.58}
\definecolor{second}{rgb}{0.98, 0.78, 0.57}
\definecolor{third}{rgb}{1.0, 1.0, 0.56}

\usepackage{verbatim}
\usepackage{graphicx}
\usepackage{setspace}

\usepackage{enumitem}

\definecolor{cvprblue}{rgb}{0.21,0.49,0.74}
\usepackage[pagebackref,breaklinks,colorlinks,citecolor=cvprblue]{hyperref}

\title{MaGS: Reconstructing and Simulating Dynamic 3D Objects with \\ Mesh-adsorbed Gaussian Splatting}

\author{
\textbf{Shaojie Ma}\textsuperscript{1}
\ 
\textbf{Yawei Luo}\textsuperscript{1$\dagger$}
\ 
\textbf{Wei Yang}\textsuperscript{2}
\ 
\textbf{Yi Yang}\textsuperscript{1} \\
\textsuperscript{1} Zhejiang University
\quad
\textsuperscript{2} Huazhong University of Science and Technology
}

\usepackage{dsfont}
\usepackage{etoolbox}
\usepackage{color}

\newif\ifshowedits

\newcommand{\addeditor}[3]{%
  \definecolor{#1color}{rgb}{#3}
  \expandafter\newcommand\csname #1\endcsname[1]{%
  \ifshowedits
    {\color{#1color} ##1}%
  \else
    {##1}%
  \fi
  }%
  \expandafter\newcommand\csname #1rmk\endcsname[1]{%
  \ifshowedits
    {\color{#1color} {\bf [#2: ##1]}}
  \fi
  }%
  \expandafter\newcommand\csname #1rpl\endcsname[2]{%
  \ifshowedits
    {\color{#1color} ##1 \sout{##2}}
  \else
    {##1}
  \fi
  }%
}

\newcommand{\createtextvar}[1]{
  \expandafter\newcommand\csname #1\endcsname{%
  {\text{#1}}
}%
}

\newcommand{\mycomment}[1]{}

\newcommand{\vcomment}[1]{}

\begin{document}

\twocolumn[{%
\renewcommand\twocolumn[1][]{#1}%
\maketitle
\vspace{-14mm}
\begin{center}
    \captionsetup{type=figure}
\label{fig:teaser_two}
\includegraphics[width=\textwidth,trim=4 100 0 155,clip]{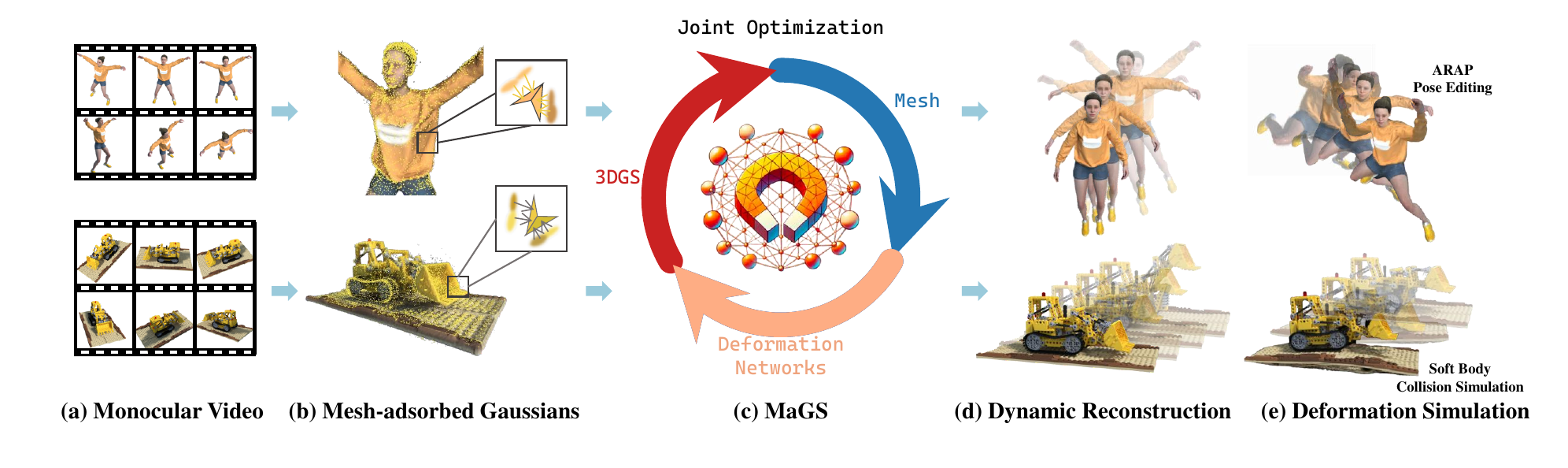}
\end{center}
\vspace{-8mm}
    \captionof{figure}{We propose the Mesh-adsorbed Gaussian Splatting (MaGS), a unified framework for reconstructing and simulating a dynamic 3D object from a monocular video.}
\vspace{5pt}
}]

\if TT\insert\footins{\noindent\footnotesize{
$\dagger$Corresponding Author}\\
Project page: \url{https://wcwac.github.io/MaGS-page/}}\fi

\begin{abstract}

3D reconstruction and simulation, although interrelated, have distinct objectives: reconstruction requires a flexible 3D representation that can adapt to diverse scenes, while simulation needs a structured representation to model motion principles effectively. This paper introduces the \textbf{M}esh-\textbf{a}dsorbed \textbf{G}aussian \textbf{S}platting (MaGS) method to address this challenge. MaGS constrains 3D Gaussians to roam near the mesh, creating a mutually adsorbed mesh-Gaussian 3D representation. Such representation harnesses both the rendering flexibility of 3D Gaussians and the structured property of meshes. To achieve this, we introduce RMD-Net, a network that learns motion priors from video data to refine mesh deformations, alongside RGD-Net, which models the relative displacement between the mesh and Gaussians to enhance rendering fidelity under mesh constraints.  To generalize to novel, user-defined deformations beyond input video without reliance on temporal data, we propose MPE-Net, which leverages inherent mesh information to bootstrap RMD-Net and RGD-Net. Due to the universality of meshes, MaGS is compatible with various deformation priors such as ARAP, SMPL, and soft physics simulation. Extensive experiments on the D-NeRF, DG-Mesh, and PeopleSnapshot datasets demonstrate that MaGS achieves state-of-the-art performance in both reconstruction and simulation.
\end{abstract}

\section{Introduction}
\label{sec:intro}

The human visual system can simultaneously capture 3D appearance (reconstruction) and infer dynamic objects' probable motions (simulation) from a monocular video. In contrast, computer vision and graphics typically treat 3D reconstruction and simulation as distinct tasks. Over the years, various reconstruction methods have emerged~\cite{snavely_photo_2006,furukawa_accurate_2010,mildenhall_nerf_2020,kerbl_3d_2023}.
These methods represent the geometry and appearance of a 3D scene from multi-view images, either implicitly or explicitly, and subsequently render photo-realistic novel views based on the 3D representation. Neural Radiance Fields (NeRF)~\cite{mildenhall_nerf_2020}, an implicit method, utilizes volume rendering techniques to bridge 2D and 3D spaces. %
Explicit methods represent a 3D scene concretely by using rendering primitives, such as 3D points, with PointRF~\cite{zhang_differentiable_2022} and Pulsar~\cite{lassner_pulsar_2021} being representative approaches. More recently, 3D Gaussian Splatting (3DGS) has demonstrated significant improvements in rendering quality and speed~\cite{kerbl_3d_2023,chen_survey_2024}, and has been extended to 4D scene reconstruction%
~\cite{wu_4d_2023,yang_deformable_2023,yang_real-time_2023,miao_pla4d_2024}. These explicit or implicit representations do not impose prior information about the entity on the reconstructed object, ensuring its flexible application in various scenarios.

The simulation utilizes reconstruction results for motion editing~\cite{park_hypernerf_2021,jiang_neuman_2022,yang_neumesh_2022,jung_deformable_2023}, ray tracing rendering~\cite{qiao_dynamic_2023,zeng_mirror-nerf_2023,meng_mirror-3dgs_2024}, and other applications~\cite{xie_physgaussian_2023,feng_gaussian_2024}. Due to the lack of structural or physical priors, such as skeletons, simulating the raw reconstruction results often appear unrealistic~\cite{waczynska_d-miso_2024,borycki_gasp_2024}. To address this, some methods introduce explicit representations to enhance simulation capabilities~\cite{yuan_nerf-editing_2022,yang_neumesh_2022,zhong_reconstruction_2024,jiang_vr-gs_2024}. For instance, Huang \emph{et al.}~\cite{huang_sc-gs_2023} applied control points as handles to guide deformation more structurally. In specific cases like human body, parametric models such as SMPL~\cite{loper_smpl_2015} are commonly used~\cite{yuan_simpoe_2021,moreau_human_2023,li_animatable_2023}. Some approaches aim to integrate spatial continuity priors by introducing explicit mesh representations. 
Guedon \emph{et al.}~\cite{guedon_sugar_2023}, Waczynska \emph{et al.}~\cite{waczynska_games_2024}, and Gao \emph{et al.}~\cite{gao_2024_mani} employed a hybrid mesh and 3DGS representation to enable more natural editing of static objects.
Liu \emph{et al.}~\cite{liu_dynamic_2024} proposed a dynamic mesh Gaussian method to extend hybrid representations for simulating dynamic objects, thus enhancing simulation capabilities. 

3D reconstruction and simulation, while interrelated, have distinct objectives: reconstruction requires a flexible 3D representation adaptable to diverse scenes, whereas simulation necessitates a structured representation to model deformation policies effectively. This dual requirement poses significant challenges for a unified framework. In this paper, we propose the Mesh-adsorbed Gaussian Splatting (MaGS) method to address this. MaGS constrains 3D Gaussians to roam near the mesh surface, creating a hybrid mesh-Gaussian 3D representation that combines the rendering flexibility of 3D Gaussians with the adaptability of meshes to different geometric priors. %
To realize this representation, we introduce a learnable Relative Mesh Deformation Network (RMD-Net) that learns motion principles from video data to refine mesh deformations, and a learnable Relative Gaussian Deformation Network (RGD-Net) to model the relative displacement between the mesh and 3D Gaussians to enhance rendering fidelity under mesh constraints. Unlike conventional anchored, fixed mesh-Gaussian methods~\cite{liu_dynamic_2024,waczynska_games_2024,guedon_sugar_2023}, MaGS allows Gaussians relative displacement through the RGD-Net, effectively bypassing the trade-off between rendering accuracy and deformation rationality during dynamic object reconstruction.

In the simulation, MaGS eliminates dependence on temporal information using Mesh Pose Embedding Network (MPE-Net), which guides relative deformation (i.e., RMD-Net \& RGD-Net) based on mesh inherent information rather than video timestamps. This strategy allows MaGS to generalize effectively to novel deformations beyond the input video. Due to the universality of meshes, MaGS is compatible with mesh-based simulation methods, such as ARAP~\cite{sorkine_as-rigid-as-possible_2007}, SMPL~\cite{loper_smpl_2015}, and soft physics simulation~\cite{terzopoulos_1987_elastically}, enabling it to handle complex deformations. Through joint optimization of meshes, 3D Gaussians, and networks, MaGS achieves both high rendering accuracy and realistic deformation. Extensive experiments on the D-NeRF, DG-Mesh, and PeopleSnapshot datasets demonstrate that MaGS surpasses SOTA methods, establishing a new paradigm for unified reconstruction and simulation tasks.

\section{Related Work}
\label{sec:relatedworks}
\subsection{Neural Rendering for Dynamic Scenes}
Since the introduction of NeRF~\cite{mildenhall_nerf_2020}, advancements in 3D scene reconstruction have progressed rapidly. Researchers have extended NeRF with temporal encoding to address dynamic scenes~\cite{gao_dynamic_2021, pumarola_d-nerf_2020, tretschk_non-rigid_2021}. Other approaches have optimized NeRF's temporal modeling, including using voxel grids for faster training~\cite{fang_fast_2022, liu_devrf_2022} and $k$-plane representations to improve efficiency~\cite{fridovich-keil_k-planes_2023, hu_tri-miprf_2023, cao_hexplane_2023, shao_tensor4d_2022}. 
Park \emph{et al.}~\cite{park_nerfies_2021, park_hypernerf_2021} incorporated geometric priors and hyperspace projections to enhance the interpretability of deformation fields, while Yan \emph{et al.}~\cite{yan_nerf-ds_2023} improved accuracy by modeling specular reflections. More recently, 3DGS~\cite{kerbl_3d_2023} has gained increasing attention. It significantly improves rendering speed compared to NeRF while providing a more explicit geometric interpretation. Dynamic field research has also been applied to 3DGS, studies discussing the use of deformation fields to represent dynamic scenes~\cite{yang_deformable_2023, jung_deformable_2023, wu_4d_2023}. Some studies have applied 3DGS to dynamic scenes of the human body~\cite{li_animatable_2023,qian_3dgs-avatar_2023,shao_splattingavatar_2024}.

\subsection{Neural Rendering Enhanced by Explicit Priors}
Recent studies have explored effective deformation capabilities by introducing explicit priors. NeRF-Editing~\cite{yuan_nerf-editing_2022} integrates NeRF with mesh deformation using ARAP~\cite{sorkine_as-rigid-as-possible_2007}, while NeuMesh~\cite{yang_neumesh_2022} directly incorporates neural fields onto meshes. SuGaR~\cite{guedon_sugar_2023} utilizes Poisson reconstruction to \textbf{\emph{bind}} Gaussian point clouds to mesh and optimizes them simultaneously. GaMeS~\cite{waczynska_games_2024} introduces pseudo-mesh and designs a mesh Gaussian \textbf{\emph{binding}} algorithm for deformation editing. Parametric models are also combined to fit scene deformations, exemplified by Qian \textit{et al.}~\cite{qian_3dgs-avatar_2023, chen_monogaussianavatar_2023, moreau_human_2023} for human body deformation scene modeling. 
SC-GS~\cite{huang_sc-gs_2023} employs sparse control points for Gaussian point cloud deformation. SP-GS~\cite{wan_superpoint_2024} accelerates dynamic scene rendering based on sparse point control. DG-Mesh~\cite{liu_dynamic_2024} enhances 3DGS and mesh integration by mapping Gaussian points to mesh facets with Gaussian-Mesh \textbf{\emph{Anchoring}} for uniformity and improved mesh optimization. D-MiSo~\cite{waczynska_d-miso_2024} extends GaMeS to be applied to dynamic scenes, but its essence is still a point cloud-based representation method.

\subsection{Neural Rendering for Mesh Reconstruction}

Mesh is one of the most widely used representations of 3D objects, with applications in animation, gaming, autonomous driving, and digital twins. Due to its implicit representation, neural rendering is challenging to apply directly in these fields. Extracting high-quality meshes using neural rendering technology has become a popular research topic. Earlier studies primarily used Marching Cubes~\cite{lorensen_marching_1987} to reconstruct meshes from the depth information generated by neural rendering. NeuS~\cite{wang_neus_2023} combines Signed Distance Function with neural rendering to achieve higher quality mesh extraction. NeRF2Mesh ~\cite{tang_delicate_2023} implements an adaptive iterative algorithm, further improving the accuracy of the extracted mesh. SuGaR~\cite{guedon_sugar_2023} and DG-Mesh~\cite{liu_dynamic_2024} bind Gaussian processes to meshes, optimizing mesh details through the Gaussian training process. 2DGS~\cite{huang_2d_2024} designs a flat Gaussian representation to avoid inaccuracies in the Gaussian space boundaries, enhancing the quality of mesh extraction. PGSR~\cite{chen_pgsr_2024} utilizes Planar-based Gaussian Splatting to achieve high-quality mesh extraction. DynaSurfGS~\cite{cai_dynasurfgs_2024} further applies Planar-based Gaussian Splatting in dynamic scenes, allowing for mesh extraction at any moment in dynamic scenarios.

\section{Preliminaries}
\label{subsec:3dgs}
\subsection{3D Gaussian Splatting}
3D Gaussian Splatting (3DGS)~\cite{kerbl_3d_2023} uses learnable 3D Gaussians to map spatial coordinates to pixel values, enhancing rendering quality and efficiency. Each Gaussian has a mean $\mu \in \mathbb{R}^3$, covariance $\Sigma \in \mathbb{R}^{3 \times 3}$, opacity $\sigma$, scaling $s \in \mathbb{R}^3$, rotation $q \in \mathbb{R}^4$, and spherical harmonics $sh \in \mathbb{R}^L$ (where $L$ varies by model).
The 3D Gaussian function is:
\begin{equation}
    G(x) = e^{-\frac{1}{2} (x - \mu)^T \Sigma^{-1} (x - \mu)},
\end{equation}
where $x \in \mathbb{R}^3$. Here, $\Sigma$ is decomposed as:
\begin{equation}
    \Sigma = R S S^T R^T,
\end{equation}
with scaling matrix $S$ and rotation matrix $R$ derived from $s$ and $q$, respectively.
Each Gaussian’s opacity $\sigma$ modulates its effect on the image, while $sh$ coefficients add view-dependent shading.

To render, each 3D Gaussian is projected onto the 2D image. The 3D covariance $\Sigma$ transforms to a 2D covariance $\Sigma^{\prime}$ by:
\begin{equation}
    \Sigma^{\prime} = JW \Sigma W^T J^T,
\end{equation}
where $W$ is the view transformation, and $J$ is the projection Jacobian.

The color $C(u)$ at pixel $u$ is:
\begin{equation}
    C(u) = \sum_{i \in N} T_i \alpha_i c_i,
\end{equation}
with transmittance $T_i$ defined as:
\begin{equation}
    T_i = \prod_{j=1}^{i-1} (1 - \alpha_j),
\end{equation}
and opacity $\alpha_i$ for each Gaussian as:
\begin{equation}
    \alpha_i = \sigma_i e^{-\frac{1}{2} (u - \mu_i)^T \Sigma^{\prime} (u - \mu_i)},
\end{equation}
where $c_i$ is the Gaussian's color and $\mu_i \in \mathbb{R}^2$ is the projected 2D mean.

\subsection{Dynamic 3D Gaussian Splatting}
Modeling dynamic scenes requires handling temporal variations in Gaussian parameters. 
To model dynamic scenes, recent methods like Deformable 3DGS~\cite{yang_deformable_2023} and DynaSurfGS~\cite{cai_dynasurfgs_2024} use 3D Gaussians that adapt over time through deformation fields. A deformation network $\mathbb{D}$ predicts temporal updates to Gaussian parameters from an initial frame ($t=0$) to subsequent frames.
Given a Gaussian $G_g = \{\mu_g, q_g, s_g, \sigma_g, c_g\}$, the deformation network outputs adjustments:
\begin{equation} 
(\delta \mu, \delta q, \delta s) = \mathbb{D}(\mathbb{E}_p(\mu_g), \mathbb{E}_t(t)),
\end{equation}
where $\mathbb{E}_p$ and $\mathbb{E}_t$ are spatial and temporal embeddings. The updated Gaussian at time $t$ is:
\begin{equation}
    \mu_g(t) = \mu_g + \delta \mu, \quad q_g(t) = q_g \cdot \delta q, \quad s_g(t) = s_g + \delta s.
\end{equation}
The deformation network aligns these updates with ground-truth frames to optimize both Gaussian and deformation parameters.

\subsection{Mesh Extraction from Dynamic 3D Gaussians}
\label{subsec:3.3}
Prior approaches to mesh extraction from Gaussian representations~\cite{cai_dynasurfgs_2024, chen_pgsr_2024,zhang_dynamic_2024} frequently leverage a truncated signed distance function (TSDF)~\cite{gao_multi-target_2019} to integrate multiple RGB-D views from diverse perspectives into a unified 3D mesh. To achieve simplification, the quadric decimation~\cite{garland_1997_surface} is typically applied. However, these approaches are limited in that they do not produce meshes with temporal facet-ID correspondence, making them unsuitable for MaGS (as such correspondences are necessary to transfer learned deformation principles from the rendering process to simulation). Instead, we adopt a single mesh generated using TSDF at a given time and employ the deformation network to adjust mesh vertices, yielding a set of coarse meshes that maintain consistent correspondence across frames.

\section{Methodology}
\begin{figure*}[!ht]
    \centering
    \includegraphics[width=0.9\linewidth]{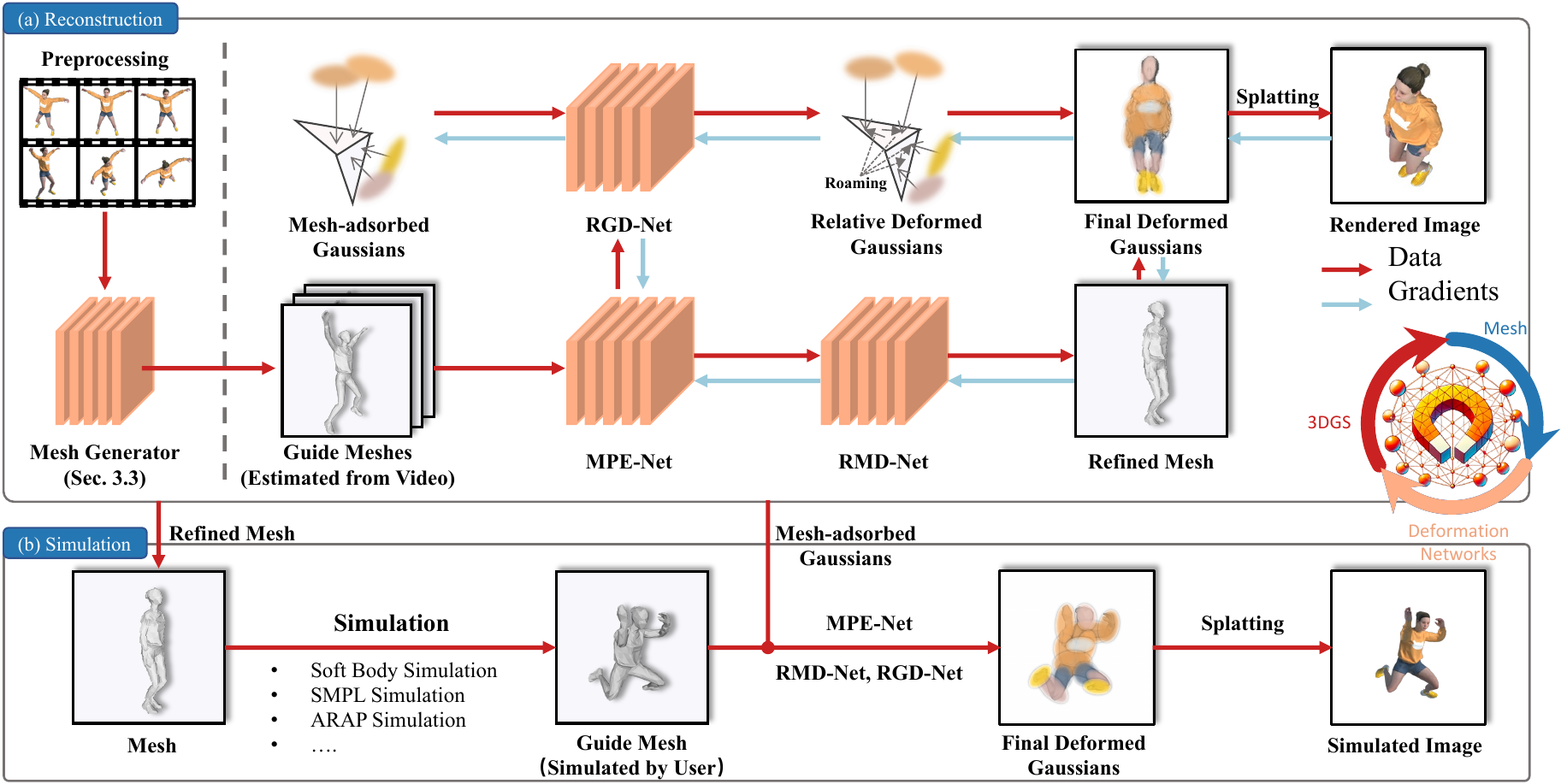}
    \caption{
    \textbf{Pipeline of MaGS.} MaGS begins by extracting a temporally consistent coarse mesh for each frame of video. These meshes, referred to as Guide Meshes, provide the foundation for dynamic reconstruction. During the \textbf{reconstruction process}, pose information from the guide meshes is extracted using MPE-Net and forwarded to RMD-Net and RGD-Net. RMD-Net and RGD-Net perform relative deformations on the guide mesh and the Mesh-adsorbed Gaussians, respectively, yielding the refined mesh and relative deformed Gaussian. These two components produce the Final Deformed Gaussians. Splatting-based rendering is then employed, with the rendering loss used to optimize the Gaussians, MPE-Net, RMD-Net, and RGD-Net via backpropagation. The reconstruction phase not only yields a high-precision mesh and Gaussians but also trains the networks to learn deformation principles from the video, effectively preparing them for simulation. In the \textbf{simulation phase}, mesh-based techniques—such as soft body simulation, ARAP, and SMPL—are used to deform the reconstructed meshes, producing new guide mesh. Mesh-adsorbed Gaussians are also inherited (adsorbed to their corresponding facets). The following process resembles the reconstruction, where MPE-Net, RMD-Net, and RGD-Net are again utilized to yield the Final Deformed Gaussians, which are then rendered to generate the final image.
    }
    \label{fig:pipeline}
    \vspace{-10pt}
\end{figure*}

Figure~\ref{fig:pipeline} gives an overview of MaGS. Section~\ref{subsec:4.1} details the Mesh-adsorbed Gaussian initialization. Section~\ref{subsec:4.2} describes the network design. Section~\ref{subsec:4.3} outlines the optimization process for Mesh-adsorbed Gaussian, and Section~\ref{subsec:4.4} explains the simulation process. In Section~\ref{subsec:4.5}, we discuss MaGS with existing methods to illustrate the distinctions of our design. Implementation details of each network can be found in the Appendix.

\subsection{Mesh-adsorbed Gaussian}
\label{subsec:4.1}

The Mesh-adsorbed Gaussian is a hybrid representation composed of a triangular mesh and 3D Gaussian. The mesh part is a triangular mesh with $ N $ vertices and $ M $ facets, defined by three attributes: $ vertices $, $ facets $, and $ normals $. Here, $ vertices $ represents the spatial coordinates of each vertex, denoted as $ v_i \in \mathbb{R}^3 $ for $ i = 1, 2, \dots, N $. The $ facets $ attribute defines each facet by referencing three vertices of the mesh. The $ normals $ attribute provides the normal vector $ n_i \in \mathbb{R}^3 $ for each facet $ i = 1, 2, \dots, M $, which can be calculated from the $ vertices $ and $ facets $. Following Shao \emph{et al.}~\cite{shao_splattingavatar_2024}, we compute the normal at any position on a facet by linearly interpolating the normals at its vertices.

The 3D Gaussian of Mesh-adsorbed Gaussians extends the standard 3D Gaussian with attributes $ w $ and $ \text{MeshId} $. Here, $ w \in [0,1]^3 $ represents logical coordinates on a mesh facet, and $ \text{MeshId} $ records the facet on which the Gaussian resides. Initially, Gaussians are randomly initialized, with $ w $ values generated randomly, and $ \text{MeshId} $ evenly distributed across facets, ensuring an equal number of Gaussians per facet.

To compute the spatial coordinates of a 3D Gaussian based on its logical coordinates, we first define the interpolation function $f$:
\begin{equation}
f(w, x, y, z) = w_1 x + w_2 y + (1 - w_1 - w_2) z,
\end{equation}
where $w$ represents the coordinates and $x, y ,z$ denotes the values at the three vertices.
For a Gaussian point $ i $ located on facet $ j $, the $ \mu_i $ of the Gaussian can be computed as follows:
\begin{equation}
l_j = \frac{|\vec{v_{j_1}v_{j_2}}| \, |\vec{v_{j_1}v_{j_3}}|}{|\vec{v_{j_1}v_{j_2}} \times \vec{v_{j_1}v_{j_3}}|}, 
\mu_i = f(w_i, v_{j_1}, v_{j_2}, v_{j_3}) + w_{i_3} n_j l_j,
\end{equation}
where $ l_j $ represents the scaling factor for facet $ j $, $ v_{j_1}, v_{j_2}, v_{j_3} $ are the spatial coordinates of the three vertices of facet $ j $, and $ n_j $ denotes the normal vector of facet $ j $.

When the component $w_{i_3}$ of the Gaussian point $i$ is nonzero, the Gaussian point does not lie directly on the mesh surface but instead hovers above it. This increases the degree of freedom for the Gaussian to move across the mesh, and we refer to this design as \textbf{Gaussian Hover}.

\subsection{Deformation for Mesh-adsorbed Gaussian}
\label{subsec:4.2}

We design three neural networks to handle deformation: MPE-Net, RMD-Net, and RGD-Net. MPE-Net extracts pose information from coarse mesh, which is input to RMD-Net and RGD-Net. RMD-Net predicts the relative mesh deformation to refine the dynamic mesh, while RGD-Net adjusts the corresponding Gaussian parameters.

\subsubsection{Mesh Pose Embedding Network}

Since timestamps are not available as input during simulation, relying on temporal cues to guide deformation would hinder simulation performance. Therefore, we abandon timestamps as input for our algorithm. Instead, we design MPE-Net to extract information directly from the guide (coarse) mesh. MPE-Net takes as input a subset of mesh vertices called handle vertices:
\begin{equation}
\begin{aligned}
\mathcal{E}_\text{M}, \mathcal{E}_\text{V} = \text{MPE}(\mathcal{H}, \mathcal{N}),
\end{aligned}
\end{equation}
where $\mathcal{H}$ and $\mathcal{N}$ denote the sets of handle vertices and their corresponding normal vectors, respectively, and $ \mathcal{E}_\text{M} $ and $ \mathcal{E}_\text{V} $ encode the mesh pose and vertex-specific deformations, respectively.

To ensure that handle vertices capture more pose information of the mesh, we use Poisson disk sampling~\cite{yuksel_sample_2015} to sample points uniformly throughout the mesh and map them to mesh vertices using the k-nearest neighbors (KNN). 

\subsubsection{Relative Mesh Deformation Network}
We design RMD-Net to learn motion priors from video within a rendering process, enabling it to refine guide mesh deformations by predicting the relative displacement between the guide and refined meshes. RMD-Net takes $\mathcal{E}_\text{M}$ and $\mathcal{E}_\text{V}$ as inputs to predict refined deformation information for each vertex $i$. This process can be represented as:
\begin{equation}
\begin{aligned}
(\Delta v, \Delta q, \Delta s, \Delta \sigma, \Delta c)_i = \text{RMD}(\mathcal{E}_\text{M}, \mathcal{E}_{\text{V}_i}),
\end{aligned}
\end{equation}
where $\Delta v$ represents the displacement of the vertex and $\Delta q$ represents the rotation applied to the vertex coordinates. Although $\Delta s$, $\Delta \sigma$, and $\Delta c$ do not correspond directly to mesh deformation, they are utilized in Mesh-adsorbed Gaussian rendering.

\subsubsection{Relative Gaussian Deformation Network}

Unlike other methods, MaGS allows Gaussian to \textbf{roam on} rather than \textbf{anchor to} a facet when the mesh deforms. Specifically, we consider that the logical coordinates $w$ of each Gaussian on the mesh change with the mesh deformations. For this purpose, we design RGD-Net to calculate changes of $w$, denoted as $\Delta w$, during mesh deformation.

RGD-Net takes $\mathcal{E}_\text{M}$, the $\mathcal{E}_{\text{V}_{1}},\mathcal{E}_{\text{V}_{2}},\mathcal{E}_{\text{V}_{3}}$ of the three vertices of the facet, and $w_i$ of Mesh-adsorbed Gaussian $i$ as input and outputs $\Delta w_i$.
\begin{equation}
\begin{aligned}
\Delta w_i = \text{RGD}(\mathcal{E}_\text{M}, \mathcal{E}_{\text{V}_{1}},\mathcal{E}_{\text{V}_{2}},\mathcal{E}_{\text{V}_{3}}, w_i).
\end{aligned}
\end{equation}

\subsection{Optimization for Mesh-adsorbed Gaussian}
\label{subsec:4.3}
This section explains the optimization process, detailing how the mesh vertices and Mesh-adsorbed Gaussian parameters are updated through backpropagation.

When the mesh deforms, the facet area and surface normal change, affecting the Final Deformed Gaussian. For facet $ j $, we calculate $ \Delta l_j = \frac{l_j'}{l_j} $, where $ l_j $ and $ l_j' $ are the scaling factors before and after deformation, respectively. Similarly, the change in the surface normal, $ \Delta n_j $, is computed. Using the methods in Section~\ref{subsec:4.2}, we obtain $ \Delta v_j, \Delta q_j, \Delta s_j, \Delta \sigma_j, \Delta c_j $ for facet $ j $, and $ \Delta w_i $ for Gaussian $ i $. The mean position $ \mu_i $ of the Final Deformed Gaussian is then computed using the following equations:
\begin{equation}
\begin{aligned}
w'_i &= w_i + \Delta w_i, \;
v'_j = v_j + \Delta v_j, \\
\mu'_i &= f(w'_i, v'_{j_1}, v'_{j_2}, v'_{j_3}) + w'_{i_3}  n'_j  l'_j,
\end{aligned}
\end{equation}
where $n'_j$ represents the normal after deformation.
Other properties $\mu'_i, \sigma'_i, c'_i, s'_i$ and $ q'_i $ a are updated as follows:
\begin{equation}
\begin{aligned}
&\sigma'_i = f(w'_i, \Delta \sigma_{j_1}, \Delta \sigma_{j_2}, \Delta \sigma_{j_3}) \cdot \sigma_i, \\
&c'_i =f(w'_i, \Delta c_{j_1}, \Delta c_{j_2}, \Delta c_{j_3})  + c_i, \\
 &s'_i = \Delta l_j \cdot f(w'_i, \Delta s_{j_1}, \Delta s_{j_2}, \Delta s_{j_3}) \cdot s_i,\\ 
&q'_i = \Delta n_j \cdot f(w'_i, \Delta q_{j_1}, \Delta q_{j_2}, \Delta q_{j_3}) \cdot q_i.
\end{aligned}
\end{equation}
where $s_i, q_i, \sigma_i,$ and $c_i$ are the undeformed properties.

Through the above steps, we derive the Final Deformed Gaussians based on Mesh-adsorbed Gaussians by utilizing the deformation information provided by the mesh. Formally, this can be expressed as:
\begin{equation}
(w_{\text{gaussians}} + \Delta w) \times (v_{\text{mesh}} + \Delta v) \rightarrow \textbf{Final Deformed Gaussians}, 
\end{equation}
where $\times$ denotes the interpolation operation MaGS uses to calculate Gaussian properties. 

This formulation captures the adjustment of Mesh-adsorbed Gaussians properties in response to mesh deformations, ensuring that the Gaussians' positions and orientations are accurately updated as the underlying mesh structure deforms. The Final Deformed Gaussians incorporate all parameters of a standard 3D Gaussians and can be splatted using a differentiable renderer~\cite{kerbl_3d_2023}. We render the Final Deformed Gaussians, then compare the rendered image with the ground truth. The loss is computed using the following equation, enabling backpropagation:
\begin{equation}
\begin{aligned}
\mathcal{L} = \mathcal{L}_{\text{L1}} \times (1 - \mathcal{\lambda}_{\text{ssim}}) + \mathcal{L}_{\text{SSIM}} \times \mathcal{\lambda}_{\text{ssim}},
\end{aligned}
\end{equation}
where $\mathcal{L}_{\text{L1}}$ represents the pixel-wise L1 difference, and $\mathcal{L}_{\text{SSIM}}$ represents the structural similarity loss. Since the entire process is differentiable, we can jointly optimize MPE-Net, RMD-Net, RGD-Net, and the Mesh-adsorbed Gaussians through backpropagation based on rendering errors.

\subsection{Mesh-guided Simulation}
\label{subsec:4.4}

\begin{table*}[t]
\centering
\caption{\textbf{Quantitative results on D-NeRF dataset~\cite{pumarola_d-nerf_2020}}. We present the average PSNR/MS-SSIM/VGG-LPIPS values for novel view synthesis on dynamic scenes from D-NeRF, with each cell colored to indicate the \colorbox{best}{best}, \colorbox{second}{second best} and \colorbox{third}{third best}.}
\label{tab:dnerf_dataset}
\resizebox{\textwidth}{!}{%
\begin{tabular}{@{}llcccccccccccc@{}}
\toprule
 &  & \multicolumn{3}{c}{Bouncingballs} & \multicolumn{3}{c}{Hellwarrior} & \multicolumn{3}{c}{Hook} & \multicolumn{3}{c}{Jumpingjacks} \\
\multirow{-2}{*}{Type} & \multirow{-2}{*}{Method} & PSNR↑ & MS-SSIM↑ & LPIPS↓ & PSNR↑ & MS-SSIM↑ & LPIPS↓ & PSNR↑ & MS-SSIM↑ & LPIPS↓ & PSNR↑ & MS-SSIM↑ & LPIPS↓ \\ \midrule
 & D-NeRF~\cite{pumarola_d-nerf_2020} & 38.18 & 0.9910 & 0.0120 & 28.47 & 0.9317 & 0.0638 & 30.42 & 0.9820 & 0.0379 & 33.73 & 0.9902 & 0.0206 \\
 & TiNeuVox-B~\cite{fang_fast_2022} & 40.62 & 0.9969 & 0.0083 & 30.68 & 0.9495 & 0.0592 & 32.45 & 0.9898 & 0.0374 & 35.50 & 0.9944 & 0.0191 \\
 & Tensor4D~\cite{shao_tensor4d_2022} & 25.36 & 0.9610 & 0.0411 & 31.40 & 0.9250 & 0.0675 & 29.03 & 0.9550 & 0.4990 & 24.01 & 0.9190 & 0.0768 \\
\multirow{-4}{*}{NeRF-Based} & K-Planes~\cite{fridovich-keil_k-planes_2023} & 40.61 & 0.9910 & 0.2970 & 25.27 & 0.9480 & 0.0755 & 28.59 & 0.9530 & 0.5810 & 32.27 & 0.9710 & 0.0389 \\ \cmidrule(lr){2-2}
 & Deformable-GS~\cite{yang_deformable_2023} & 37.09 & 0.9974 & \cellcolor[HTML]{FFFF8F}0.0060 & 41.17 & 0.9934 & \cellcolor[HTML]{FFFF8F}0.0152 & 36.48 & 0.9962 & 0.0117 & 37.99 & 0.9978 & 0.0065 \\
 & 4D-GS~\cite{wu_4d_2023} & 40.67 & 0.9968 & 0.0069 & 31.84 & 0.9629 & 0.0474 & 33.90 & 0.9922 & 0.0201 & 36.66 & 0.9962 & 0.0106 \\
 & SC-GS~\cite{huang_sc-gs_2023} & \cellcolor[HTML]{F59194}44.91 & \cellcolor[HTML]{FAC791}0.9980 & 0.0166 & \cellcolor[HTML]{FFFF8F}42.93 & \cellcolor[HTML]{FFFF8F}0.9940 & 0.0155 & \cellcolor[HTML]{FAC791}39.89 & \cellcolor[HTML]{FFFF8F}0.9970 & \cellcolor[HTML]{FFFF8F}0.0076 & \cellcolor[HTML]{FFFF8F}41.13 & 0.9980 & 0.0067 \\
 & SP-GS~\cite{wan_superpoint_2024} & 41.72 & 0.9970 & 0.0097 & 40.25 & 0.9904 & 0.0289 & 35.42 & 0.9928 & 0.0202 & 34.70 & 0.9926 & 0.0181 \\
\multirow{-5}{*}{3DGS Based} & Grid4D~\cite{xu_grid4d_2024} & \cellcolor[HTML]{FFFF8F}41.92 & \cellcolor[HTML]{F59194}0.9981 & \cellcolor[HTML]{F59194}0.0053 & \cellcolor[HTML]{FAC791}43.43 & \cellcolor[HTML]{F59194}0.9959 & \cellcolor[HTML]{FAC791}0.0099 & \cellcolor[HTML]{FFFF8F}39.14 & \cellcolor[HTML]{FAC791}0.9974 & \cellcolor[HTML]{FAC791}0.0067 & 39.92 & \cellcolor[HTML]{FFFF8F}0.9984 & \cellcolor[HTML]{FFFF8F}0.0050 \\ \cmidrule(lr){2-2}
 & D-MiSo~\cite{waczynska_d-miso_2024} & 38.80 & 0.9957 & 0.0140 & 40.69 & 0.9919 & 0.0233 & 37.53 & 0.9961 & 0.0116 & \cellcolor[HTML]{FAC791}41.86 & \cellcolor[HTML]{FAC791}0.9988 & \cellcolor[HTML]{FAC791}0.0042 \\
 & DG-Mesh~\cite{liu_dynamic_2024} & 31.66 & 0.9762 & 0.0351 & 27.80 & 0.9705 & 0.0546 & 29.34 & 0.9613 & 0.0516 & 27.54 & 0.9702 & 0.1149 \\
 & DynaSurfGS~\cite{cai_dynasurfgs_2024} & 40.92 & 0.9948 & 0.0139 & 29.45 & 0.9758 & 0.0360 & 32.97 & 0.9773 & 0.0277 & 35.49 & 0.9864 & 0.0202 \\
\multirow{-4}{*}{Mesh-3DGS Based} & Ours & \cellcolor[HTML]{FAC791}41.97 & \cellcolor[HTML]{FFFF8F}0.9976 & \cellcolor[HTML]{FAC791}0.0055 & \cellcolor[HTML]{F59194}43.69 & \cellcolor[HTML]{FAC791}0.9957 & \cellcolor[HTML]{F59194}0.0098 & \cellcolor[HTML]{F59194}41.23 & \cellcolor[HTML]{F59194}0.9984 & \cellcolor[HTML]{F59194}0.0049 & \cellcolor[HTML]{F59194}44.29 & \cellcolor[HTML]{F59194}0.9993 & \cellcolor[HTML]{F59194}0.0022 \\ \midrule
 &  & \multicolumn{3}{c}{Mutant} & \multicolumn{3}{c}{Standup} & \multicolumn{3}{c}{Trex} & \multicolumn{3}{c}{Average} \\
\multirow{-2}{*}{Type} & \multirow{-2}{*}{Method} & PSNR↑ & MS-SSIM↑ & LPIPS↓ & PSNR↑ & MS-SSIM↑ & LPIPS↓ & PSNR↑ & MS-SSIM↑ & LPIPS↓ & PSNR↑ & MS-SSIM↑ & LPIPS↓ \\ \midrule
 & D-NeRF & 32.31 & 0.9871 & 0.0256 & 34.42 & 0.9896 & 0.0197 & 32.07 & 0.9910 & 0.0178 & 32.80 & 0.9804 & 0.0282 \\
 & TiNeuVox-B & 33.75 & 0.9920 & 0.0288 & 35.95 & 0.9930 & 0.0187 & 33.18 & 0.9955 & 0.0174 & 34.59 & 0.9873 & 0.0270 \\
 & Tensor4D & 29.99 & 0.9510 & 0.0422 & 30.86 & 0.9640 & 0.0214 & 23.51 & 0.9340 & 0.0640 & 27.74 & 0.9441 & 0.1160 \\
\multirow{-4}{*}{NeRF-Based} & K-Planes & 33.79 & 0.9820 & 0.0207 & 34.31 & 0.9840 & 0.0194 & 31.41 & 0.9800 & 0.0234 & 32.32 & 0.9727 & 0.1508 \\ \cmidrule(lr){2-2}
 & Deformable-GS & 41.02 & 0.9990 & 0.0038 & 42.01 & 0.9988 & 0.0036 & 36.07 & 0.9978 & 0.0056 & 38.83 & 0.9972 & \cellcolor[HTML]{FFFF8F}0.0075 \\
 & 4D-GS & 37.16 & 0.9963 & 0.0106 & 37.79 & 0.9959 & 0.0102 & 35.00 & 0.9968 & 0.0087 & 36.15 & 0.9910 & 0.0163 \\
 & SC-GS & \cellcolor[HTML]{FFFF8F}45.19 & 0.9990 & 0.0028 & \cellcolor[HTML]{FAC791}47.89 & 0.9990 & 0.0023 & \cellcolor[HTML]{FAC791}41.24 & 0.9980 & 0.0046 & \cellcolor[HTML]{FAC791}43.31 & \cellcolor[HTML]{FFFF8F}0.9976 & 0.0080 \\
 & SP-GS & 38.69 & 0.9970 & 0.0118 & 42.22 & 0.9977 & 0.0095 & 32.93 & 0.9930 & 0.0163 & 37.99 & 0.9944 & 0.0164 \\
\multirow{-5}{*}{3DGS Based} & Grid4D & \cellcolor[HTML]{FAC791}45.33 & \cellcolor[HTML]{FAC791}0.9995 & \cellcolor[HTML]{FAC791}0.0020 & 47.35 & \cellcolor[HTML]{FAC791}0.9996 & \cellcolor[HTML]{FAC791}0.0015 & \cellcolor[HTML]{FFFF8F}40.82 & \cellcolor[HTML]{FAC791}0.9992 & \cellcolor[HTML]{FAC791}0.0027 & \cellcolor[HTML]{FFFF8F}42.56 & \cellcolor[HTML]{FAC791}0.9983 & \cellcolor[HTML]{FAC791}0.0047 \\ \cmidrule(lr){2-2}
 & D-MiSo & 44.45 & \cellcolor[HTML]{FFFF8F}0.9992 & \cellcolor[HTML]{FFFF8F}0.0027 & \cellcolor[HTML]{FFFF8F}47.46 & \cellcolor[HTML]{FFFF8F}0.9995 & \cellcolor[HTML]{FFFF8F}0.0017 & 40.52 & \cellcolor[HTML]{FFFF8F}0.9990 & \cellcolor[HTML]{FFFF8F}0.0031 & 41.62 & 0.9972 & 0.0087 \\
 & DG-Mesh & 31.44 & 0.9693 & 0.0378 & 32.31 & 0.9791 & 0.0355 & 29.10 & 0.9674 & 0.0507 & 29.88 & 0.9706 & 0.0543 \\
 & DynaSurfGS & 38.61 & 0.9903 & 0.0150 & 37.76 & 0.9884 & 0.0191 & 34.21 & 0.9848 & 0.0230 & 35.63 & 0.9854 & 0.0221 \\
\multirow{-4}{*}{Mesh-3DGS Based} & Ours & \cellcolor[HTML]{F59194}46.42 & \cellcolor[HTML]{F59194}0.9996 & \cellcolor[HTML]{F59194}0.0019 & \cellcolor[HTML]{F59194}49.16 & \cellcolor[HTML]{F59194}0.9997 & \cellcolor[HTML]{F59194}0.0010 & \cellcolor[HTML]{F59194}41.65 & \cellcolor[HTML]{F59194}0.9993 & \cellcolor[HTML]{F59194}0.0025 & \cellcolor[HTML]{F59194}44.06 & \cellcolor[HTML]{F59194}0.9985 & \cellcolor[HTML]{F59194}0.0040 \\ \midrule
\end{tabular}%
}
\vspace{-10pt}
\end{table*}

Finally, we discuss how the optimized Mesh-adsorbed Gaussians can be leveraged for simulation tasks, taking advantage of the mesh's universality to support various physics-based priors. Generally, we utilize physics simulation tools like Blender or Taichi for soft-body simulation, As-Rigid-As-Possible (ARAP) editing simulation, cloth simulation, etc, to enhance the reconstructed mesh with physical prior. Moreover, the RMD-Net introduced in Section~\ref{subsec:4.2} has learned to refine guide mesh with deformation principles within training videos, which can further boost the reasonability of deformation. Since our pipeline retains the original vertex-facet relationships, we can apply the offsets $\Delta v$ calculated by RMD-Net directly to the vertices $v$ of the guide mesh to obtain the optimized deformed mesh. 

\begin{figure}[t]
  \centering
  \begin{subfigure}{0.48\linewidth}
  \centering
    \includegraphics[width=0.49\linewidth,trim=200 0 200 0,clip]{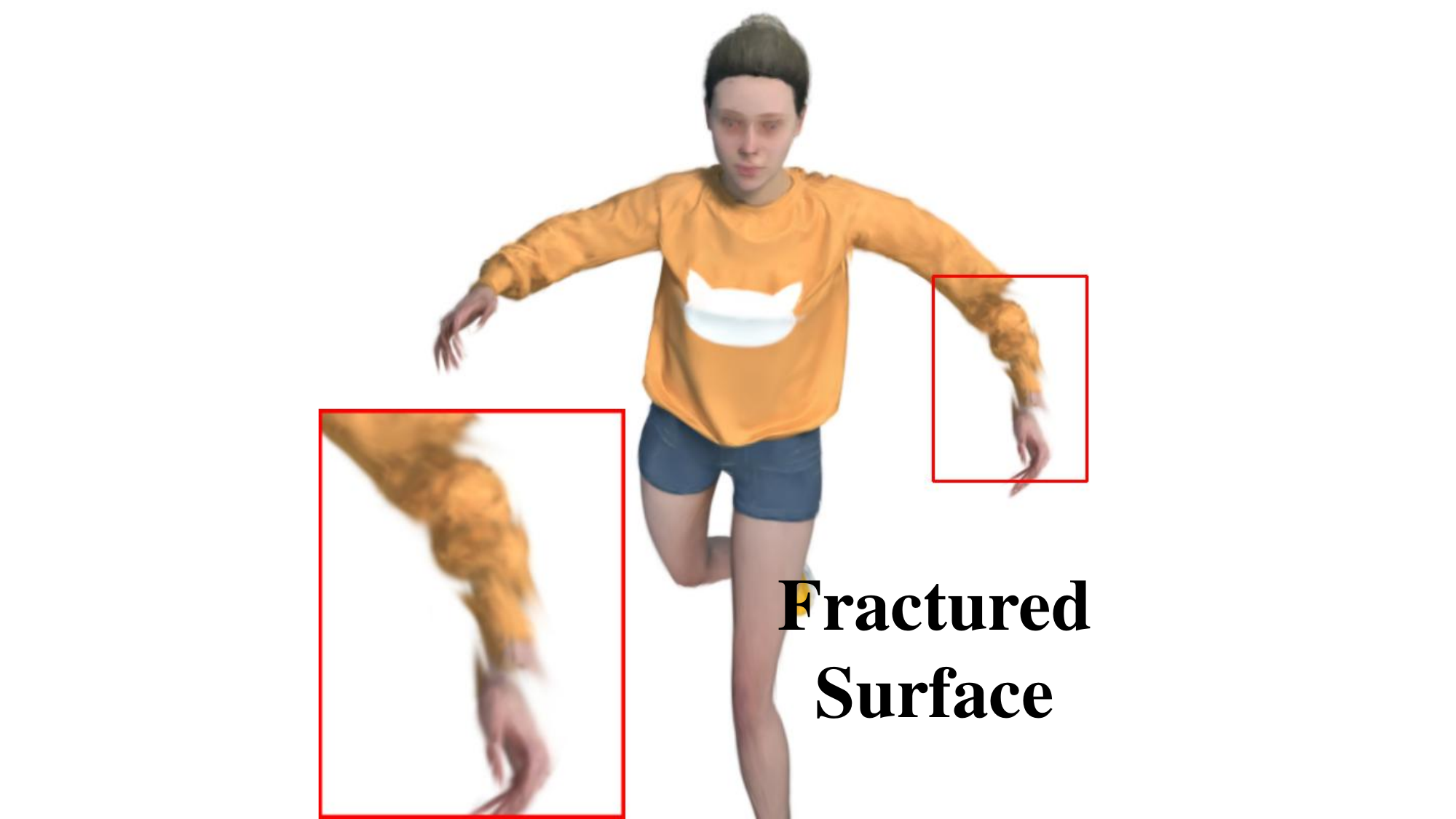}
    \includegraphics[width=0.48\linewidth,trim=200 0 200 0,clip]{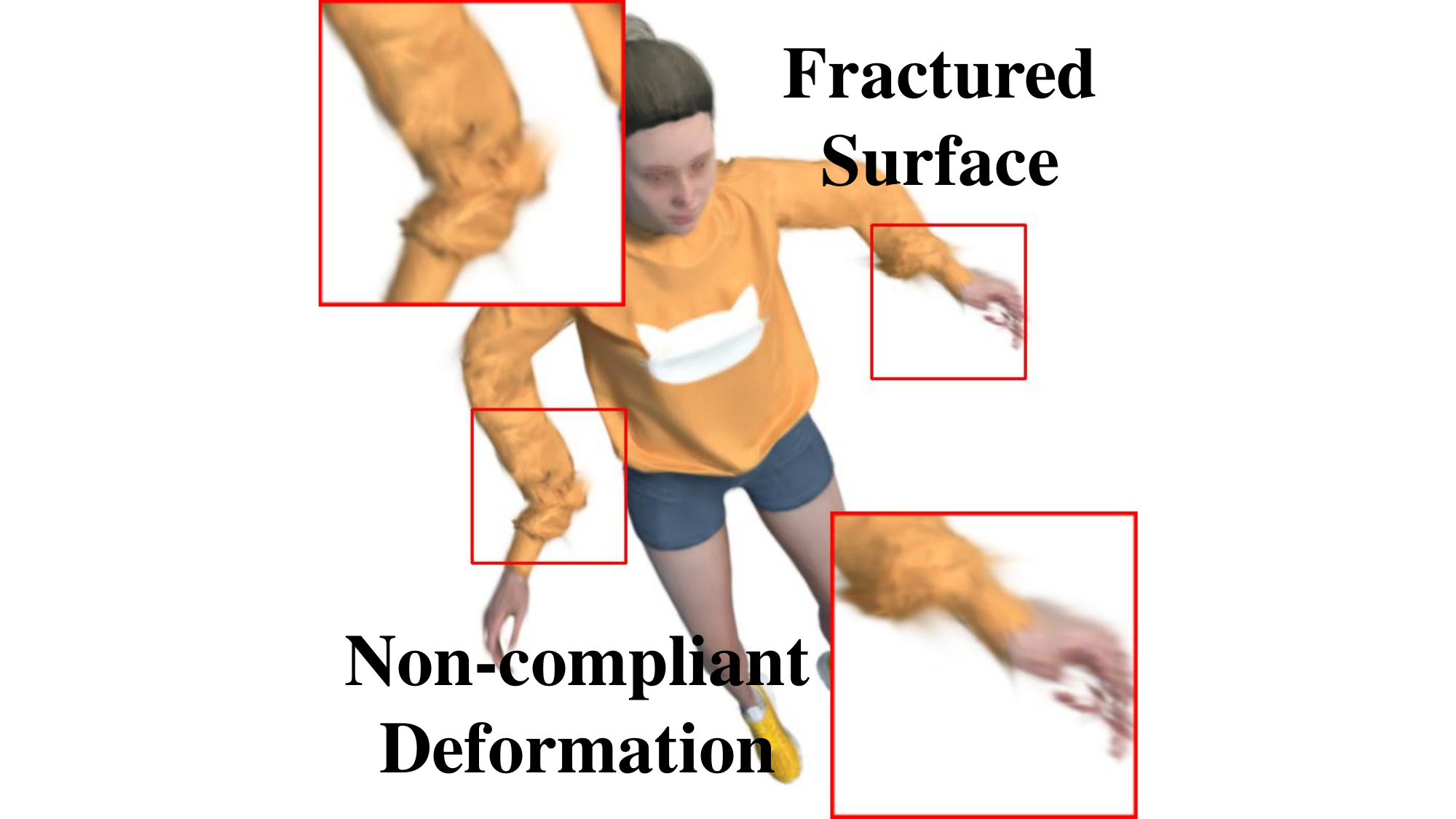}
    \caption{SC-GS}
  \end{subfigure}
  \hfill
  \begin{subfigure}{0.48\linewidth}
  \centering
    \includegraphics[width=0.48\linewidth]{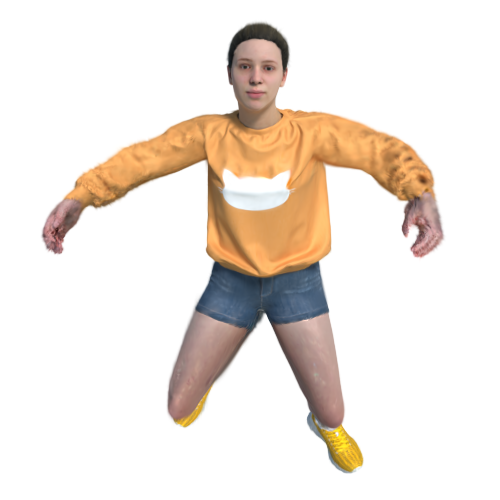}
    \includegraphics[width=0.48\linewidth]{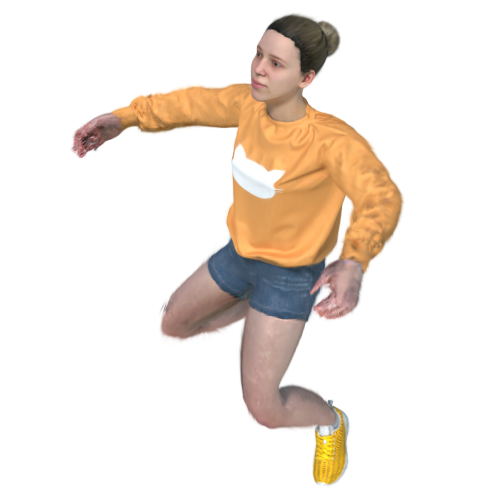}
    \caption{MaGS}
  \end{subfigure}
  \caption{\textbf{Simulation comparison on the D-NeRF dataset~\cite{pumarola_d-nerf_2020}.}}
  \label{fig:compare_simu}
\end{figure}

After deforming the mesh, we adjust the Mesh-adsorbed Gaussians accordingly and render the simulation results with the method described in Section~\ref{subsec:4.3}. This approach enables high-precision rendering of simulation outcomes based on Mesh-adsorbed Gaussians.
\subsection{Discussion}
\label{subsec:4.5}

\begin{figure}
  \centering
  \begin{subfigure}{0.48\linewidth}
        \includegraphics[width=1\linewidth,trim=0 60 290 510,clip]{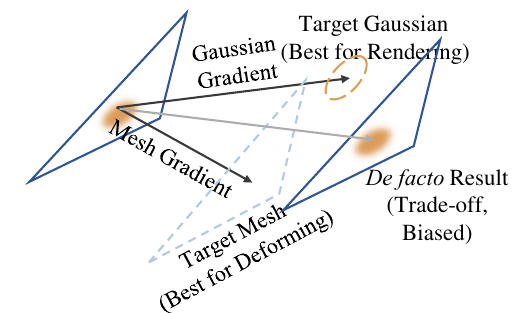}
        \caption{\textbf{Mesh-anchored GS}}
    \end{subfigure}
    \hfill
   \begin{subfigure}{0.50\linewidth}
        \includegraphics[width=1\linewidth,trim=270 70 0 510,clip]{figures/biased_image.pdf}
        \caption{\textbf{Mesh-adsorbed GS (Ours)}}
    \end{subfigure}
    \caption{
    Existing methods integrate 3D Gaussians with mesh by \textbf{anchoring} each Gaussian rigidly to a mesh facet, limiting \textbf{relative displacement} during mesh deformation. This fixed setup causes misalignment between Gaussians and the mesh during joint optimization, compromising rendering accuracy and deformation realism, as in (a). MaGS resolves this by allowing 3D Gaussians to roam on the mesh surface (using RGD), supporting joint optimization of mesh and Gaussians, as shown in (b).
    }
    \label{fig:biased}
\end{figure}

\begin{table*}[]
\centering
\caption{\textbf{Quantitative results on PeopleSnapshot dataset~\cite{alldieck_video_2018}.} We present the average PSNR/SSIM/VGG-LPIPS values for novel pose synthesis on  PeopleSnapshot, with each cell colored to indicate the \colorbox{best}{best}, \colorbox{second}{second best} and \colorbox{third}{third best}.}
\label{tab:peoplesnapshot_dataset_full}
\resizebox{\textwidth}{!}{%
\begin{tabular}{@{}lcccccccccccc@{}}
\toprule
 &
  \multicolumn{3}{c}{male-3-casual} &
  \multicolumn{3}{c}{male-4-casual} &
  \multicolumn{3}{c}{female-3-casual} &
  \multicolumn{3}{c}{female-4-casual} \\
\multirow{-2}{*}{Methods} &
  PSNR ↑ &
  SSIM ↑ &
  LPIPS ↓ &
  PSNR ↑ &
  SSIM ↑ &
  LPIPS ↓ &
  PSNR ↑ &
  SSIM ↑ &
  LPIPS ↓ &
  PSNR ↑ &
  SSIM ↑ &
  LPIPS ↓ \\ \midrule
Anim-NeRF \cite{chen_animatable_2021}&
  29.37 &
  0.9700 &
  \cellcolor[HTML]{FFFF8F}0.0170 &
  28.37 &
  0.9600 &
  0.0270 &
  28.91 &
  0.9740 &
  \cellcolor[HTML]{FFFF8F}0.0220 &
  28.90 &
  0.9680 &
  \cellcolor[HTML]{FFFF8F}0.0170 \\
InstantAvatar \cite{jiang_instantavatar_2022}&
  30.91 &
  \cellcolor[HTML]{FAC791}0.9770 &
  0.0220 &
  29.77 &
  \cellcolor[HTML]{FAC791}0.9800 &
  \cellcolor[HTML]{FFFF8F}0.0250 &
  29.73 &
  \cellcolor[HTML]{FFFF8F}0.9750 &
  0.0250 &
  30.92 &
  \cellcolor[HTML]{FAC791}0.9770 &
  0.0210 \\
3DGS-Avatar~\cite{qian_3dgs-avatar_2023}&
  \cellcolor[HTML]{FFFF8F}34.28 &
  0.9724 &
  \cellcolor[HTML]{FAC791}0.0149 &
  \cellcolor[HTML]{FFFF8F}30.22 &
  0.9653 &
  \cellcolor[HTML]{FAC791}0.0231 &
  \cellcolor[HTML]{FFFF8F}30.57 &
  0.9581 &
  \cellcolor[HTML]{F59194}0.0209 &
  \cellcolor[HTML]{FFFF8F}33.16 &
  0.9678 &
  \cellcolor[HTML]{FAC791}0.0157 \\
SplattingAvatar~\cite{shao_splattingavatar_2024}&
  \cellcolor[HTML]{FAC791}36.48 &
  \cellcolor[HTML]{FFFF8F}0.9766 &
  0.0247 &
  \cellcolor[HTML]{FAC791}33.98 &
  \cellcolor[HTML]{FFFF8F}0.9776 &
  0.0340 &
  \cellcolor[HTML]{FAC791}37.36 &
  \cellcolor[HTML]{FAC791}0.9754 &
  0.0345 &
  \cellcolor[HTML]{FAC791}35.25 &
  \cellcolor[HTML]{FFFF8F}0.9734 &
  0.0271 \\
Ours&
  \cellcolor[HTML]{F59194}38.94 &
  \cellcolor[HTML]{F59194}0.9854 &
  \cellcolor[HTML]{F59194}0.0110 &
  \cellcolor[HTML]{F59194}35.19 &
  \cellcolor[HTML]{F59194}0.9842 &
  \cellcolor[HTML]{F59194}0.0175 &
  \cellcolor[HTML]{F59194}39.31 &
  \cellcolor[HTML]{F59194}0.9829 &
  \cellcolor[HTML]{FAC791}0.0215 &
  \cellcolor[HTML]{F59194}37.83 &
  \cellcolor[HTML]{F59194}0.9828 &
  \cellcolor[HTML]{F59194}0.0116 \\ \bottomrule
\end{tabular}%
}
\vspace{-10pt}
\end{table*}

\begin{figure*}[]
    \centering
    \addtolength{\tabcolsep}{-6.5pt}
    \footnotesize{
        \setlength{\tabcolsep}{1pt} %
        \begin{tabular}{p{8.2pt}cccccc}
            & 4D-GS & SC-GS & D-Miso & Grid4D & Ours & GT 
            \\
            \raisebox{35pt}{\rotatebox[origin=c]{90}{Jumpingjacks}}&
             \includegraphics[width=0.145\textwidth]{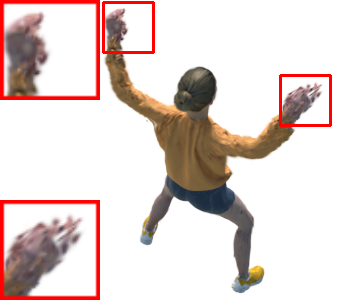} &
            \includegraphics[width=0.145\textwidth]{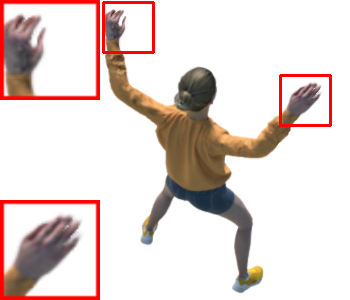} & 
            \includegraphics[width=0.145\textwidth]{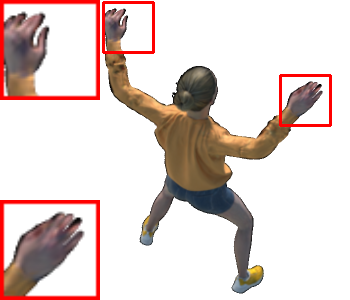} &
            \includegraphics[width=0.145\textwidth]{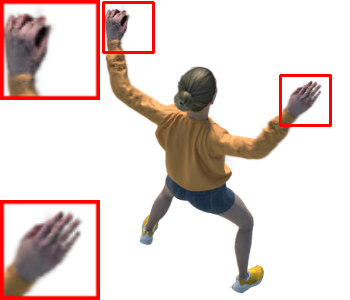} &
            \includegraphics[width=0.145\textwidth]{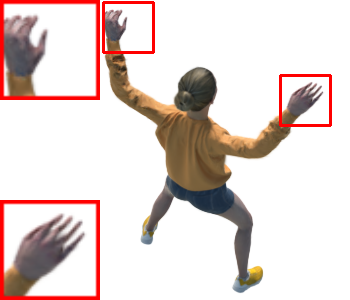} &
            \includegraphics[width=0.145\textwidth]{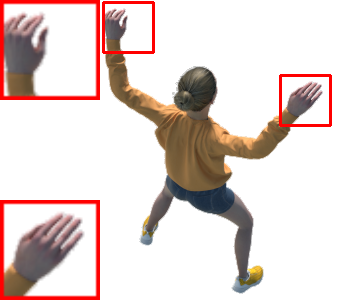}      
            \\
            \raisebox{35pt}{\rotatebox[origin=c]{90}{HellWarrior}}&
             \includegraphics[width=0.145\textwidth]{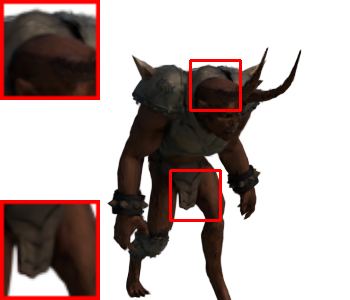} &
            \includegraphics[width=0.145\textwidth]{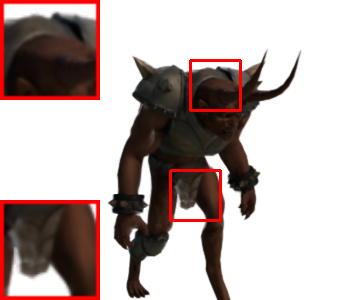} & 
            \includegraphics[width=0.145\textwidth]{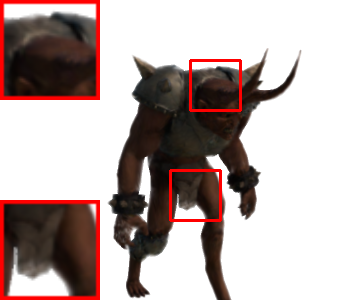} &
            \includegraphics[width=0.145\textwidth]{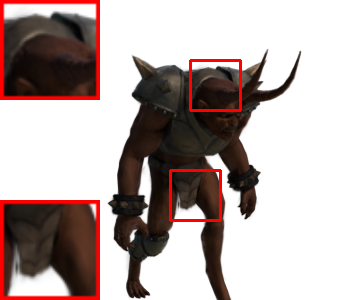} &
            \includegraphics[width=0.145\textwidth]{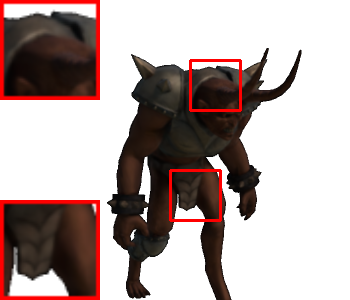} &
            \includegraphics[width=0.145\textwidth]{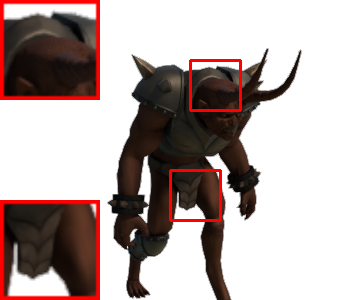}
             \\
        \end{tabular}
    }
	\caption{\textbf{Qualitative comparisons on D-NeRF~\cite{pumarola_d-nerf_2020}.} We compare MaGS with 4D-GS~\cite{wu_4d_2023}, SC-GS~\cite{huang_sc-gs_2023}, D-Miso~\cite{waczynska_d-miso_2024}, and Grid4D~\cite{xu_grid4d_2024}. }
\label{fig:dnerf_qualitative}
\vspace{-10pt}
\end{figure*}

There have been several attempts to integrate 3DGS with geometric priors, such as SC-GS~\cite{huang_sc-gs_2023}, which uses sparse point control methods, and D-Miso~\cite{waczynska_d-miso_2024}, which employs discontinuous mesh surfaces for reconstruction and editing. DG-Mesh~\cite{liu_dynamic_2024} and SplattingAvatar~\cite{shao_splattingavatar_2024} bind Gaussians to mesh surfaces, enabling mesh-based dynamic reconstruction. DynaSurfGS~\cite{cai_dynasurfgs_2024} and Dynamic 2DGS~\cite{zhang_dynamic_2024} also extract meshes at arbitrary time points via rendering. Additionally, SuGaR~\cite{guedon_sugar_2023}, PGSR~\cite{chen_pgsr_2024}, Mani-GS~\cite{gao_2024_mani}, and GaMeS~\cite{waczynska_games_2024} integrate Gaussian representations with mesh structures to achieve simulation capabilities; however, as these methods are designed for static scenes, they lack mechanisms to enable simulated deformation.

MaGS offers three key advantages over existing methods: \textbf{1) }It incorporates a continuous mesh that can be directly utilized for simulation, distinguishing it from methods like SC-GS and D-Miso. For point cloud-based methods, when large edits are performed, the lack of surface information provided by the mesh leads to the issue shown in Figure~\ref{fig:compare_simu}. \textbf{2)} MaGS allows Gaussians to roam along the mesh surface. In contrast, DG-Mesh and SplattingAvatar do not support the mesh-Gaussian displacement, restricting the flexibility of Gaussians as discussed in Figure~\ref{fig:biased}. \textbf{3)} MaGS preserves mesh continuity across frames, ensuring consistent point-and-facet correspondence over time. \textbf{This continuity is crucial for us to inherit the Mesh-adsorbed Gaussian from reconstruction to simulation}, and represents a notable improvement over methods like DynaSurfGS and Dynamic 2DGS, which generate independent meshes per frame using TSDF without maintaining cross-frame consistency.

{
\setstretch{0.96}
\begin{figure*}[ht]
    \centering
\includegraphics[width=\linewidth,trim=20 102 15 195,clip]{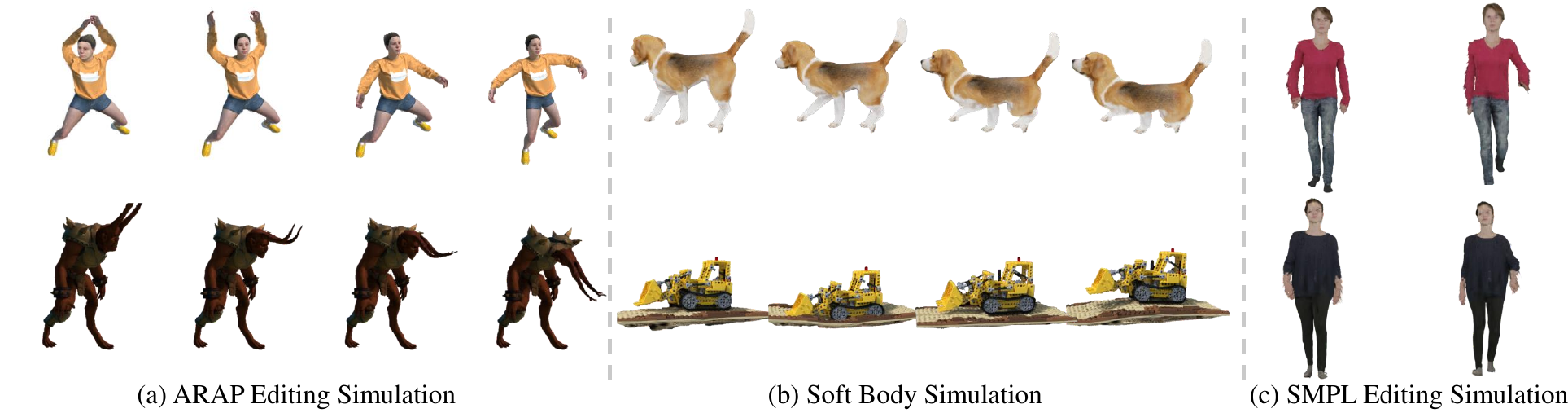}
	\caption{\textbf{Mesh-guided Simulation of MaGS.} We utilize user-edited meshes—modified through methods such as dragging, soft-body collisions, or SMPL-based motion editing—to guide deformation and achieve natural rendering results.}
        \label{fig:jumpingjacks_simu}
\vspace{-5pt}
\end{figure*}

\section{Experiments}
\label{sec:experiments}
\subsection{Experimental Settings}
We evaluated MaGS on three datasets: D-NeRF dataset~\cite{pumarola_d-nerf_2020}, DG-Mesh dataset~\cite{liu_dynamic_2024}, and PeopleSnapshot dataset~\cite{alldieck_video_2018}. D-NeRF dataset consists of eight synthetic monocular scenes, each evaluated at a resolution of $400 \times 400$. DG-Mesh dataset, also evaluated at a resolution of $400 \times 400$, provides ground-truth meshes for each frame, enabling the evaluation of dynamic reconstruction accuracy. PeopleSnapshot dataset contains monocular videos of individuals and is evaluated at the original resolution of $1080 \times 1080$. Following previous work~\cite{shao_splattingavatar_2024}, we use optimized SMPL parameters from InstantAvatar~\cite{jiang_instantavatar_2022} to generate guide meshes. All experiments were performed on an NVIDIA RTX 4090.

The performance metrics used for evaluation include Peak Signal-to-Noise Ratio (PSNR), Structural Similarity Index (SSIM), Multiscale Structural Similarity Index (MS-SSIM)~\cite{wang_multiscale_2003}, and Learned Perceptual Image Patch Similarity (LPIPS)~\cite{zhang_unreasonable_2018}. We also use Earth Mover's Distance (EMD) and Chamfer Distance (CD) for DG-Mesh to evaluate reconstruction accuracy based on the ground-truth mesh.

\subsection{Quantitative Comparisons}

\textbf{D-NeRF. }
We evaluate MaGS against state-of-the-art methods on the D-NeRF dataset.
Table~\ref{tab:dnerf_dataset} presents the evaluation results across seven scenes, excluding Lego\footnote{Yang \emph{et al.}~\cite{yang_deformable_2023} noted inconsistencies in D-NeRF's Lego scene and provided a corrected version; we tested both. See Appendix for details.}.
MaGS achieves higher PSNR, SSIM, and LPIPS metrics across most scenes, surpassing SC-GS \cite{huang_sc-gs_2023} (the second-best method) with an average PSNR improvement of 0.7 dB.

\begin{table}[t!]
\centering
\caption{\textbf{Quantitative results on DG-Mesh~\cite{liu_dynamic_2024}.} We use \textbf{*} to indicate the data provided by DynaSurfGS.}
\label{tab:dgmesh_dataset_short}
\resizebox{0.9\columnwidth}{!}{%
\begin{tabular}{@{}lccccc@{}}
\toprule
Methods    & CD↓    & EMD↓   & PSNR↑ & Time↓ & FacesNum \\ \midrule
D-NeRF*~\cite{pumarola_d-nerf_2020}     & 1.1506 & 0.1710 & 28.44 &   /    &     /         \\
K-Plane*~\cite{fridovich-keil_k-planes_2023}    & 0.9224 & 0.1440 & 31.13 &   /    &     /         \\
HexPlane*~\cite{cao_hexplane_2023}   & 1.9072 & 0.1474 & 30.18 &   /    &     /         \\
TiNeuVox-B*~\cite{fang_fast_2022} & 2.5186 & 0.1666 & 31.96 &   /    &     /         \\
DG-Mesh~\cite{liu_dynamic_2024}              & \cellcolor[HTML]{FAC791}0.6022 & \cellcolor[HTML]{FFFF8F}0.1192 & 31.43                         & \cellcolor[HTML]{FFFF8F}89.3 & 170,232  \\
DynaSurfGS*~\cite{cai_dynasurfgs_2024}           & 0.7570                         & \cellcolor[HTML]{FAC791}0.1136 & \cellcolor[HTML]{FFFF8F}33.18 &              /                &     /    \\
Dynamic 2D Gaussians~\cite{zhang_dynamic_2024} & \cellcolor[HTML]{F59194}0.5254 & 0.1260                         & \cellcolor[HTML]{FAC791}36.40 & \cellcolor[HTML]{FAC791}72.7 & 1,419,454 \\
Ours                 & \cellcolor[HTML]{FFFF8F}0.6662 & \cellcolor[HTML]{F59194}0.1106 & \cellcolor[HTML]{F59194}40.76 & \cellcolor[HTML]{F59194}47.6 & 981     \\ \bottomrule
\end{tabular}%
}
\end{table}

\textbf{DG-Mesh. }
Table~\ref{tab:dgmesh_dataset_short} shows that MaGS achieves leading performance in PSNR and EMD on the DG-Mesh dataset. MaGS performs closely to the state-of-the-art in Chamfer Distance (CD), using only 981 mesh faces compared to 170,232 for DG-Mesh and 1,419,454 for Dynamic 2D Gaussians. Moreover, MaGS requires only 47.6 minutes for optimization, which is considerably faster than DG-Mesh (89.3 min) and Dynamic 2D Gaussians (72.7 min).

\textbf{PeopleSnapshot. }
We evaluate MaGS on the PeopleSnapshot dataset~\cite{alldieck_video_2018}. Table~\ref{tab:peoplesnapshot_dataset_full} shows that MaGS achieves the highest PSNR and SSIM metrics across all scenarios, surpassing SplattingAvatar. MaGS demonstrates an approximately 2 dB average PSNR improvement over the next best method and also outperforms in LPIPS.

\subsection{Qualitative Comparisons and Simulation}

We present qualitative comparisons on two datasets: D-NeRF and DG-Mesh. Figure~\ref{fig:dnerf_qualitative} illustrates the results on D-NeRF, with magnified images highlighting the finer details of the synthesized outputs. Our method achieves superior visual quality, producing sharper and more accurate reconstructions.  Similarly, Figure~\ref{fig:dgmesh_qualitative} shows a comparison on DG-Mesh, where we compare our method with the state-of-the-art dynamic mesh extraction techniques. Our method generates meshes that more closely resemble the ground truth, demonstrating improved accuracy in reconstruction.

In terms of simulation, Figure~\ref{fig:jumpingjacks_simu} showcases the results produced by MaGS. As shown, our method effectively preserves texture throughout the deformation process. Additionally, Figure~\ref{fig:compare_simu} compares MaGS with SC-GS, a leading simulation method based on sparse guide points. MaGS avoids the surface fracture problems in SC-GS when subjected to large deformations.

\subsection{Ablation Studies}
We ablate MaGS to understand the contribution of key design choices with the default settings, including the RMD-Net, Gaussians Hover, and RGD-Net, as shown in Table~\ref{tab:ablation_dnerf_short}, removing both RMD-Net and RGD-Net results in a PSNR drop of 2.92, which is a 6.6\% decrease. Disabling Gaussians Hover reduces PSNR by 2.19, a 5.0\% reduction while excluding RGD-Net alone leads to a PSNR drop of 1.08, or 2.4\%. The full MaGS configuration achieves the highest PSNR of 44.06, highlighting the importance of each component in enhancing the model's performance.
\begin{table}[t]
\centering
\caption{\textbf{Ablation experiments on the D-NeRF~\cite{pumarola_d-nerf_2020}}.}
\label{tab:ablation_dnerf_short}
\resizebox{0.9\columnwidth}{!}{%
\begin{tabular}{@{}lccc@{}}
\toprule
Method               & PSNR↑          & SSIM↑           & LPIPS↓          \\ \midrule
MaGS w/o RMD-Net and RGD-Net & 41.14          & 0.9974          & 0.0064          \\
MaGS w/o Gaussian Hover       & 41.87          & 0.9977          & 0.0059          \\
MaGS w/o RGD-Net         & 42.98          & 0.9982          & 0.0047 \\
MaGS Full       & \cellcolor[HTML]{F59194}{44.06} & \cellcolor[HTML]{F59194}{0.9985} & \cellcolor[HTML]{F59194}{0.0040}          \\ \bottomrule
\end{tabular}%
}
\end{table}

\begin{figure}[t]
  \centering
  
  \begin{subfigure}{0.24\linewidth} 
    \centering 
    \includegraphics[width=\linewidth,trim=100 100 100 100,clip]{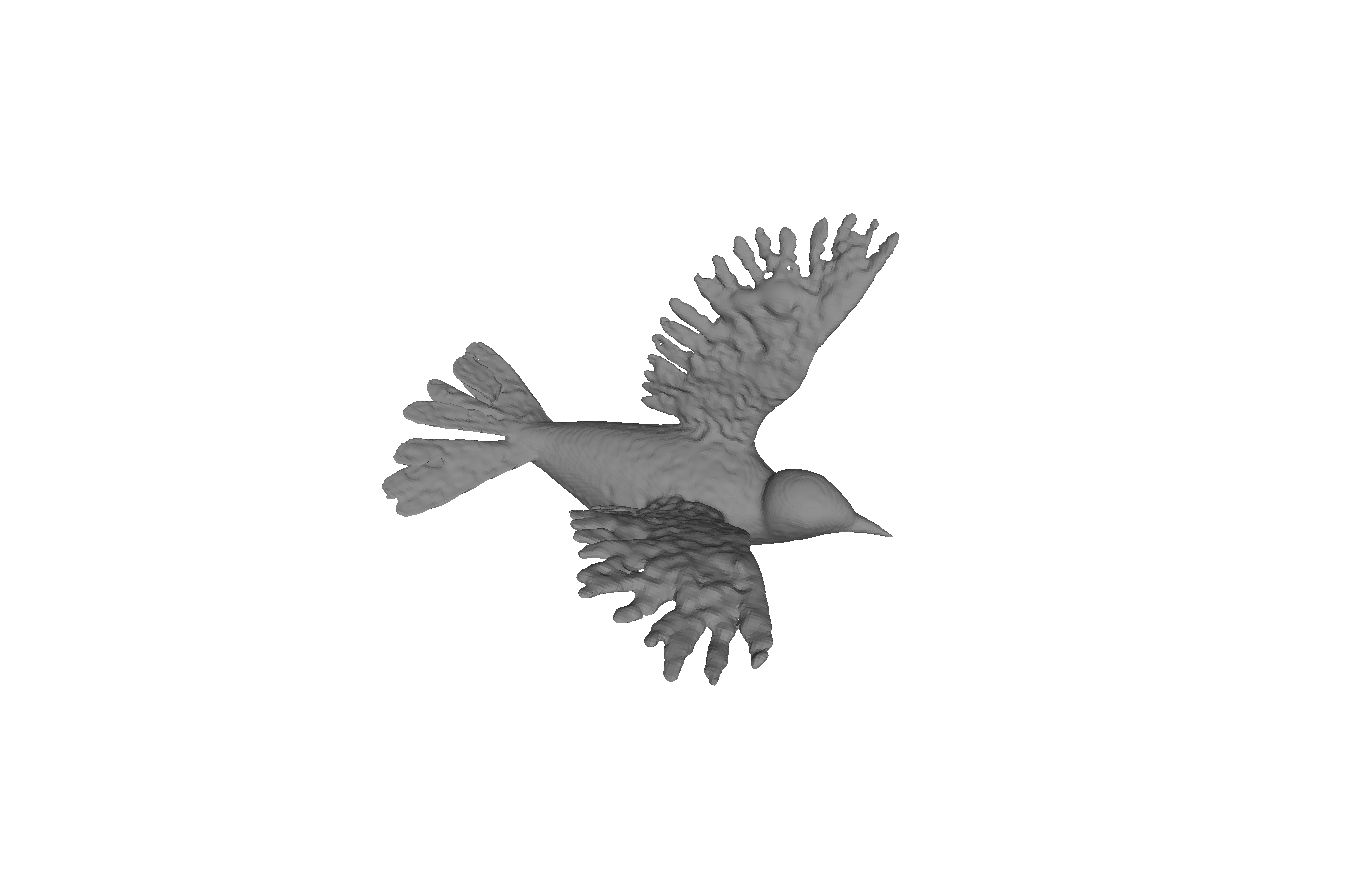} 
    \caption{\scriptsize DG-Mesh} 
  \end{subfigure}
  \hfill
  \begin{subfigure}{0.24\linewidth} 
    \centering 
    \includegraphics[width=\linewidth,trim=100 100 100 100,clip]{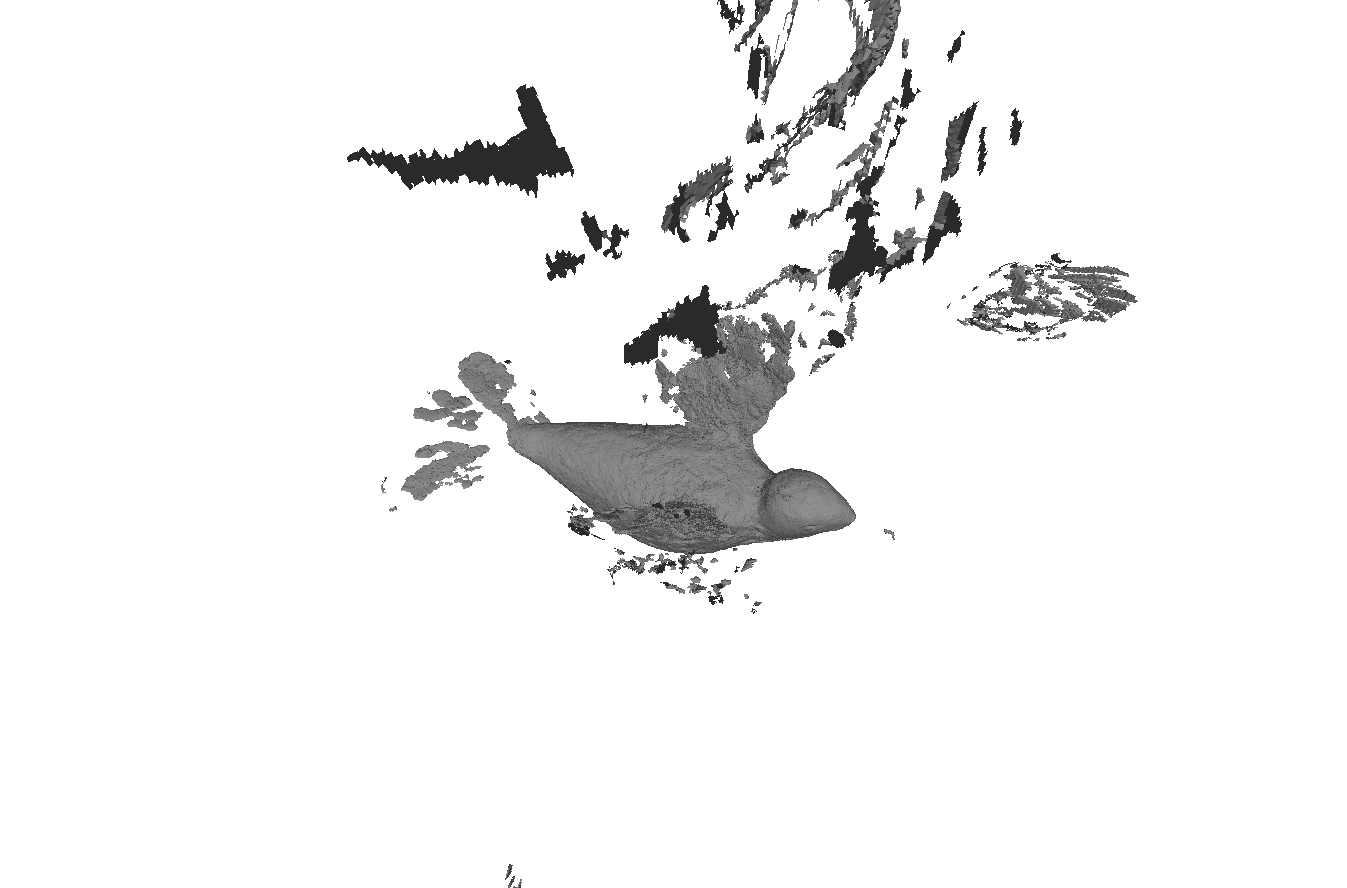} 
    \caption{\scriptsize Dynamic 2DGS} 
  \end{subfigure}
  \hfill
  \begin{subfigure}{0.24\linewidth} 
    \centering 
    \includegraphics[width=\linewidth,trim=100 100 100 100,clip]{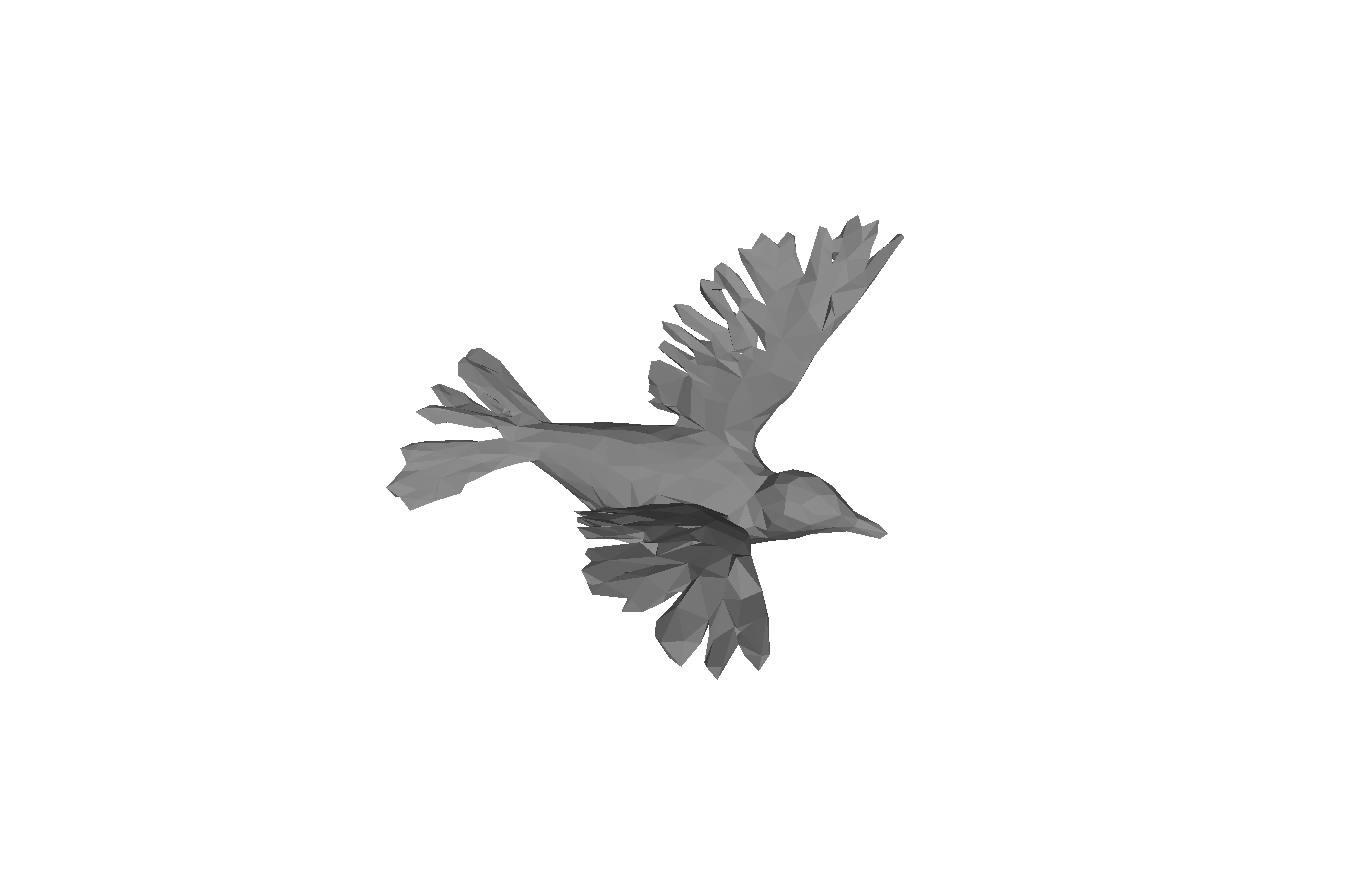} 
    \caption{\scriptsize Ours} 
  \end{subfigure}
  \hfill
  \begin{subfigure}{0.24\linewidth} 
    \centering 
    \includegraphics[width=\linewidth,trim=100 100 100 100,clip]{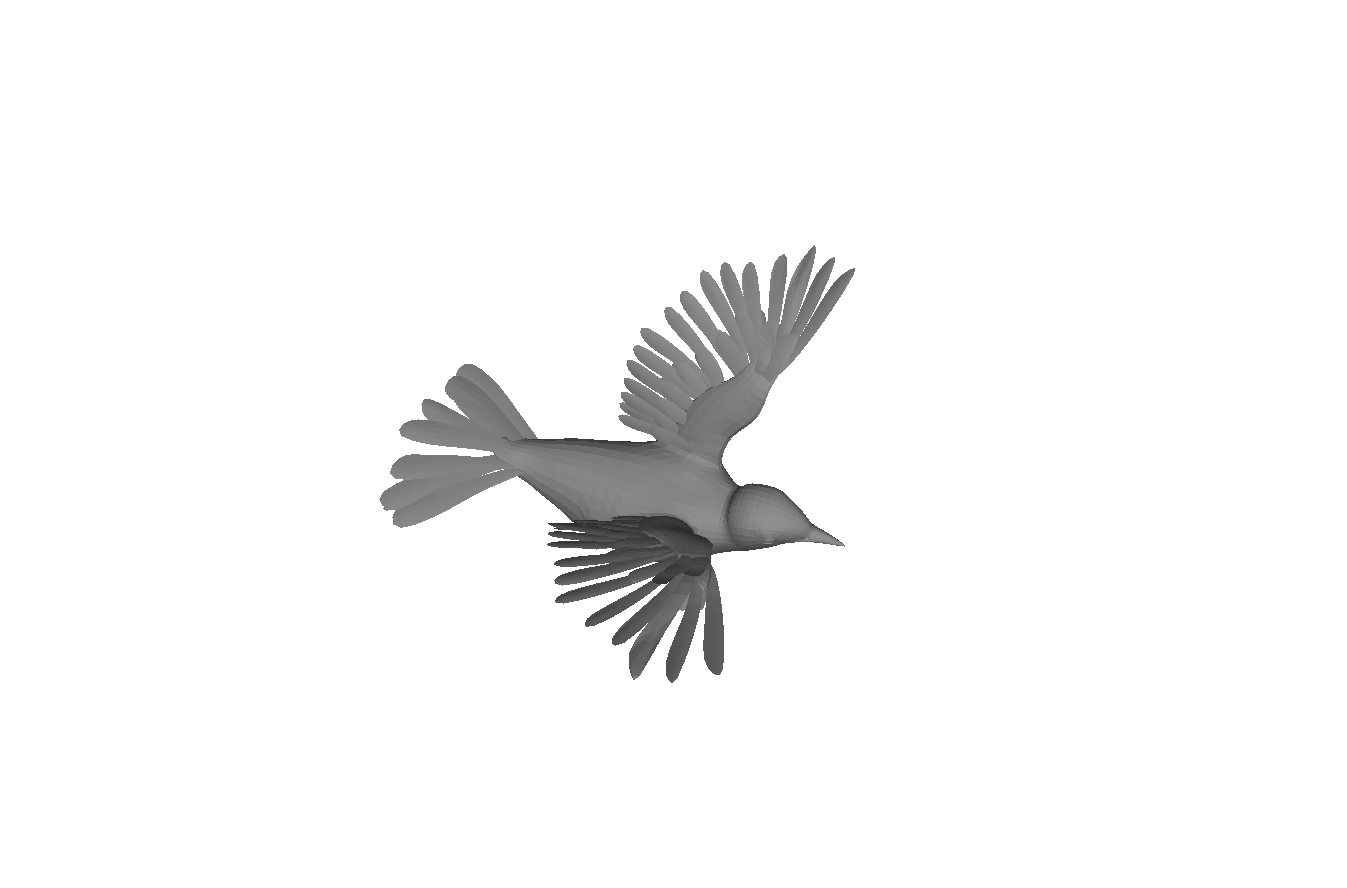} 
    \caption{\scriptsize GT} 
  \end{subfigure}
  
  \caption{\textbf{Qualitative Comparison on the DG-Mesh dataset~\cite{liu_dynamic_2024} in terms of mesh quality.} %
  }
  \label{fig:dgmesh_qualitative}
\end{figure}

}
\section{Conclusion}
This paper introduces the MaGS method, which addresses 3D reconstruction and simulation within a unified framework. It creates a novel adsorbed mesh-Gaussian 3D representation by constraining 3D Gaussians to roam near the mesh surface, which combines the rendering flexibility of 3D Gaussians with the adaptability of meshes to different geometric priors. %
MaGS is compatible with various deformation priors like ARAP, SMPL, and soft physics simulation. Extensive experiments on D-NeRF, DG-Mesh, and PeopleSnapshot demonstrate that MaGS achieves SOTA performance in both reconstruction and simulation.

\textbf{Limitations and Future Work. }Despite these successes, MaGS has limitations: 1) it struggles with multi-object interaction scenarios; 2) it requires video data with diverse viewpoints for reliable mesh construction—an inherent challenge for mesh-based methods. Future work will focus on addressing these challenges by integrating generative models to enhance MaGS's capabilities and robustness under uncertain or incomplete viewpoint conditions.

\newpage
{
    \small
    \bibliographystyle{ieeenat_fullname}
    \bibliography{main}
}

\clearpage
\setcounter{page}{1}
\setcounter{section}{0}
\setcounter{figure}{0}
\setcounter{table}{0}
\maketitlesupplementary
\renewcommand{\thesection}{\Alph{section}}

\section*{Supplementary Material Overview}
This supplementary material provides additional visuals and analyses to complement the main content of the paper. The sections are organized as follows:

\begin{itemize}
\item %
Section~\ref{sec:implementation} includes implementation details for MPE-Net, RMD-Net, and RGD-Net. Refer to Figures~\ref{fig:MPE-Net}, \ref{fig:RMD-Net}, and \ref{fig:enter-label} for visual representations of the models.

\item %
Section~\ref{sec:training-config} describes hyperparameters. Specific details on opacity resets, densification, and MaGS parameters are included.

\item %
Section~\ref{sec:algorithm} outlines the steps of the MaGS pipeline using pseudo-code (Algorithm~\ref{alg:MaGS}). It covers mesh refinement and Mesh-adsorbed Gaussian optimization from input to output.

\item %
Section~\ref{sec:add_experiments} includes qualitative results comparing our method with existing approaches across datasets such as DG-Mesh, D-NeRF, and PeopleSnapshot. Visualizations in Figures~\ref{fig:add_dgmesh_qualitative}, \ref{fig:add_dnerf_heatmap}, and \ref{fig:add_peoplesnapshot_heatmap} highlight reconstruction accuracy and rendering quality.

\item %
Section~\ref{sec:simulations} showcases the results of simulations. Figures~\ref{fig:add_lego_simu}, \ref{fig:add_horse_simu}, and \ref{fig:add_multigs_simu} demonstrate complex object shapes and dynamic motions in a simulated environment.

\item %
Section~\ref{sec:quantitative} presents a detailed quantitative analysis of MaGS. Metrics such as L1 loss, PSNR, and SSIM are reported in Tables~\ref{tab:ablation_dnerf_full}, \ref{tab:dgmesh_dataset_full2}, \ref{tab:add_lego}.

\item %
 Section~\ref{sec:performance} summarizes performance evaluations of MaGS in terms of FPS across scenes in the D-NeRF dataset. Results in Table~\ref{tab:add_performance_benchmark}.
\end{itemize}

\section{Implementation Details}
\label{sec:implementation}  %
\begin{figure}[!ht]
    \centering
    \includegraphics[width=1\linewidth,trim=200 30 200 0,clip]{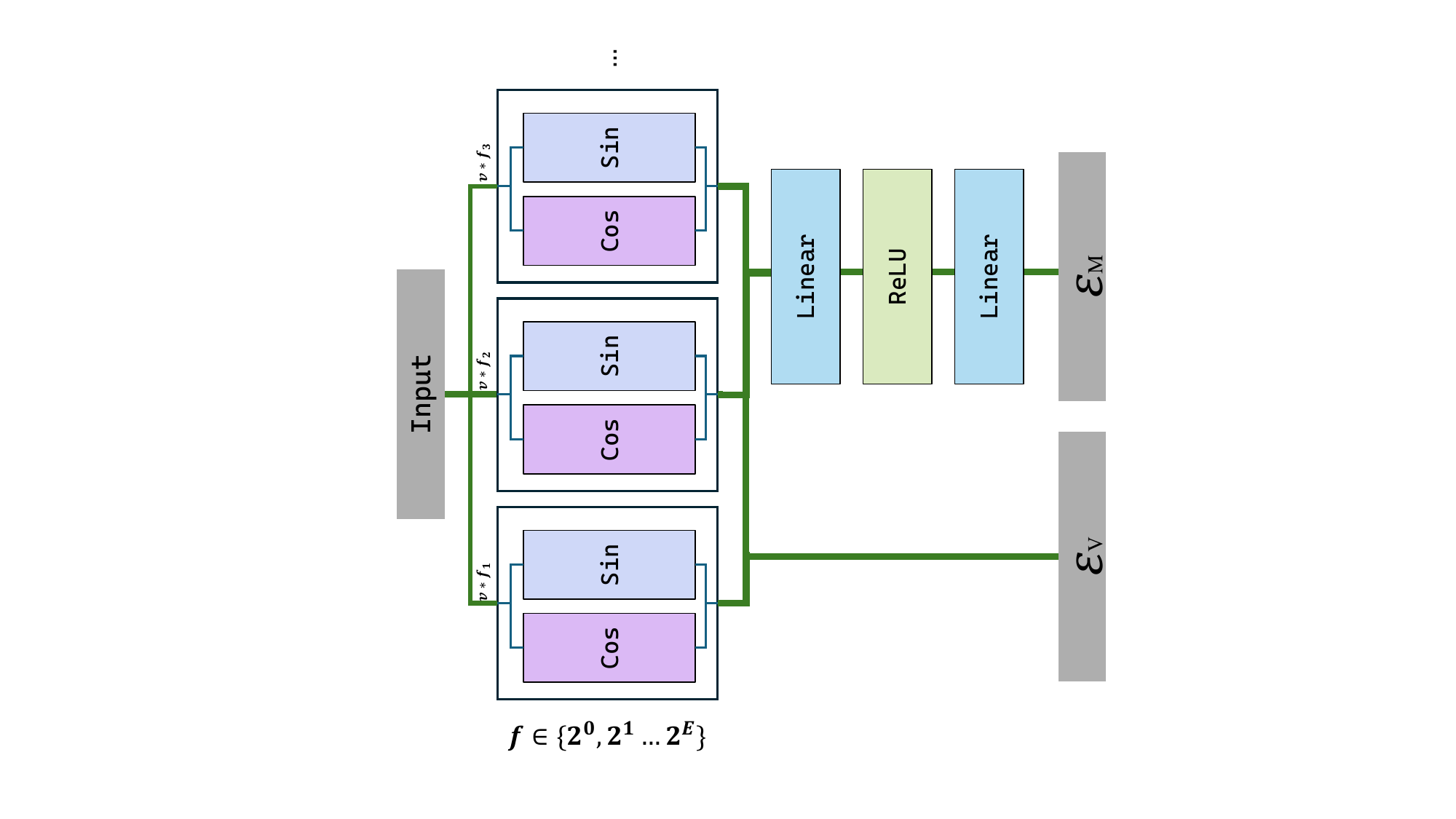}
    \caption{Structure of MPE-Net.}
    \label{fig:MPE-Net}
\end{figure}

In MPE-Net, as illustrated in Figure~\ref{fig:MPE-Net}, we predict mesh pose ($ \mathcal{E}_\text{M} $) and vertex-specific deformations ($ \mathcal{E}_\text{V} $) based on a coarse guide mesh. The input consists of handle vertices ($ \mathcal{H} $) along with their corresponding normals ($ \mathcal{N} $), as well as the complete mesh for calculating vertex information. Each input vertex $v$ is encoded using positional encoding, represented as $
\Phi(v) = \begin{bmatrix} v, & p_1(f_1 v), & \dots, & p_2(f_{E+1} v) \end{bmatrix} $
where $ p_j \in \{\sin, \cos\} $ and $ f_i = 2^{i-1} $ and $E$ is a hypermeter. This encoded input $ \Phi(v) $ produces the vertex-specific embedding $ \mathcal{E}_\text{V} $, which is then passed through fully connected layers with ReLU activations. The output of these layers generates the global mesh pose embedding $ \mathcal{E}_\text{M} $. The model thus outputs both the mesh pose ($ \mathcal{E}_\text{M} $) and the vertex-specific deformations ($ \mathcal{E}_\text{V} $), enabling effective deformation prediction without requiring temporal cues. 

\begin{figure}[!ht]
    \centering
    \includegraphics[width=1\linewidth,trim=200 30 200 0,clip]{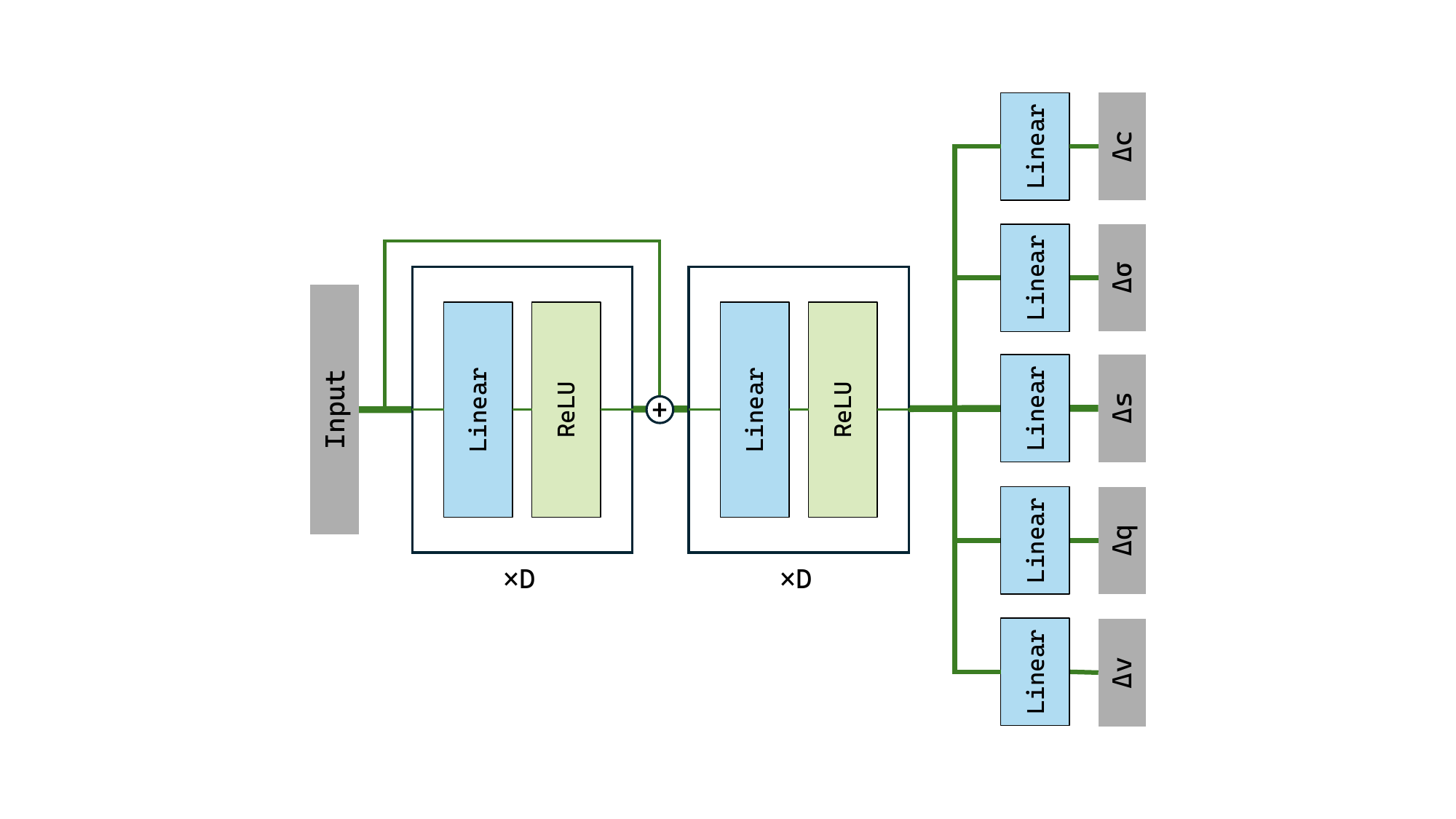}
    \caption{Structure of RMD-Net.}
    \label{fig:RMD-Net}
\end{figure}

We train RMD-Net using an MLP network $ F_\theta : (\mathcal{E}_\text{M}, \mathcal{E}_\text{V}) \to (\Delta v, \Delta q, \Delta s, \Delta \sigma, \Delta c) $. As depicted in Figure~\ref{fig:RMD-Net}, the MLP $ F_\theta $ processes the input through $ D $ fully connected layers with ReLU activations, producing an initial feature vector. In the fourth layer, we concatenate this feature vector with the input. The resulting combined representation is then passed through five additional fully connected layers, independently generating the outputs for $ \Delta v $, $ \Delta q $, $ \Delta s $, $ \Delta \sigma $, and $ \Delta c $.

\begin{figure}[!ht]
    \centering
    \includegraphics[width=1\linewidth,trim=200 100 200 100,clip]{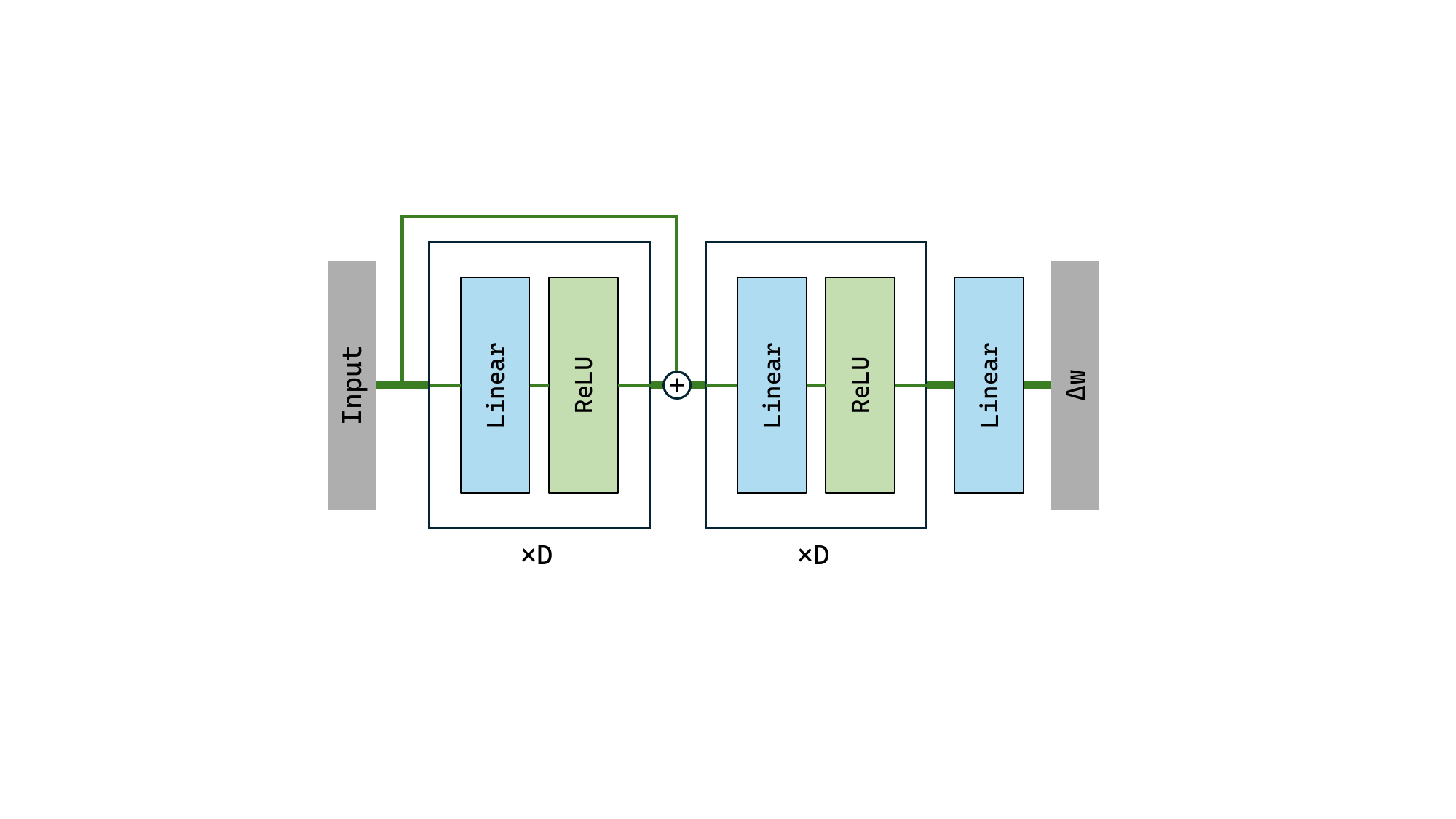}
    \caption{Structure of RGD-Net.}
    \label{fig:enter-label}
\end{figure}

As shown in Figure~\ref{fig:enter-label}, the structure of RGD-Net closely resembles that of RMD-Net, utilizing a similar MLP architecture. The key difference is that RGD-Net also takes $w$ as input. Additionally, instead of generating multiple outputs like RMD-Net, RGD-Net predicts a single output, $ \Delta w $.

\section{Training Configuration}
\label{sec:training-config}  %
The model is trained using a set of hyperparameters for both the optimization and loss functions. The optimization process utilizes a combination of learning rates for the Mesh-adsorbed Gaussians components, including $ w $-parameters ($ lr = 0.00016 $), feature optimization ($ lr = 0.0025 $), opacity ($ lr = 0.05 $), scaling ($ lr = 0.005 $), and rotation ($ lr = 0.001 $). The optimization of $ w $ follows a learning rate scheduler, which starts with an initial rate of $ 0.00016 $ and decays to $ 0.0000016 $ over $ 40,000 $ iterations. Densification occurs from iteration 100 to iteration 15,000, with a size threshold of 20 and a gradient threshold of $ 0.0002 $. Opacity is periodically reset starting from iteration 300, with a reset interval of 300,000 iterations. The loss function incorporates a structural similarity index (SSIM) term with a weight of $ \lambda_{ssim} = 0.2 $. For RMD-Net and RGD-Net, the network depth is controlled by the parameter $ D = 8 $, while for MPE-Net, $E = 10$ is selected. The optimizer is configured with a batch size of 2, and the opacity reset interval is set to 3000 iterations. The background is black, with both the \textit{random\_bg} and \textit{white\_bg} set to false. Additionally, 50 Mesh-adsorbed Gaussians are randomly initialized on each facet.

\section{Pseudo Code of Pipeline}
\label{sec:algorithm}
The following pseudo-code outlines the key steps involved in the MaGS pipeline for mesh refinement and Gaussian optimization.

\begin{algorithm}
\caption{MaGS Pipeline}\label{alg:MaGS}
\begin{algorithmic}[1]
\Require Video frames $F$, Initial mesh $M_0$
\Ensure Refined mesh $M$ and Gaussians $G$

\Procedure{MaGS\_Pipeline}{$F, M_0$}
    \State // Step 1: Initialize Gaussians
    \State $G \gets \textsc{InitializeGaussians}(M_0)$
    
    \For{each $f \in F$}
        \State // Step 2: Extract feature embeddings
        \State $(E_M, E_V) \gets \textsc{MPE\_Net}(M_0, f)$
        
        \State // Step 3: Predict mesh and Gaussian updates
        \State $(\Delta v, \Delta params) \gets \textsc{RMD\_Net}(E_M, E_V)$
        \State $\Delta w \gets \textsc{RGD\_Net}(E_M, E_V, G)$
        
        \State // Step 4: Update mesh and Gaussians
        \State $M \gets M + \Delta v$
        \State $G \gets G + \textsc{Interp}(\Delta params, G + \Delta w, M)$
        
        \State // Step 5: Render and compute loss
        \State $I \gets \textsc{RenderGaussians}(G)$
        \State $L \gets \textsc{Loss}(I, f.\text{gt})$
        
        \State // Step 6: Backpropagate loss
        \State \textsc{Backpropagate}(L, \{MPE\_NET\})
        \State \textsc{Backpropagate}(L, \{RMD\_NET\})
        \State \textsc{Backpropagate}(L, \{RGD\_NET\})
    \EndFor
    
    \State // Step 7: Final refinement
    \State $M, G \gets \textsc{Refine}(M, G)$
    \State \Return $M, G$
\EndProcedure
\end{algorithmic}
\end{algorithm}

\section{Additional Visualizations of Reconstruction}
\label{sec:add_experiments}  %
In this section, we present additional visualizations and comparisons to validate our findings further and demonstrate the performance of our methods.

\begin{figure*}
    \centering
    \addtolength{\tabcolsep}{-6.5pt}
    \footnotesize{
        \setlength{\tabcolsep}{1pt} %
        \begin{tabular}{p{8.2pt}cccc}
            & DG-Mesh & Dynamic 2DGS & Ours & GT\\
            \raisebox{35pt}{\rotatebox[origin=c]{90}{Beagle}}&
            \includegraphics[width=0.24\linewidth]{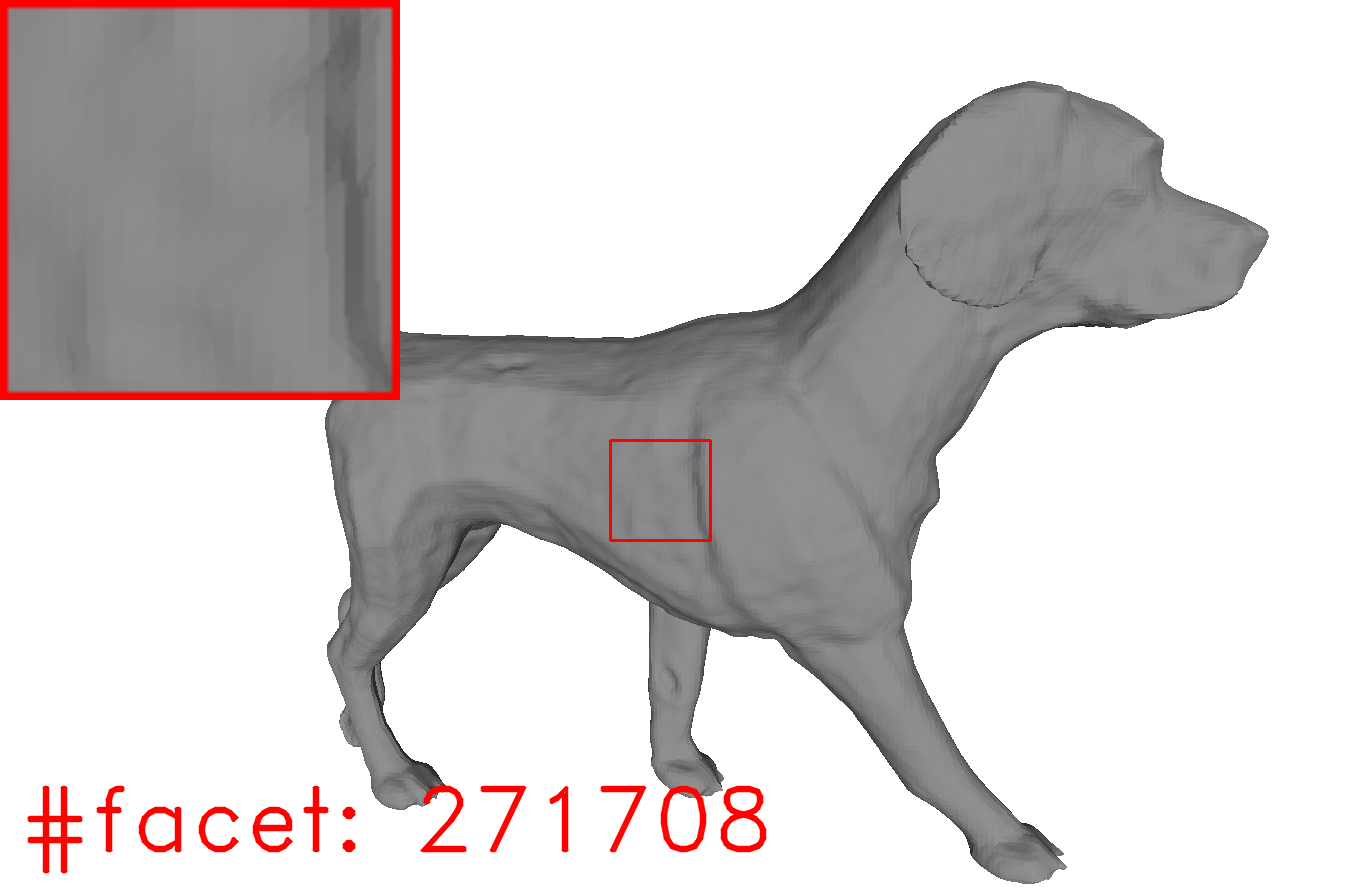}&
            \includegraphics[width=0.24\linewidth]{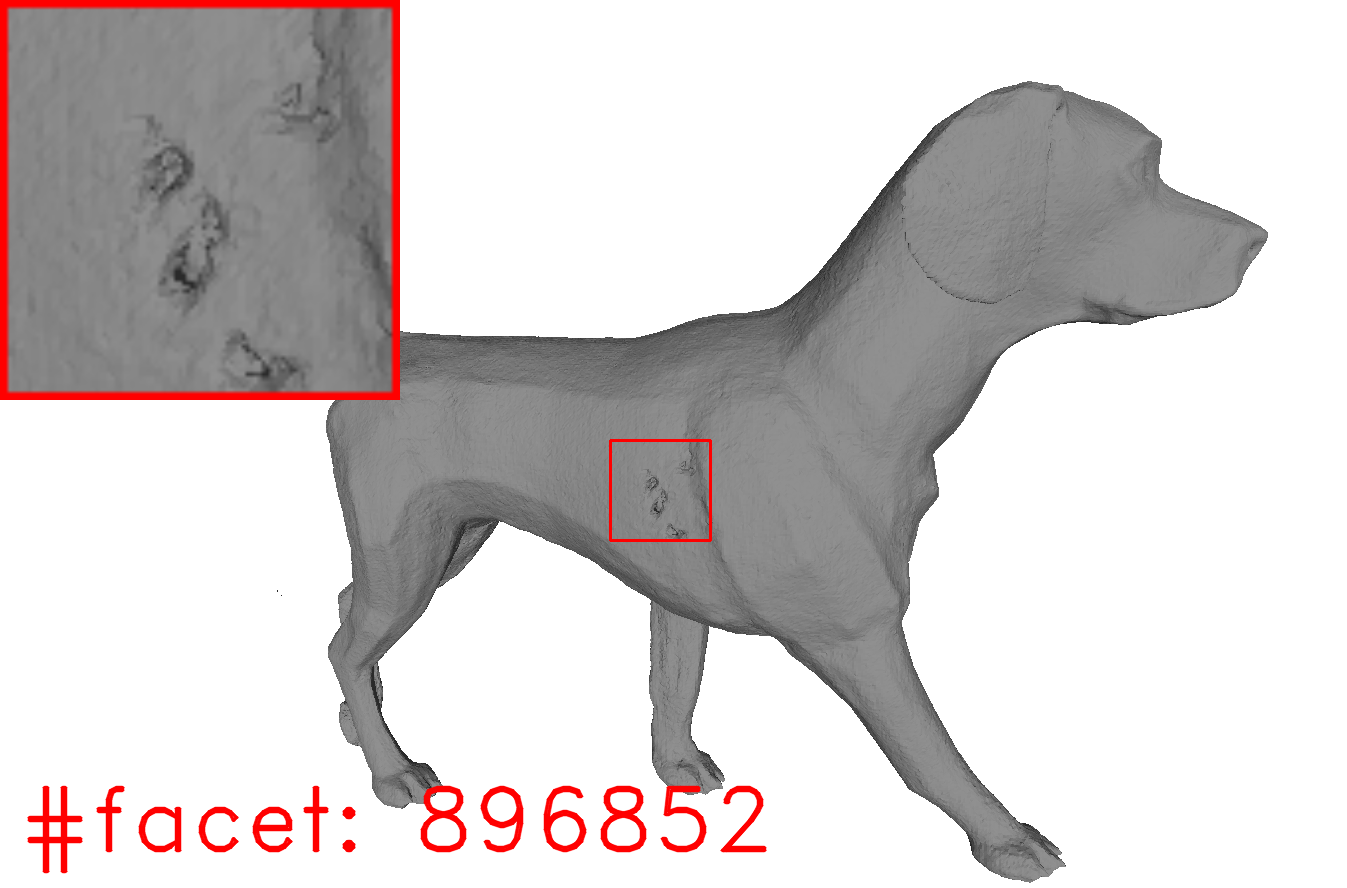}&
            \includegraphics[width=0.24\linewidth]{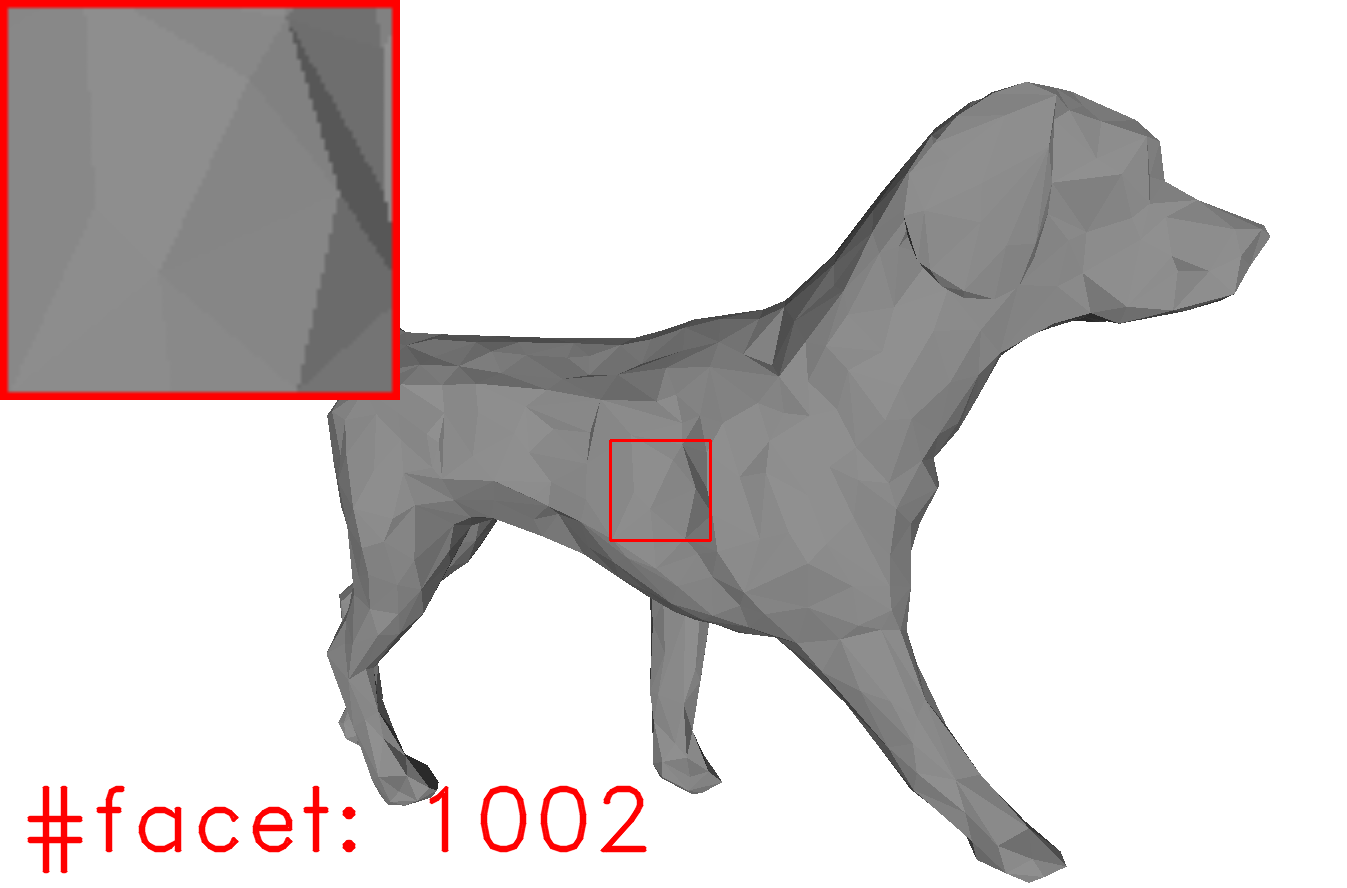}&
            \includegraphics[width=0.24\linewidth]{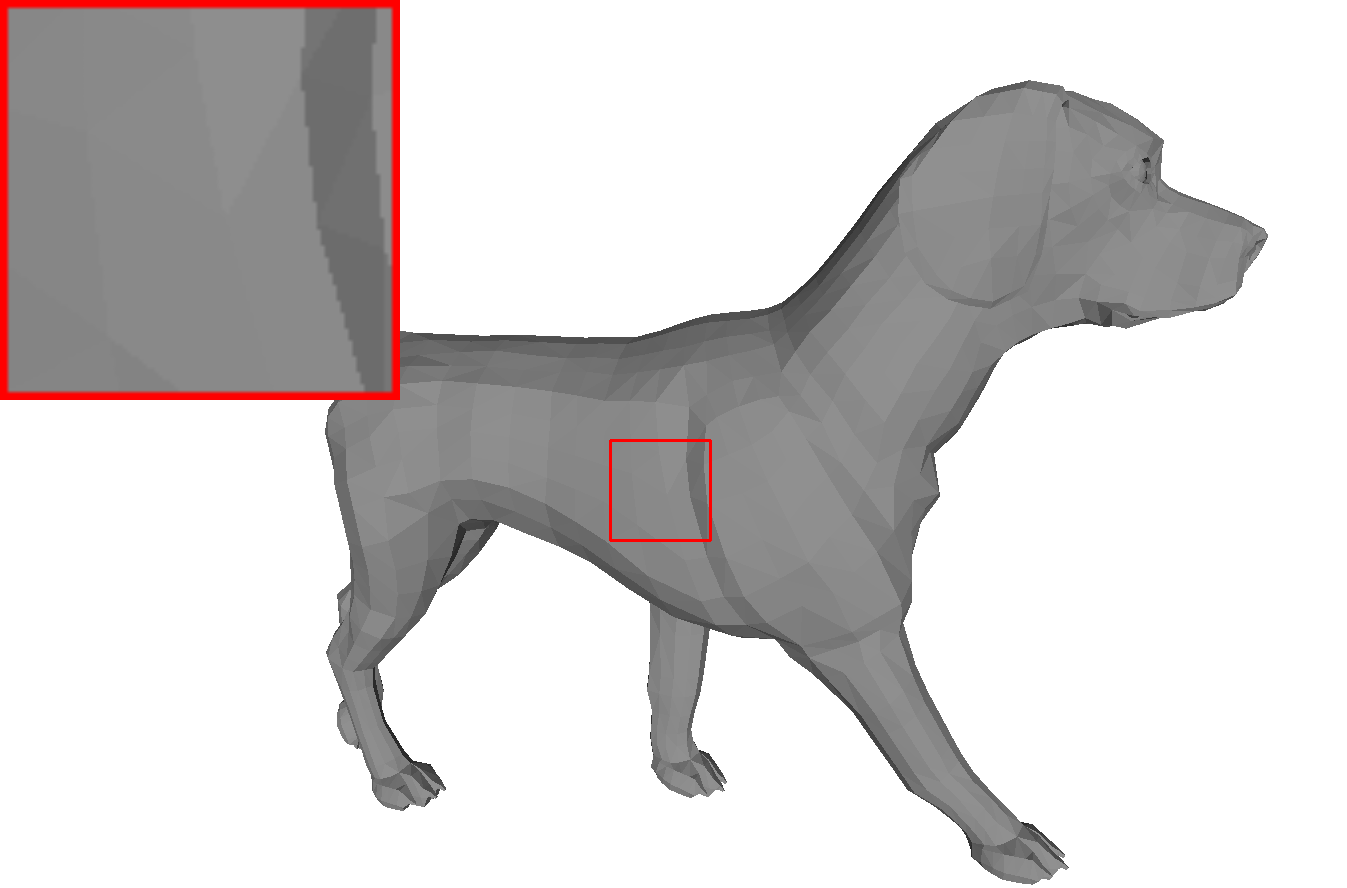}\\
            \raisebox{35pt}{\rotatebox[origin=c]{90}{Duck}}&
            \includegraphics[width=0.24\linewidth]{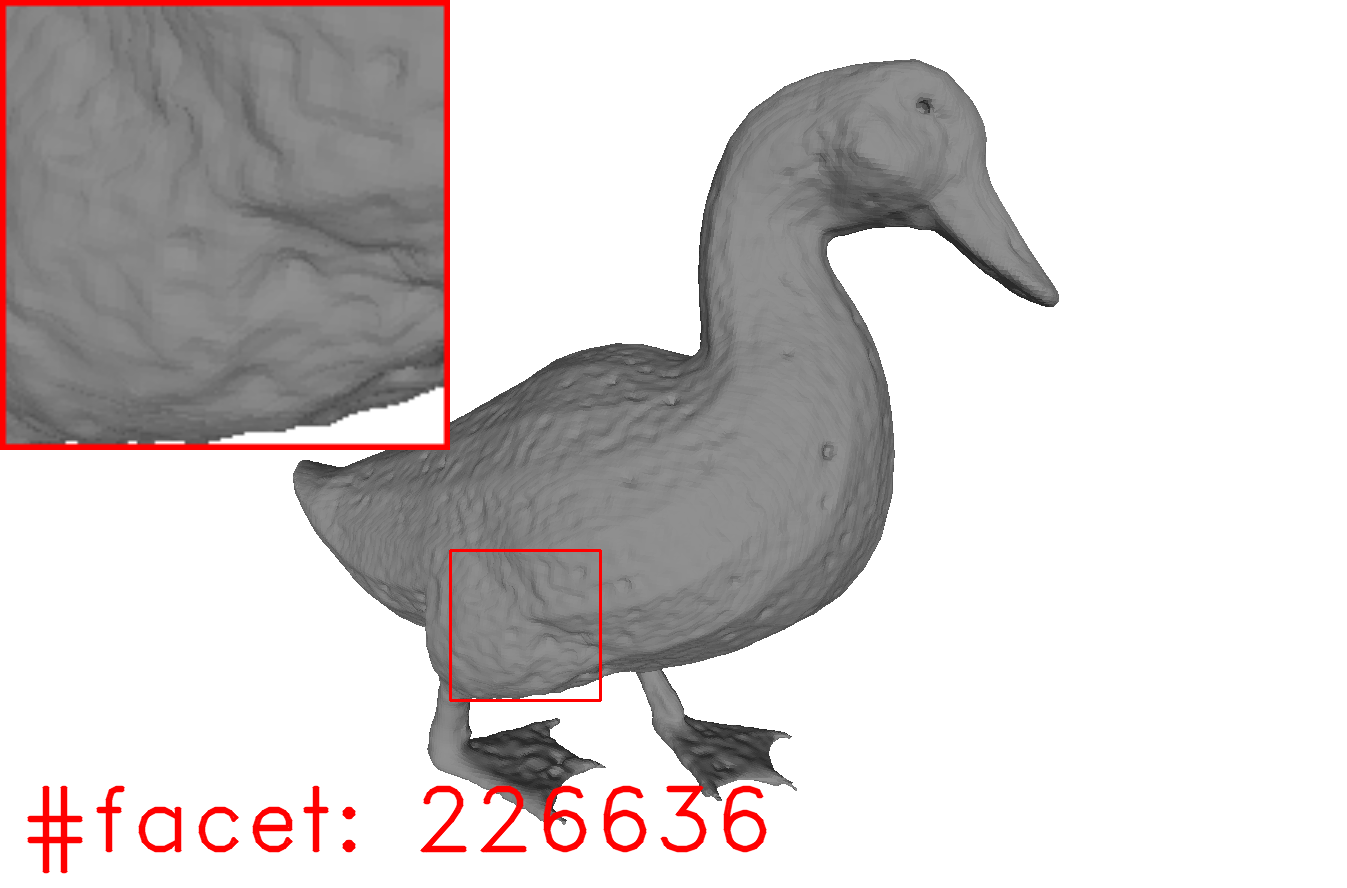}&
            \includegraphics[width=0.24\linewidth]{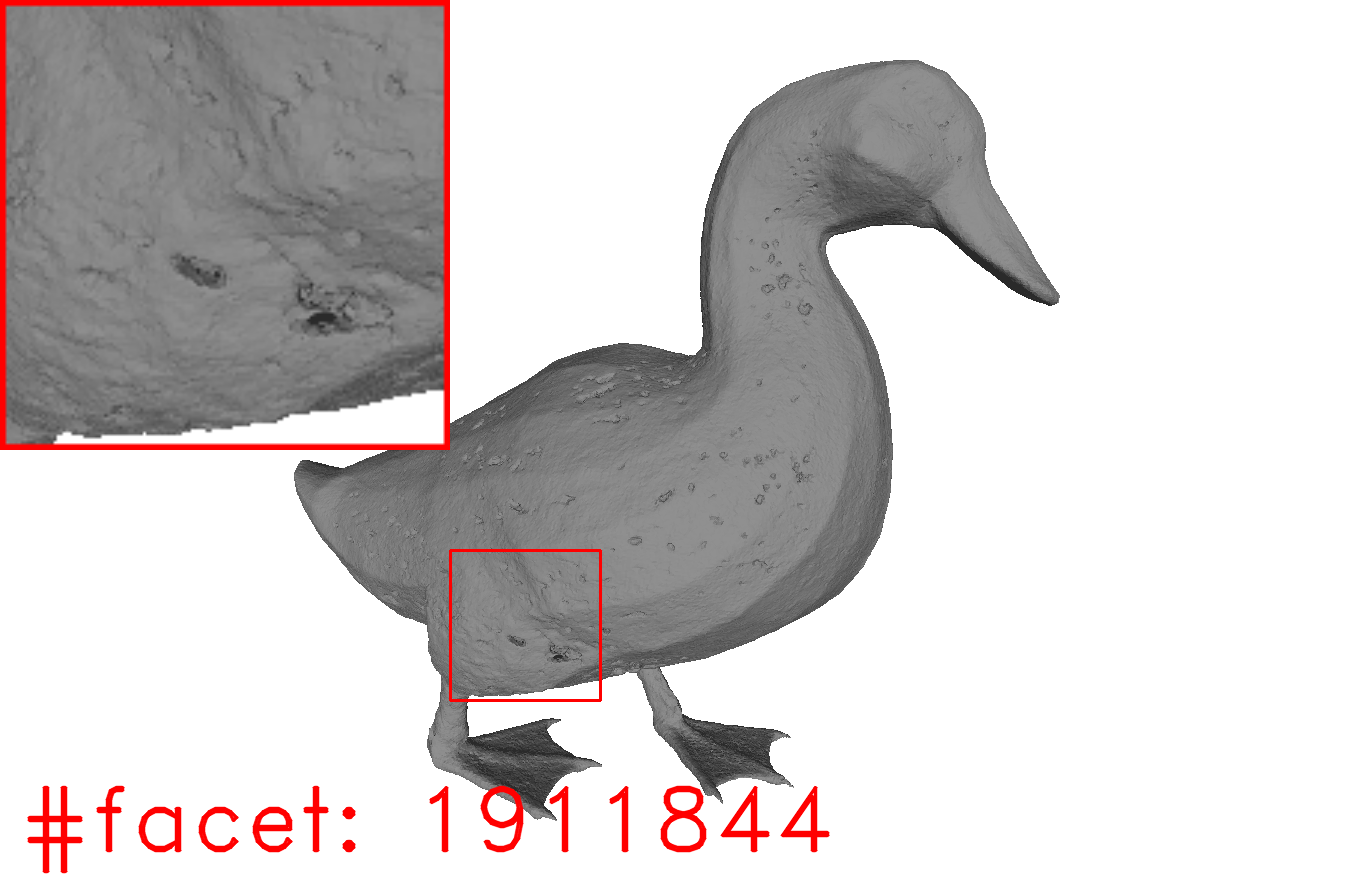}&
            \includegraphics[width=0.24\linewidth]{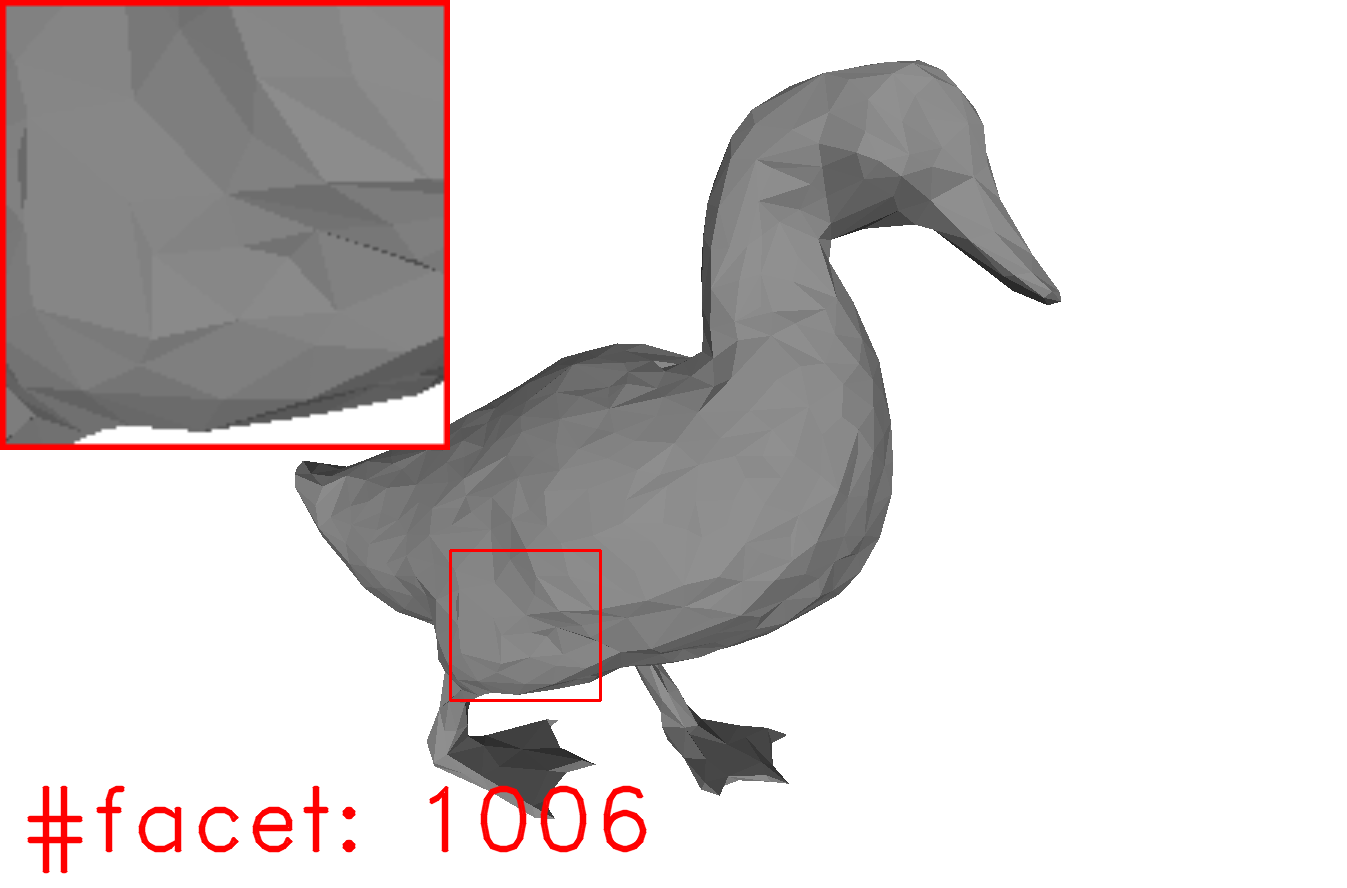}&
            \includegraphics[width=0.24\linewidth]{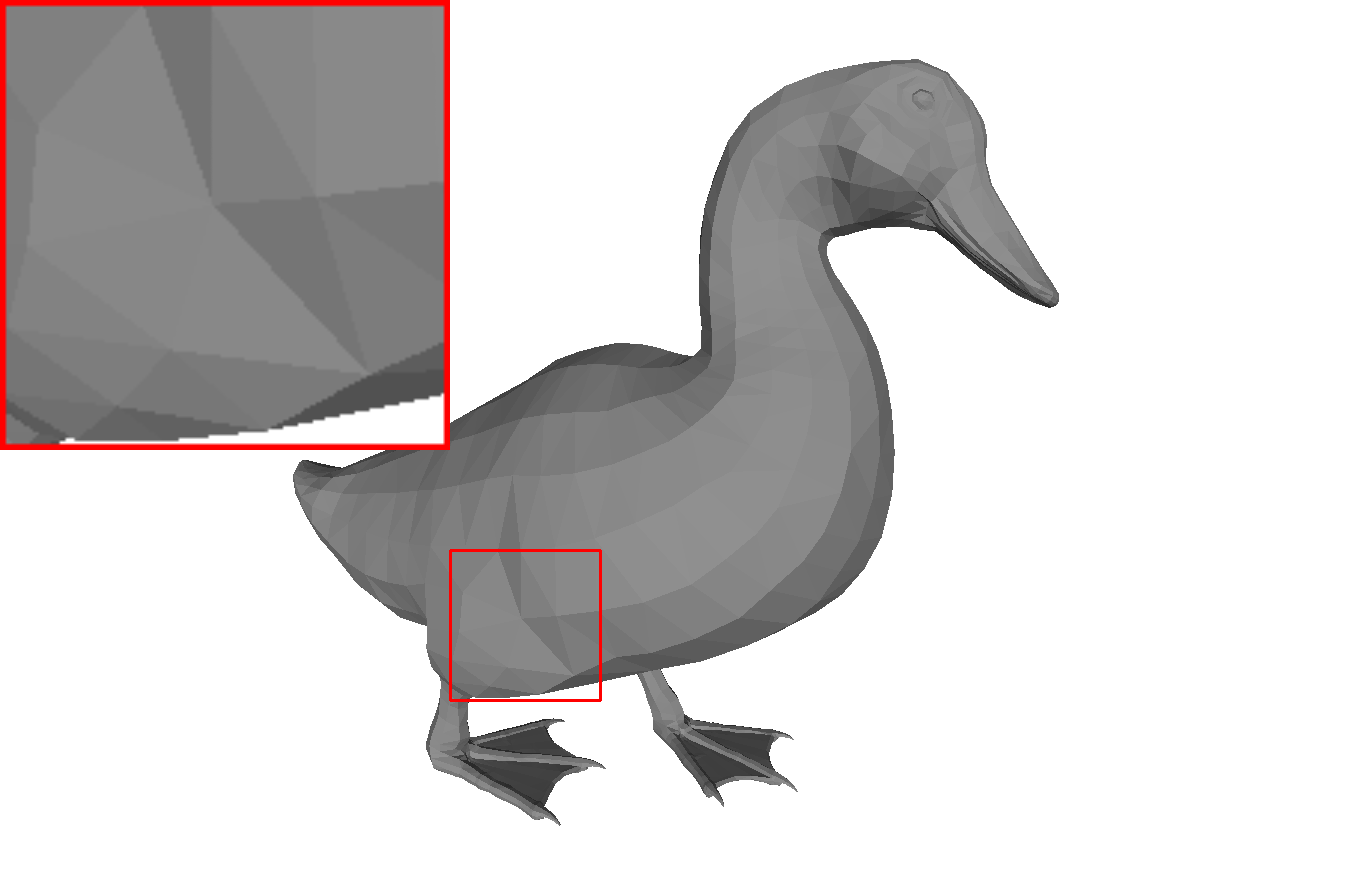}\\
            \raisebox{35pt}{\rotatebox[origin=c]{90}{Girlwalk}}&
            \includegraphics[width=0.24\linewidth]{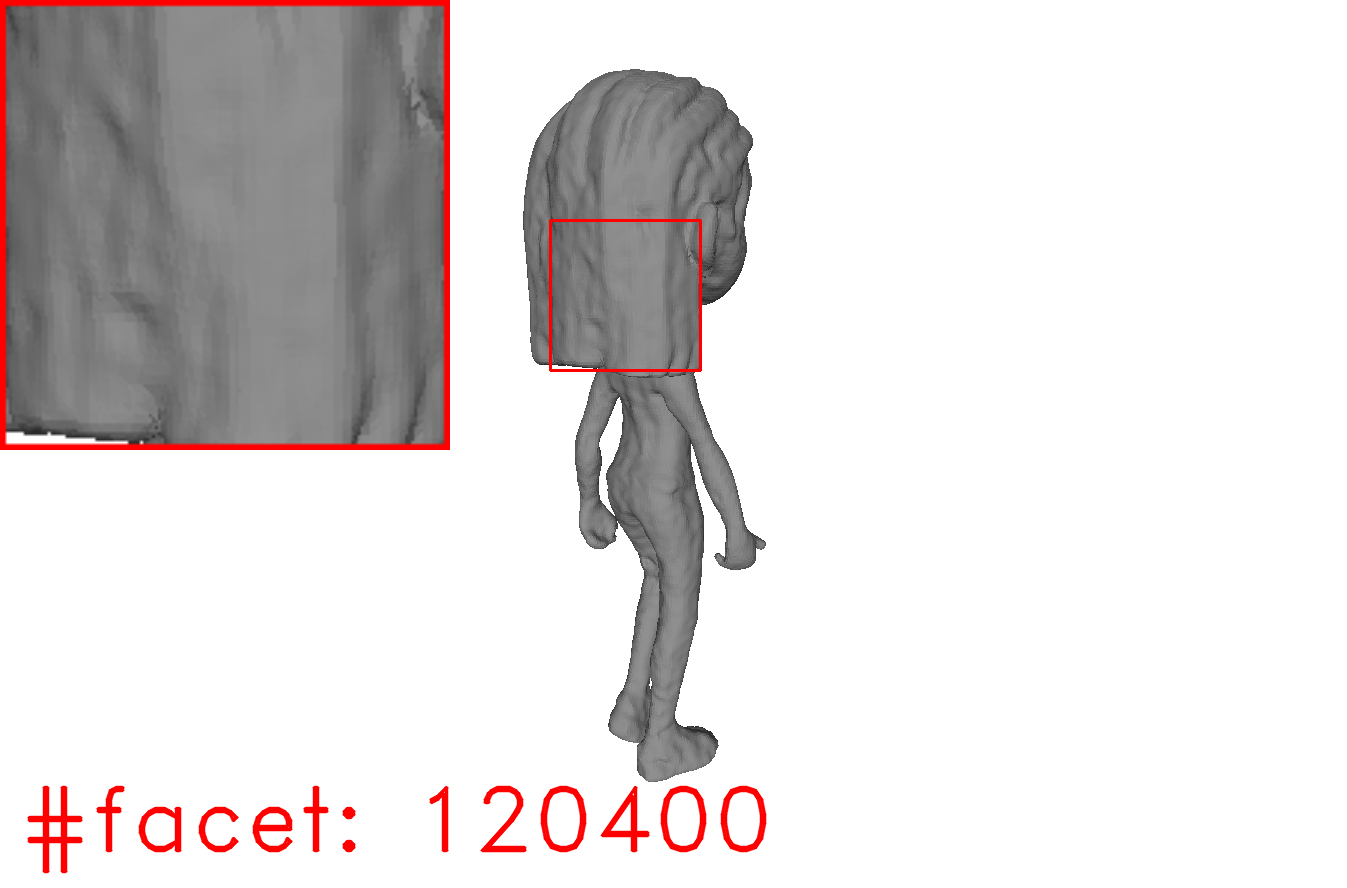}&
            \includegraphics[width=0.24\linewidth]{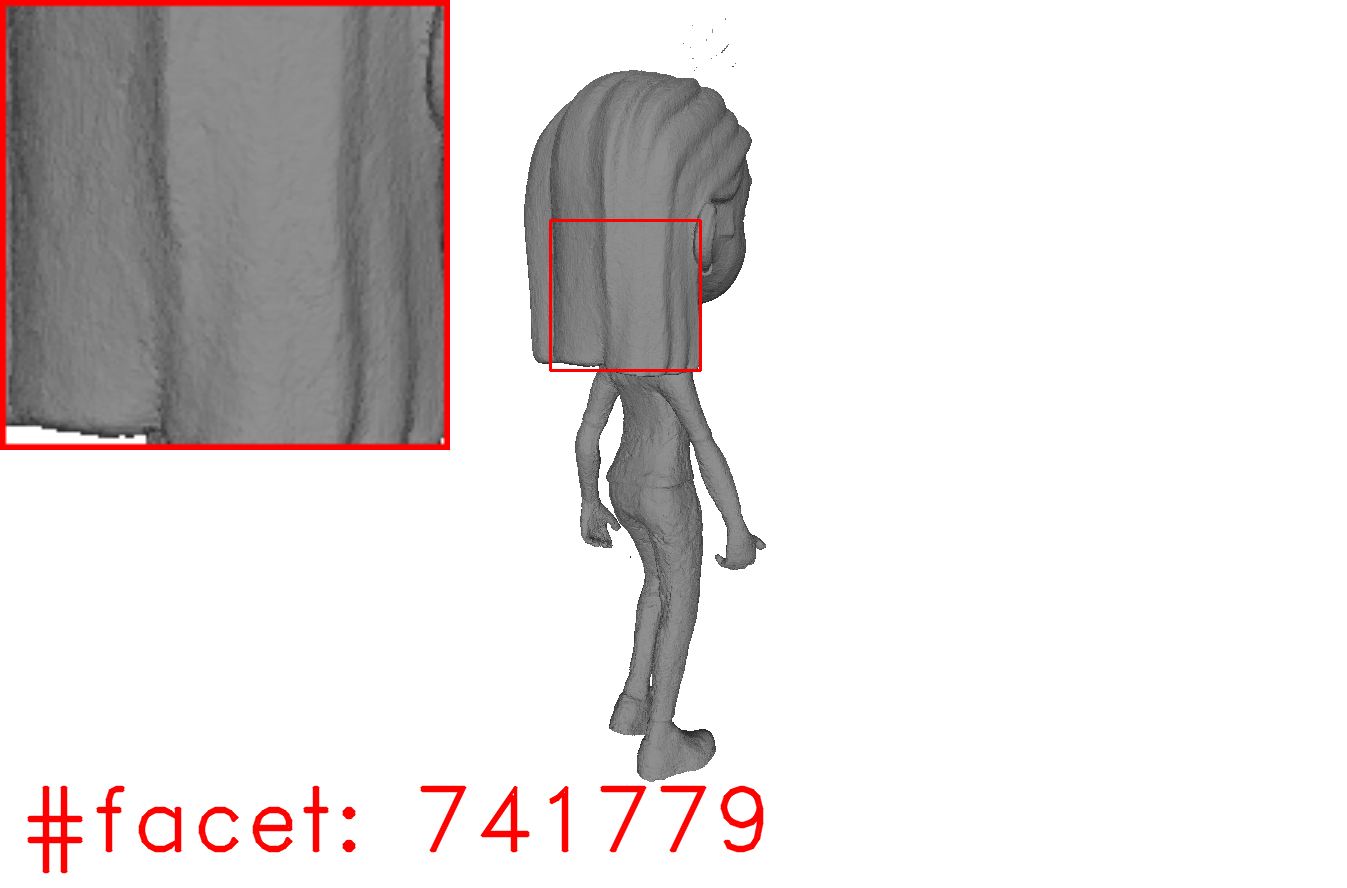}&
            \includegraphics[width=0.24\linewidth]{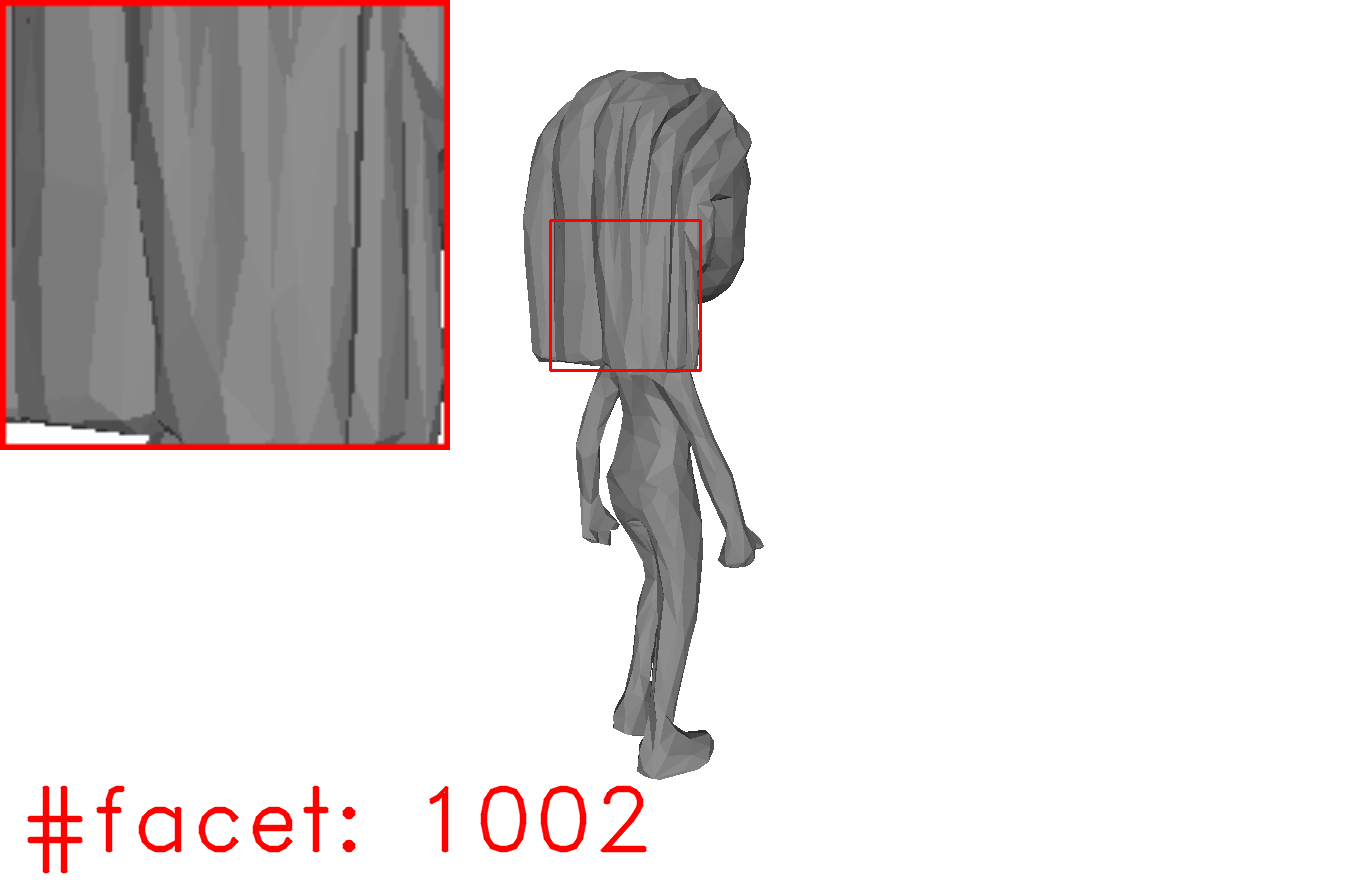}&
            \includegraphics[width=0.24\linewidth]{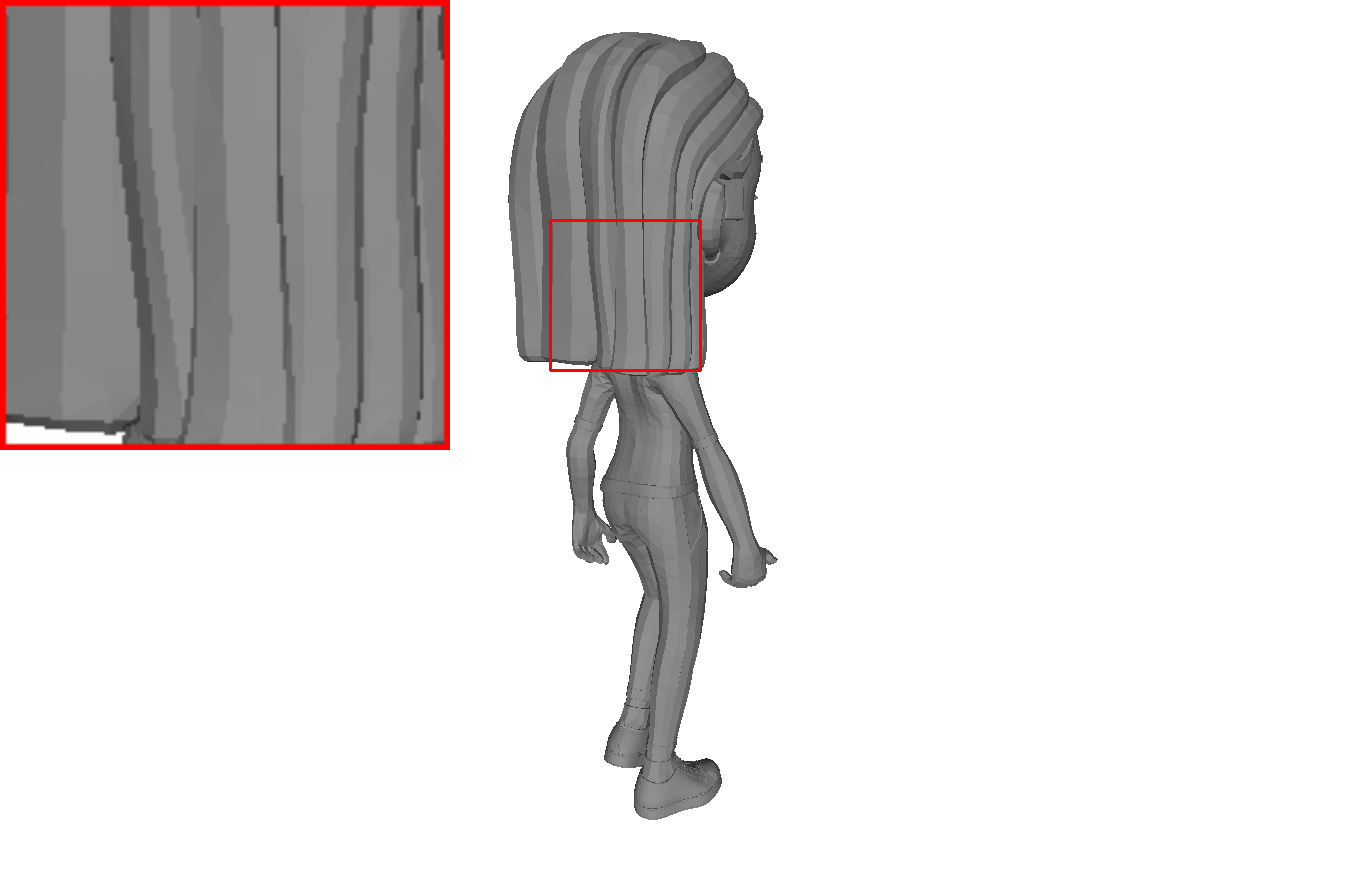}\\
            \raisebox{35pt}{\rotatebox[origin=c]{90}{Horse}}&
            \includegraphics[width=0.24\linewidth]{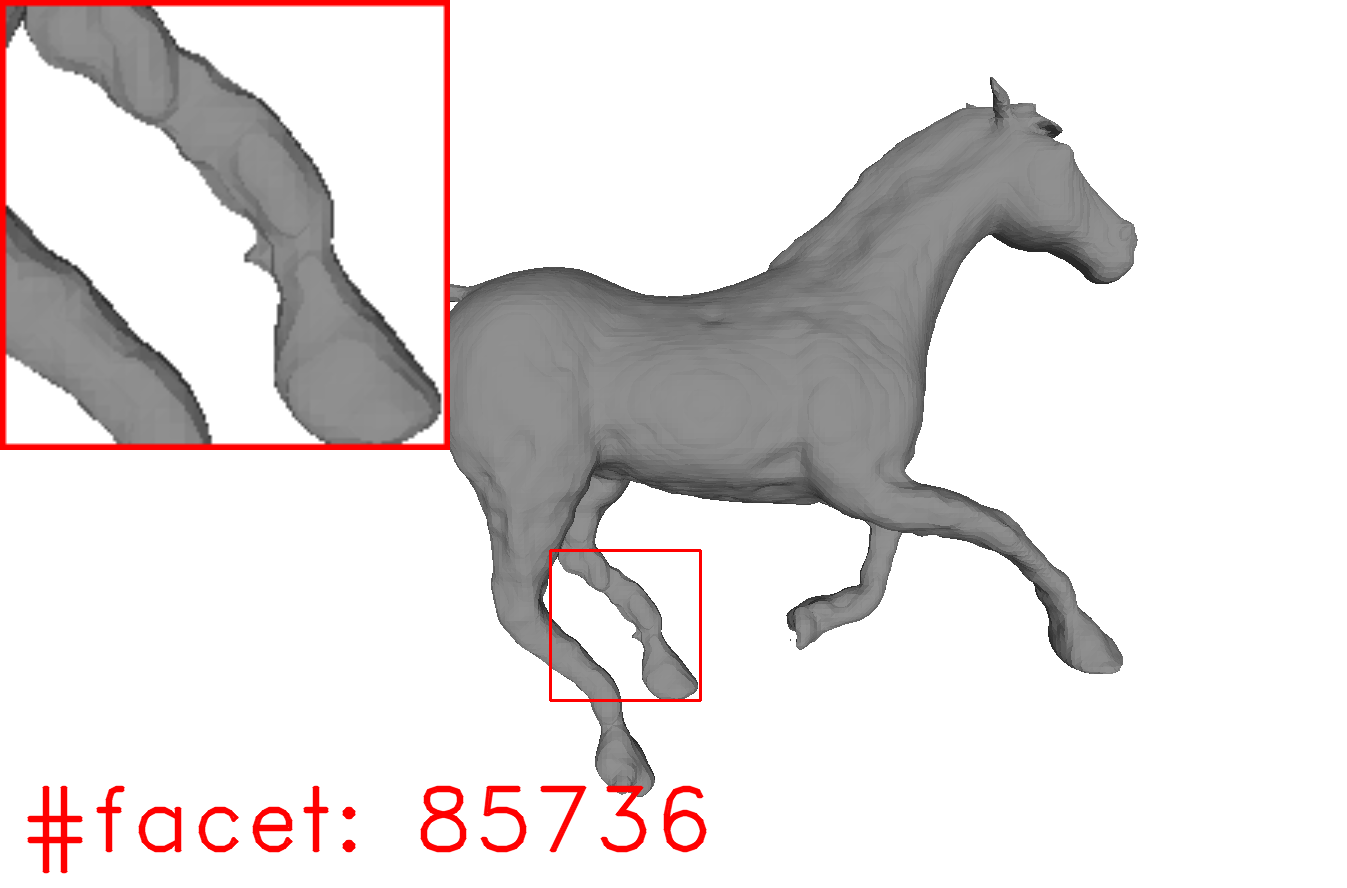}&
            \includegraphics[width=0.24\linewidth]{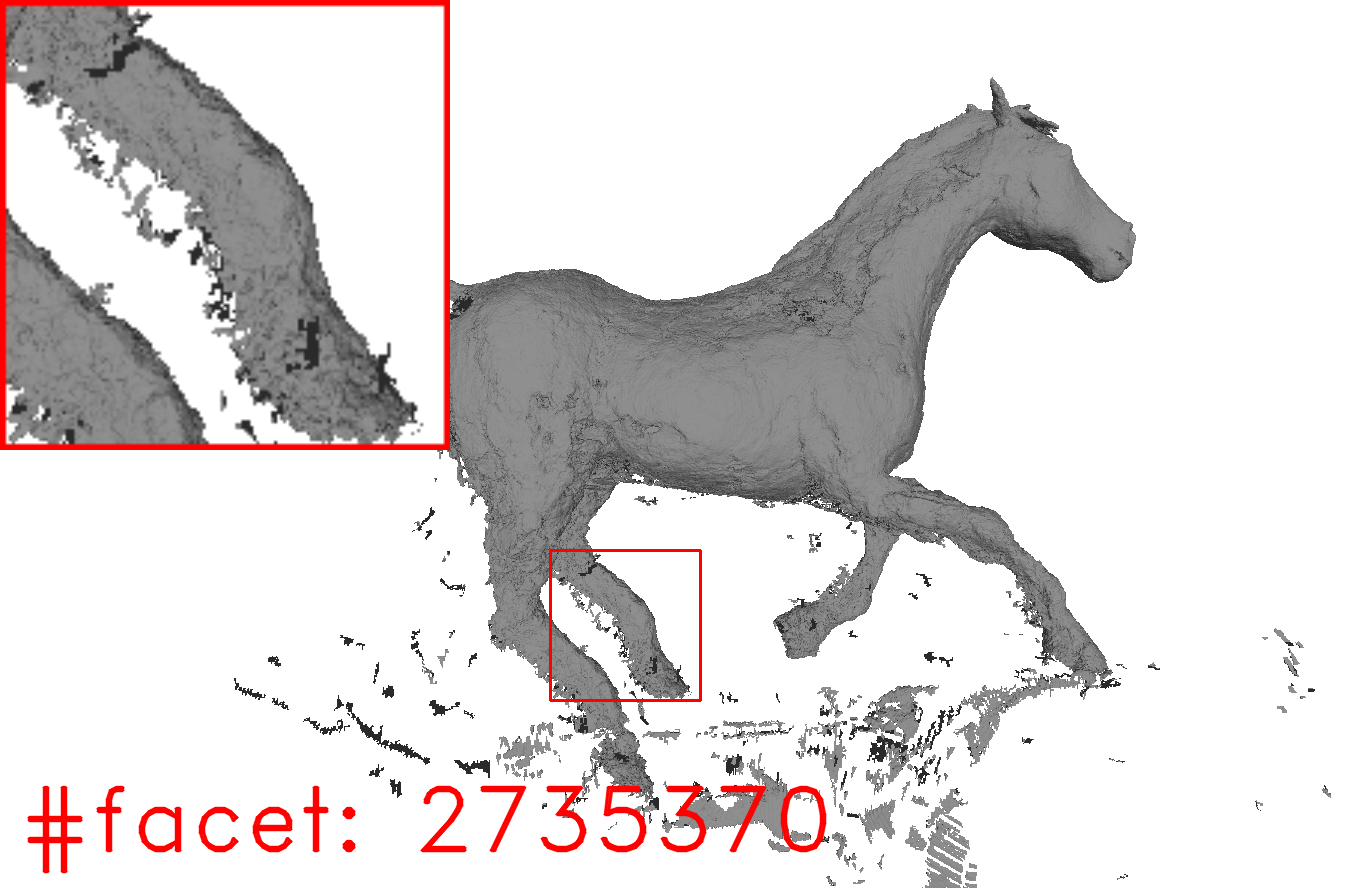}&
            \includegraphics[width=0.24\linewidth]{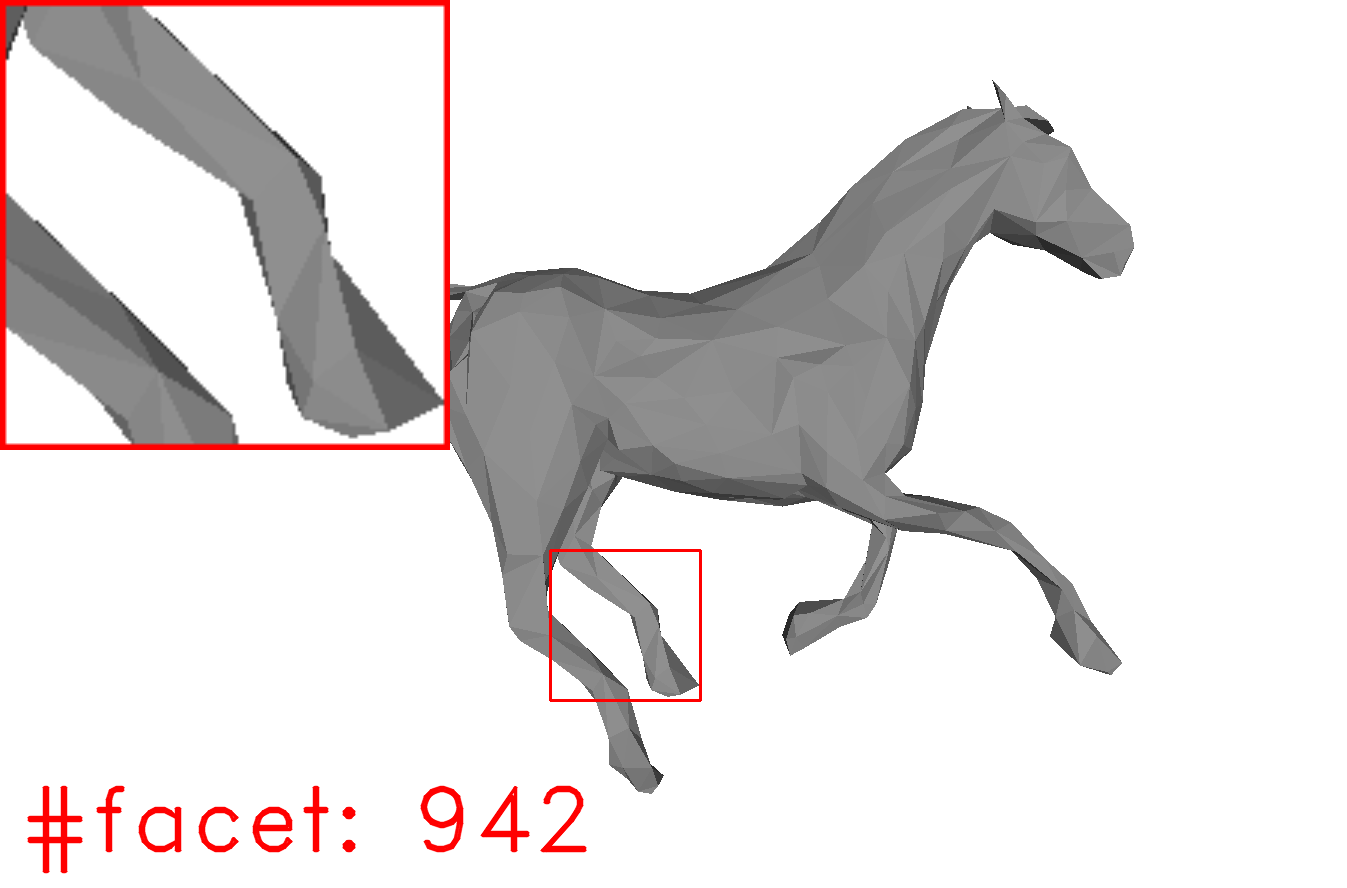}&
            \includegraphics[width=0.24\linewidth]{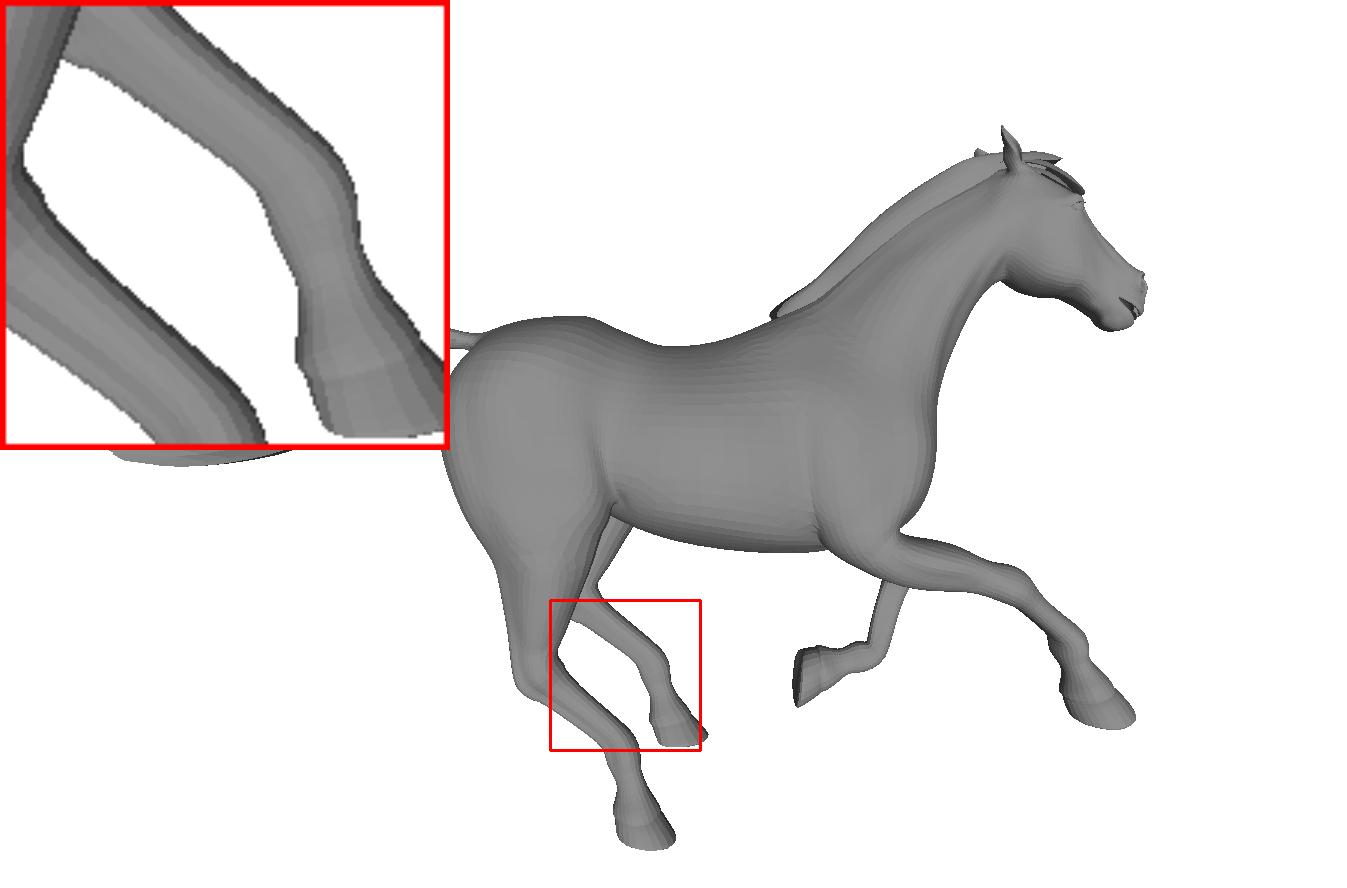}\\
        \end{tabular}
    }
	\caption{\textbf{Qualitative Comparison on the DG-Mesh Dataset.} MaGS achieves superior rendering and simulation results with better mesh quality, while using several orders of magnitude fewer facets, demonstrating its efficiency and effectiveness.}
        \label{fig:add_dgmesh_qualitative}
\end{figure*}

Figure~\ref{fig:add_dgmesh_qualitative} presents additional qualitative results showcasing the performance of our method on the DG-Mesh dataset. The comparison demonstrates that our approach achieves mesh reconstruction close to the ground truth, providing higher accuracy in the reconstructed meshes than previous methods with fewer facets.

\begin{figure*}[]
    \centering
    \addtolength{\tabcolsep}{-6.5pt}
    \footnotesize{
        \setlength{\tabcolsep}{1pt} %
        \begin{tabular}{p{8.2pt}cccccl}
            & 4D-GS & SC-GS & D-Miso & Grid4D & Ours &
        \\
        \raisebox{35pt}{\rotatebox[origin=c]{90}{bouncingballs}}&
        \includegraphics[height=2.6cm]{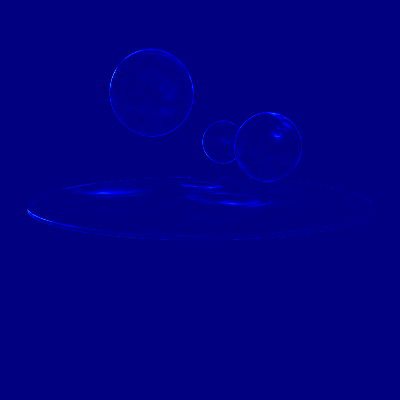} &
        \includegraphics[height=2.6cm]{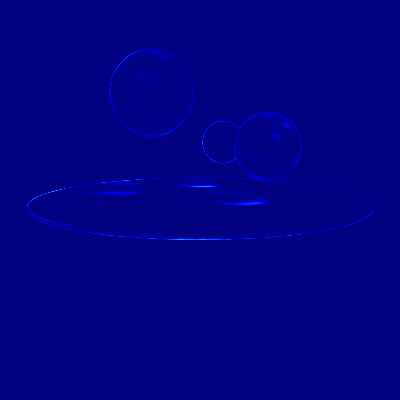} &
        \includegraphics[height=2.6cm]{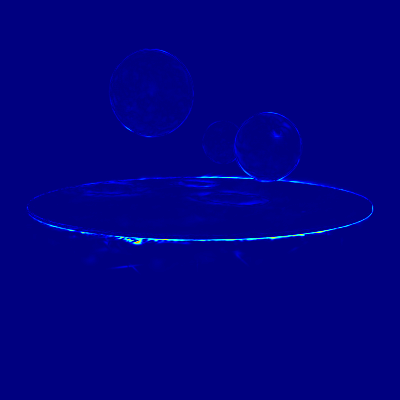} &
        \includegraphics[height=2.6cm]{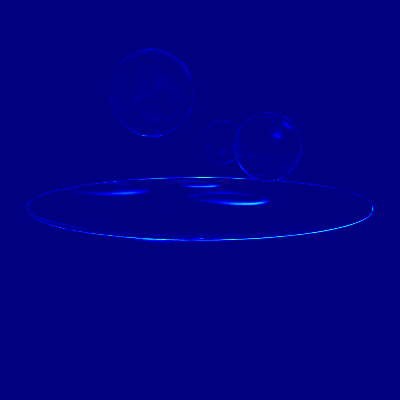} &
        \includegraphics[height=2.6cm]{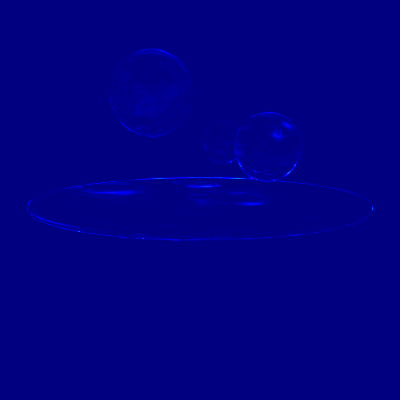} &
        \includegraphics[height=2.6cm]{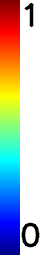}
        
        \\
        \raisebox{35pt}{\rotatebox[origin=c]{90}{hellwarrior}}&
        \includegraphics[height=2.6cm]{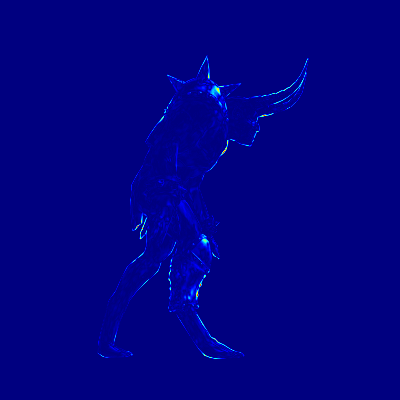} &
        \includegraphics[height=2.6cm]{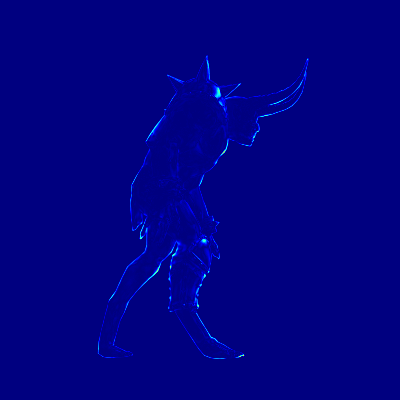} &
        \includegraphics[height=2.6cm]{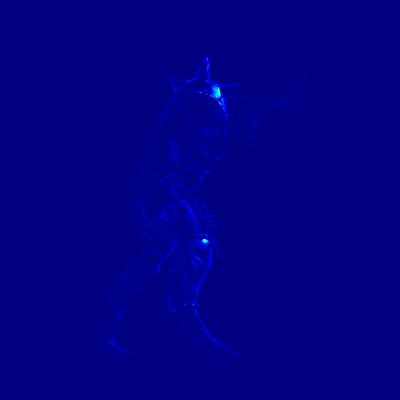} &
        \includegraphics[height=2.6cm]{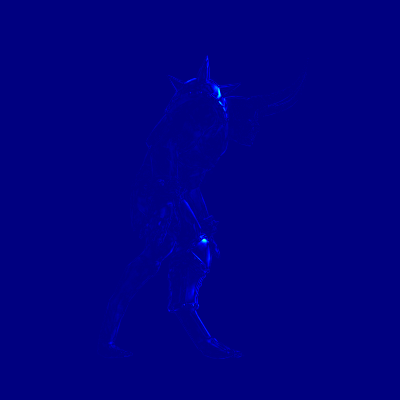} &
        \includegraphics[height=2.6cm]{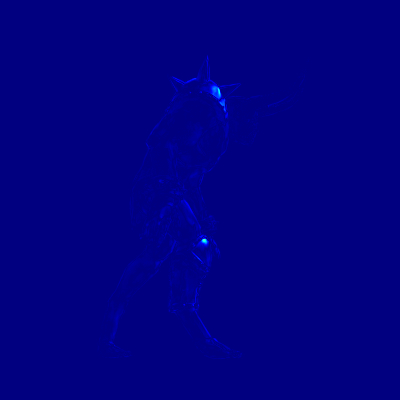} &
        \includegraphics[height=2.6cm]{images/heatmap/heatmap.png}
        
        \\
        \raisebox{35pt}{\rotatebox[origin=c]{90}{hook}}&
        \includegraphics[height=2.6cm]{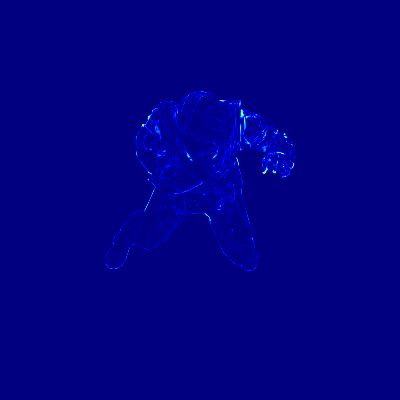} &
        \includegraphics[height=2.6cm]{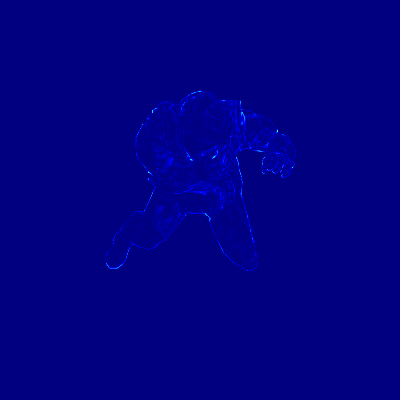} &
        \includegraphics[height=2.6cm]{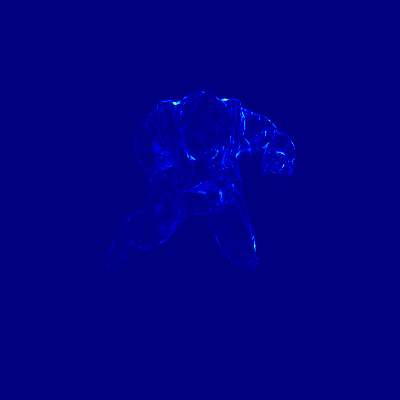} &
        \includegraphics[height=2.6cm]{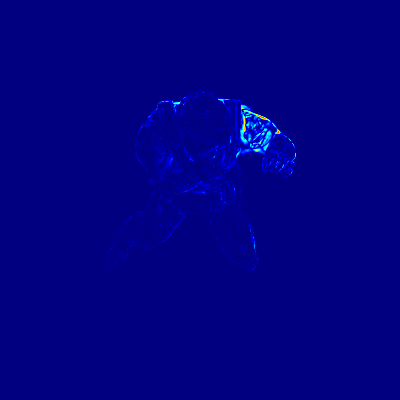} &
        \includegraphics[height=2.6cm]{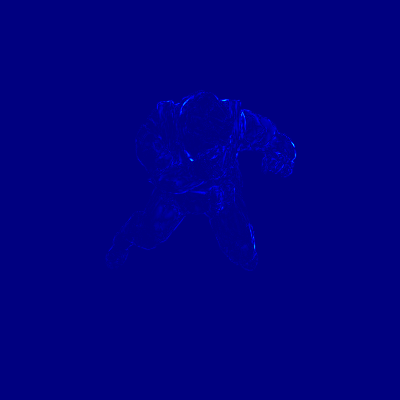} &
        \includegraphics[height=2.6cm]{images/heatmap/heatmap.png}
        
        \\
        \raisebox{35pt}{\rotatebox[origin=c]{90}{jumpingjacks}}&
        \includegraphics[height=2.6cm]{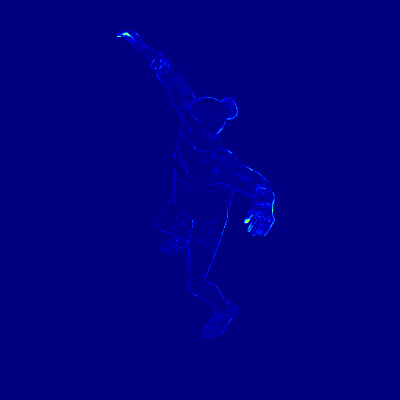} &
        \includegraphics[height=2.6cm]{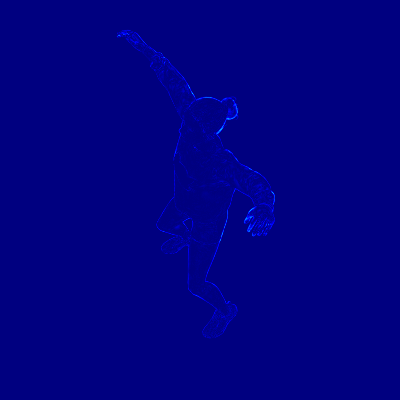} &
        \includegraphics[height=2.6cm]{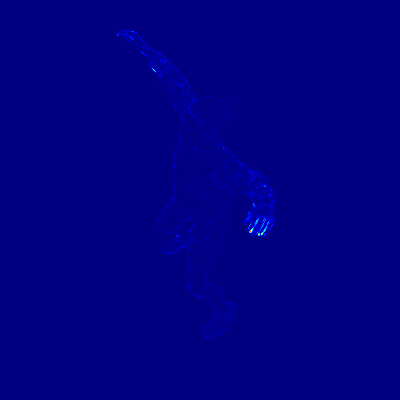} &
        \includegraphics[height=2.6cm]{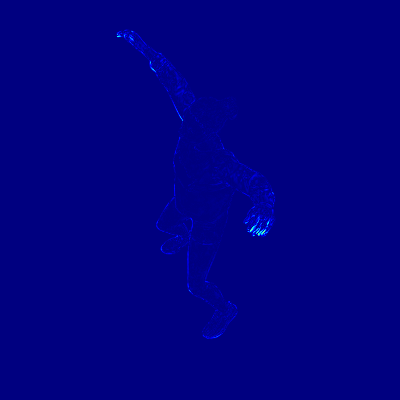} &
        \includegraphics[height=2.6cm]{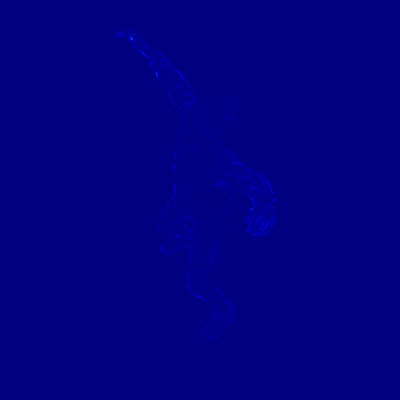} &
        \includegraphics[height=2.6cm]{images/heatmap/heatmap.png}
        
        \\
        \raisebox{35pt}{\rotatebox[origin=c]{90}{lego}}&
        \includegraphics[height=2.6cm]{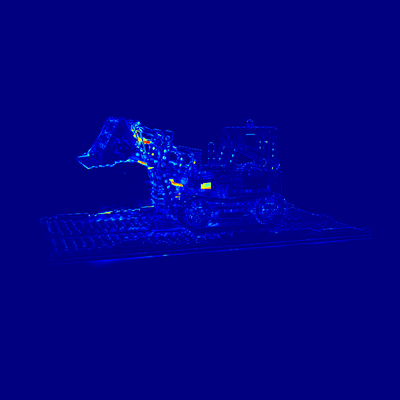} &
        \includegraphics[height=2.6cm]{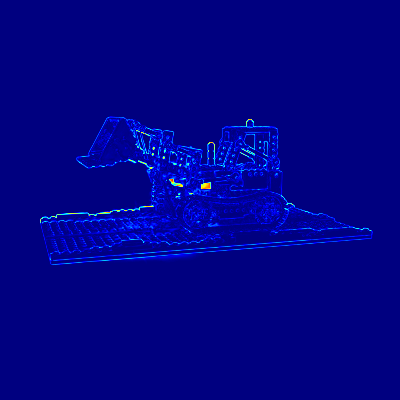} &
        \includegraphics[height=2.6cm]{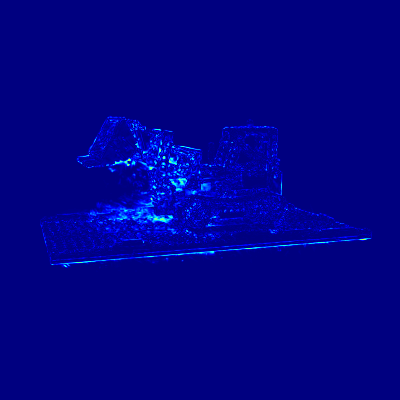} &
        \includegraphics[height=2.6cm]{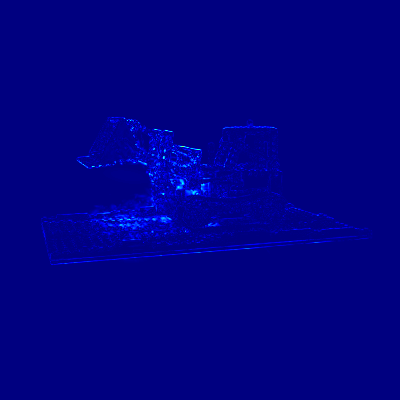} &
        \includegraphics[height=2.6cm]{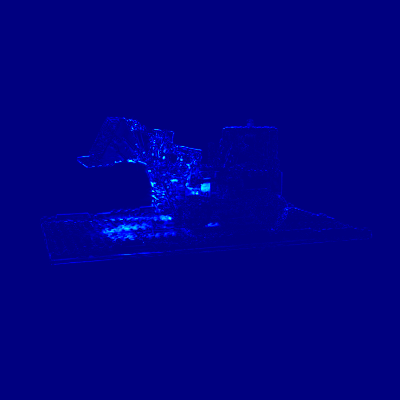} &
        \includegraphics[height=2.6cm]{images/heatmap/heatmap.png}
        
        \\
        \raisebox{35pt}{\rotatebox[origin=c]{90}{mutant}}&
        \includegraphics[height=2.6cm]{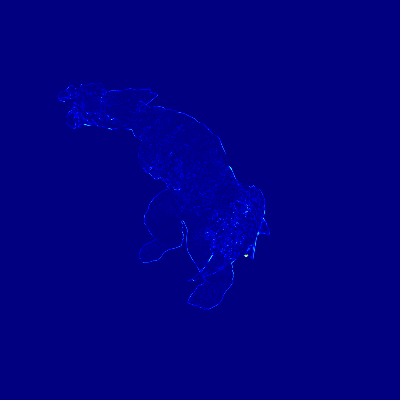} &
        \includegraphics[height=2.6cm]{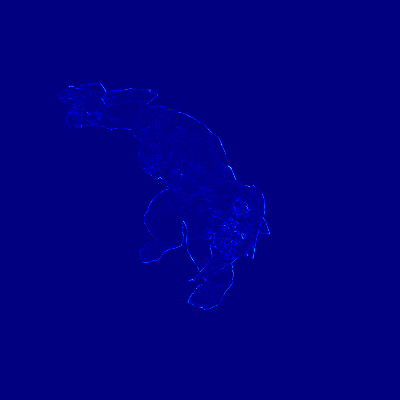} &
        \includegraphics[height=2.6cm]{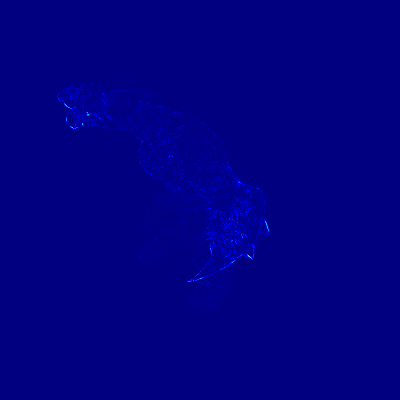} &
        \includegraphics[height=2.6cm]{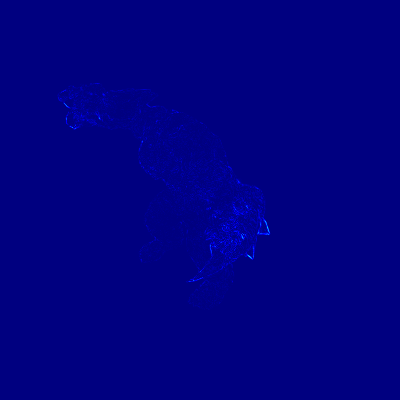} &
        \includegraphics[height=2.6cm]{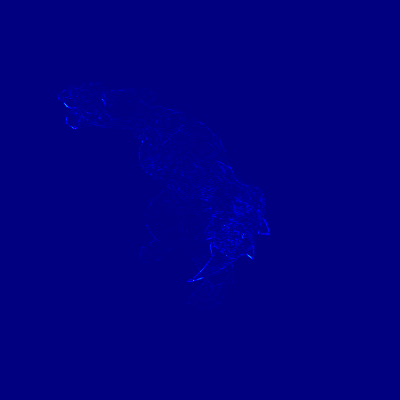} &
        \includegraphics[height=2.6cm]{images/heatmap/heatmap.png}
        
        \\
        \raisebox{35pt}{\rotatebox[origin=c]{90}{standup}}&
        \includegraphics[height=2.6cm]{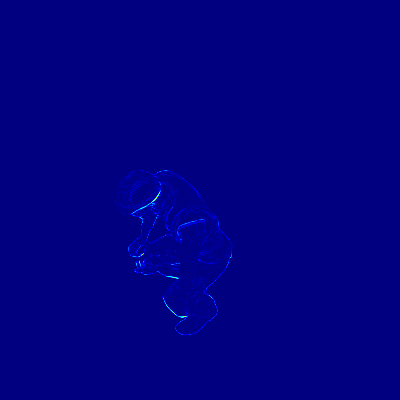} &
        \includegraphics[height=2.6cm]{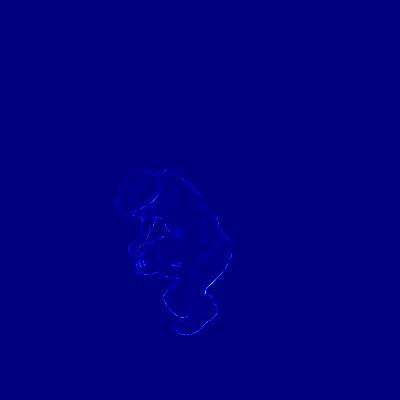} &
        \includegraphics[height=2.6cm]{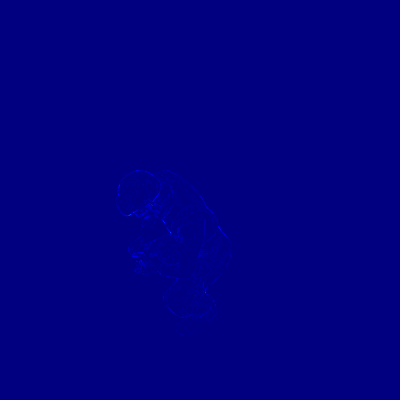} &
        \includegraphics[height=2.6cm]{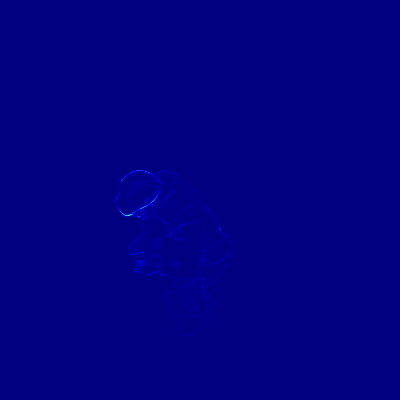} &
        \includegraphics[height=2.6cm]{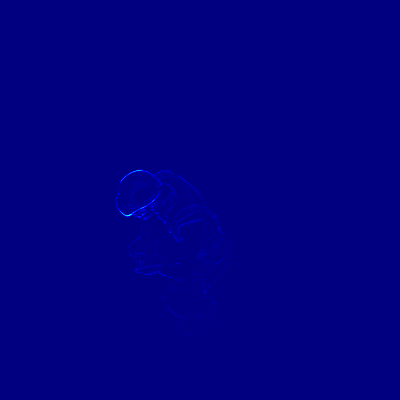} &
        \includegraphics[height=2.6cm]{images/heatmap/heatmap.png}
        
        \\
        \raisebox{35pt}{\rotatebox[origin=c]{90}{trex}}&
        \includegraphics[height=2.6cm]{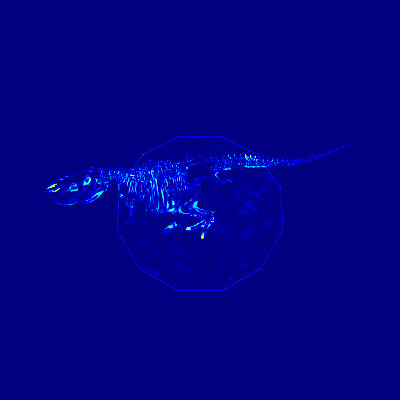} &
        \includegraphics[height=2.6cm]{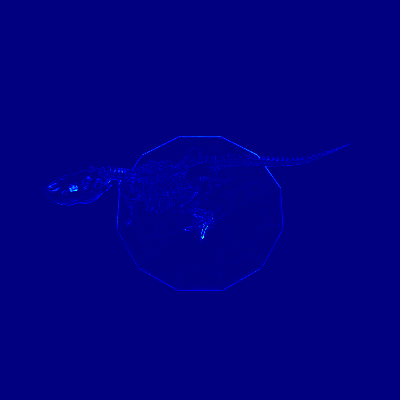} &
        \includegraphics[height=2.6cm]{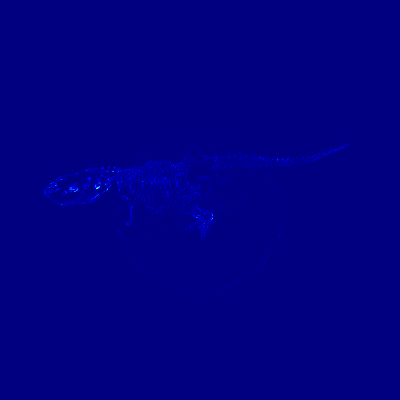} &
        \includegraphics[height=2.6cm]{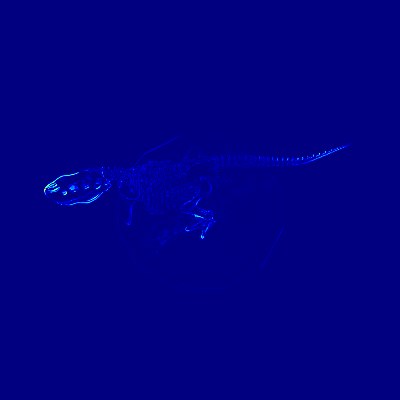} &
        \includegraphics[height=2.6cm]{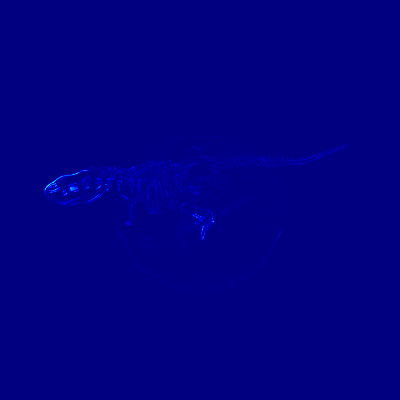} &
        \includegraphics[height=2.6cm]{images/heatmap/heatmap.png}

        \end{tabular}
    }
\caption{\textbf{Visualization of L1 Loss (between rendered and GT image) Qualitative Results on D-NeRF.} We compare MaGS with 4D-GS, SC-GS, D-Miso, and Grid4D. All images are processed using the same pseudo-color conversion algorithm (CV2's COLORMAP-JET).}
\label{fig:add_dnerf_heatmap}
\vspace{-10pt}
\end{figure*}

Figure~\ref{fig:add_dnerf_heatmap} presents an L1 loss visualization on the D-NeRF dataset, comparing the deformation predictions of our model with those of existing methods. This detailed qualitative comparison demonstrates that our approach achieves the most accurate rendering, highlighting its superior performance. 

\begin{figure*}[]
    \centering
    \addtolength{\tabcolsep}{-6.5pt}
    \footnotesize{
        \setlength{\tabcolsep}{1pt} %
        \begin{tabular}{p{8.2pt}cccl}
            & 3DGS-Avatar & SplattingAvatar & Ours &
        \\
        \raisebox{35pt}{\rotatebox[origin=c]{90}{female-3-casual}}&
        \includegraphics[height=3cm]{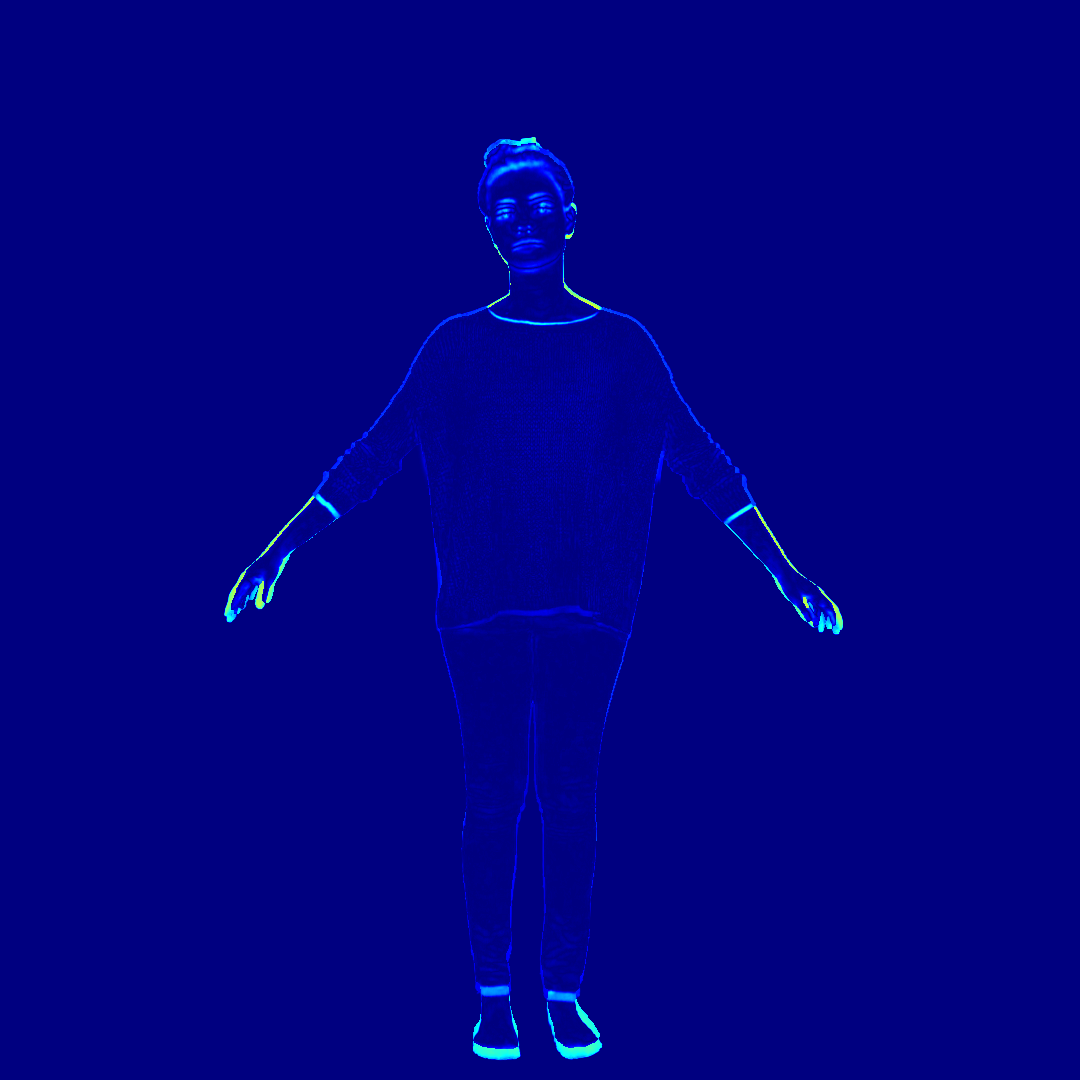} &
        \includegraphics[height=3cm]{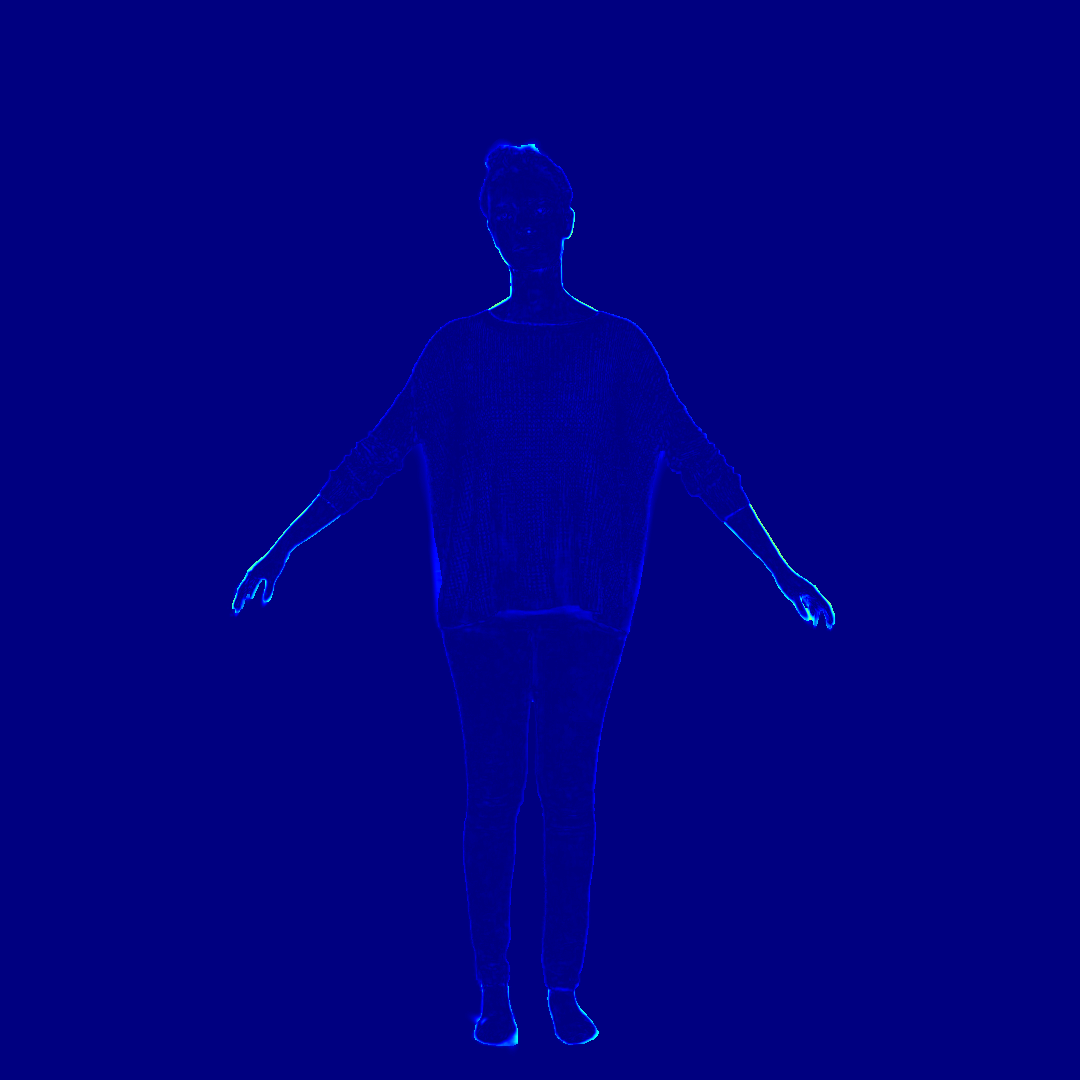} &
        \includegraphics[height=3cm]{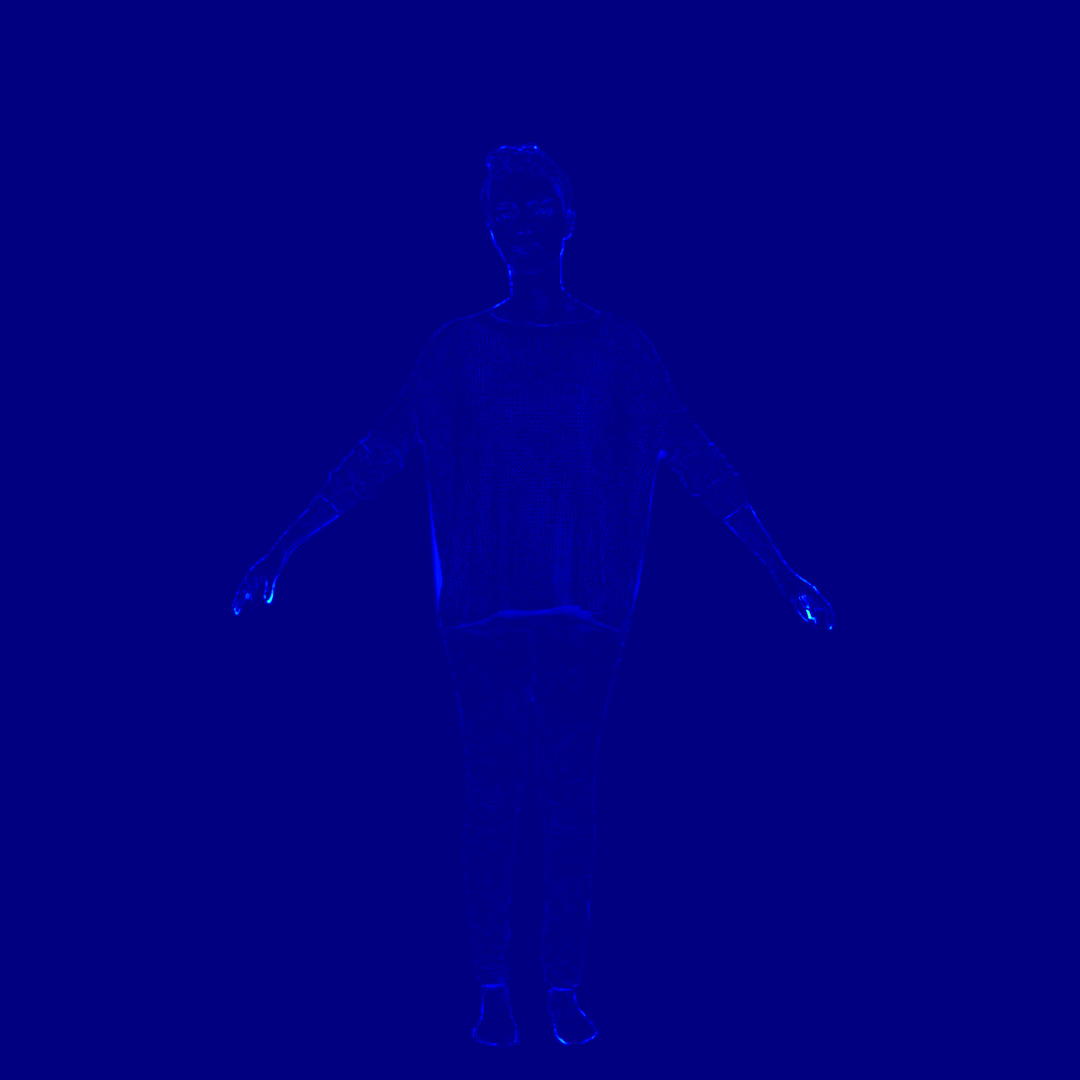} &
        \includegraphics[height=3cm]{images/heatmap/heatmap.png} 
        
        \\
        \raisebox{35pt}{\rotatebox[origin=c]{90}{female-4-casual}}&
        \includegraphics[height=3cm]{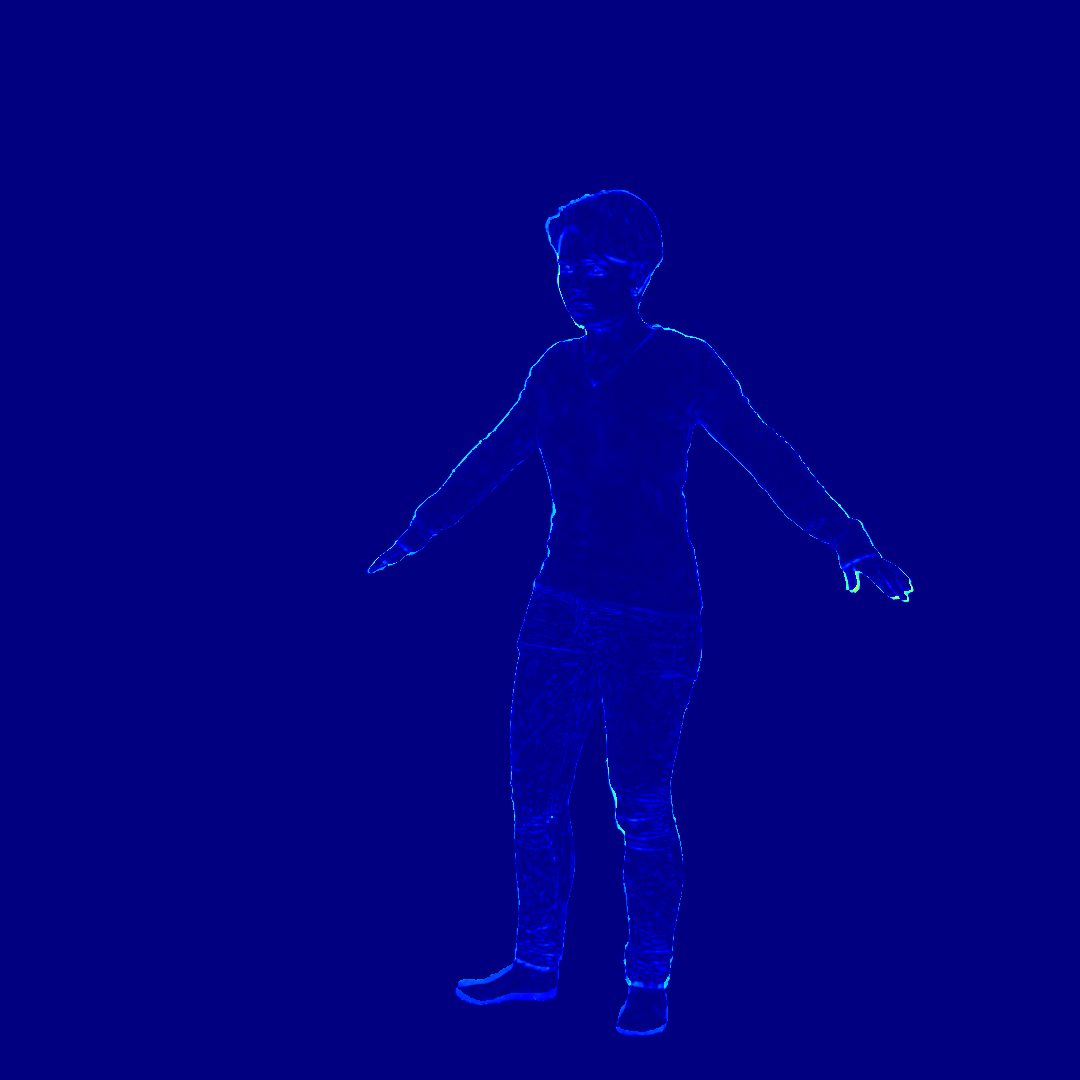} &
        \includegraphics[height=3cm]{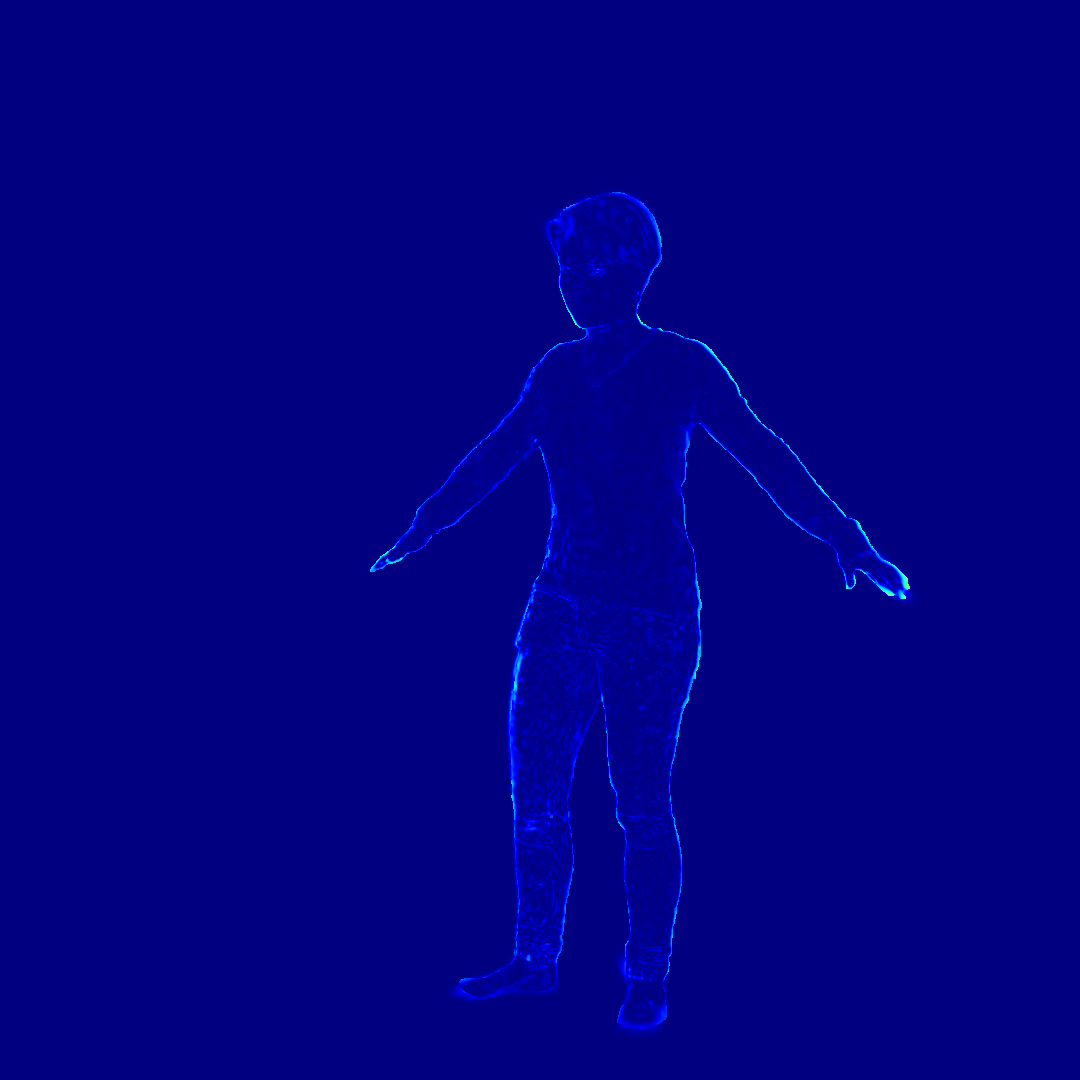} &
        \includegraphics[height=3cm]{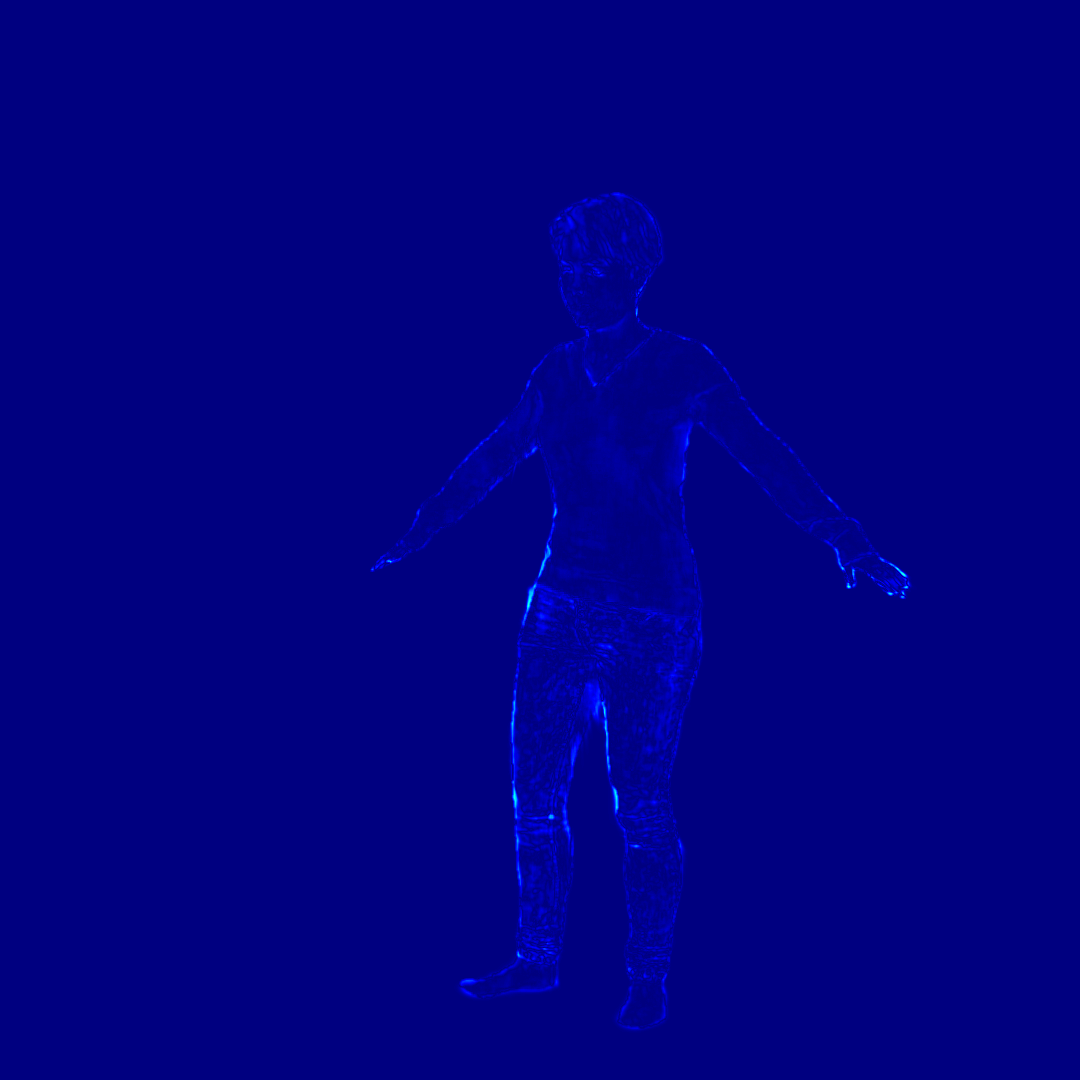} &
        \includegraphics[height=3cm]{images/heatmap/heatmap.png} 
        
        \\
        \raisebox{35pt}{\rotatebox[origin=c]{90}{male-3-casual}}&
        \includegraphics[height=3cm]{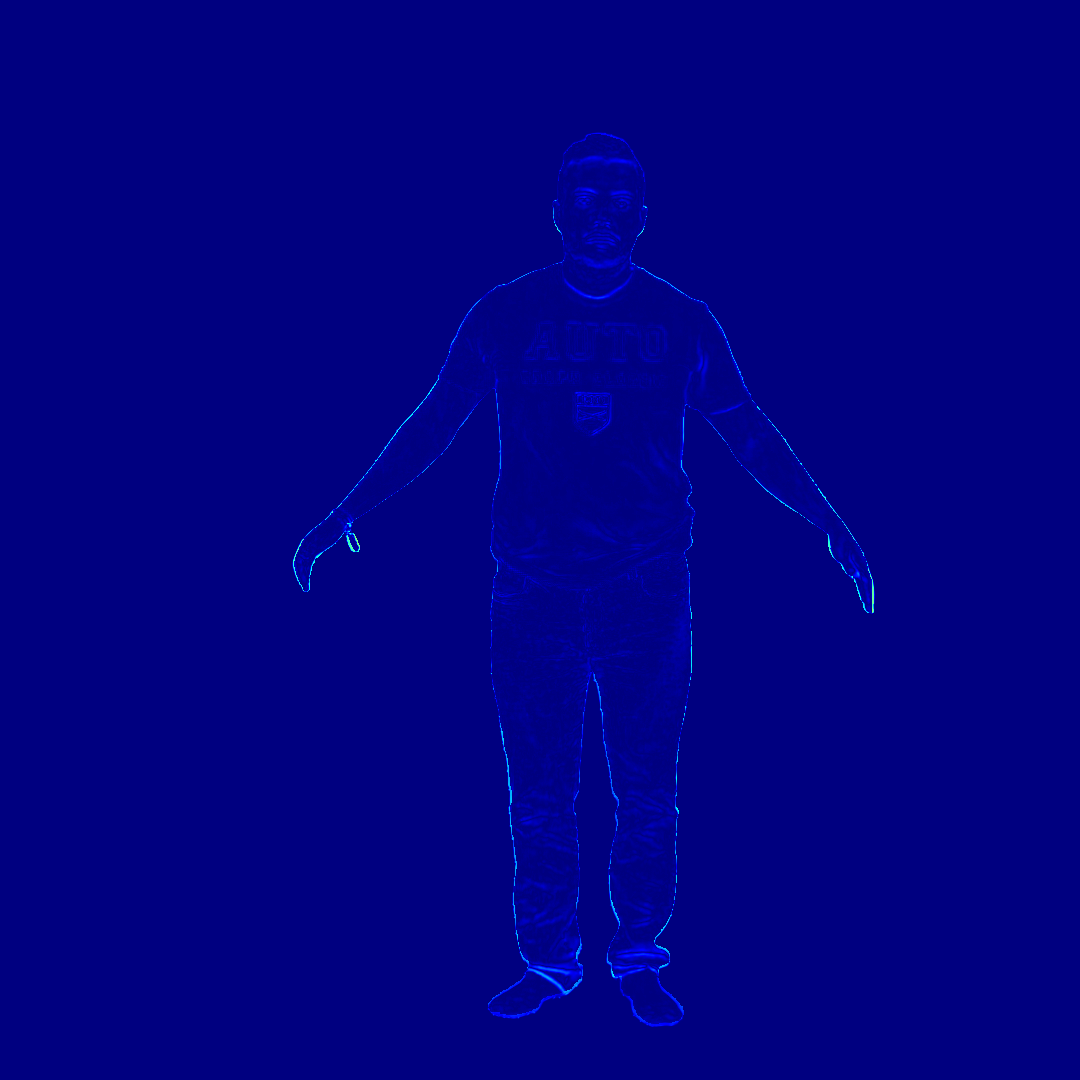} &
        \includegraphics[height=3cm]{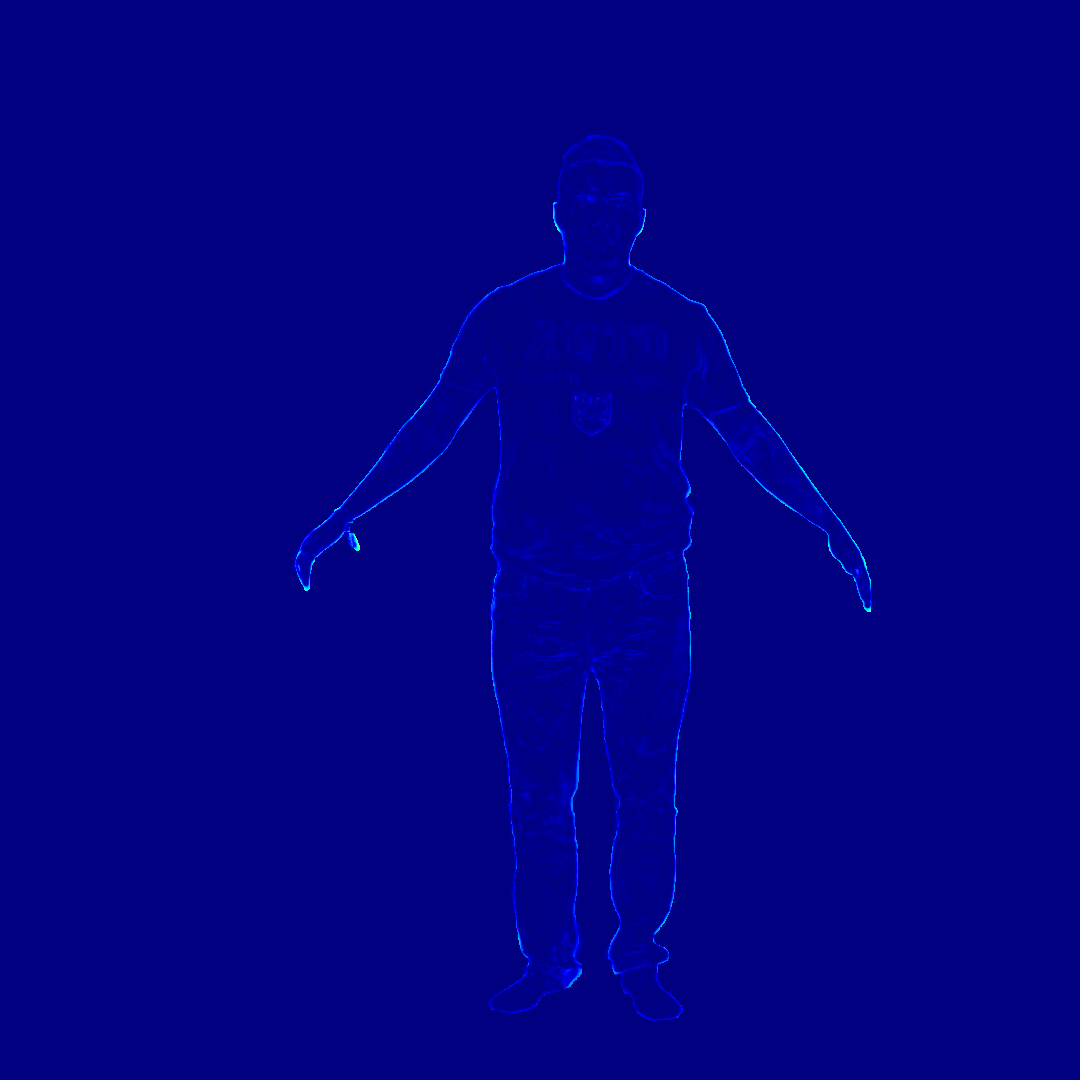} &
        \includegraphics[height=3cm]{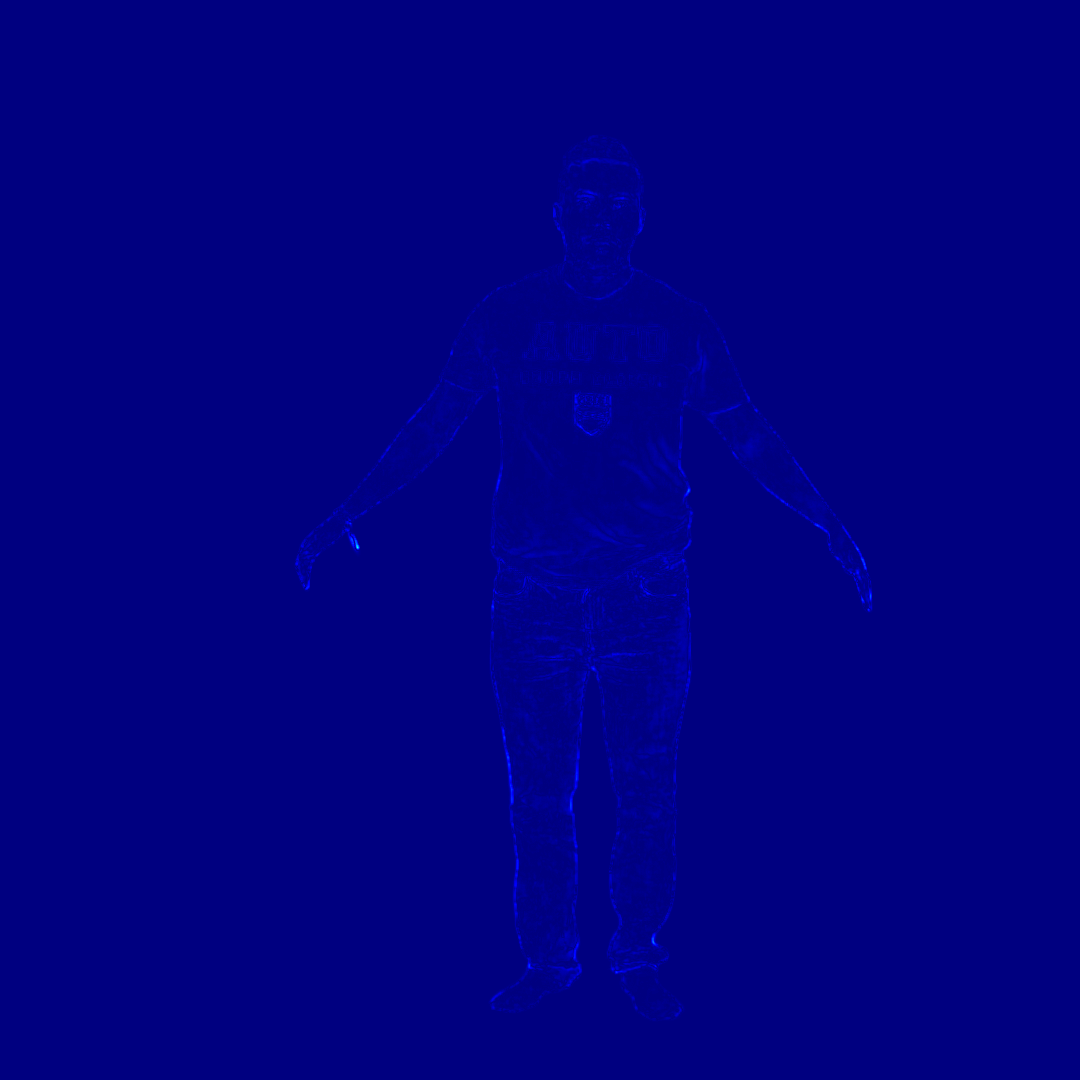} &
        \includegraphics[height=3cm]{images/heatmap/heatmap.png} 
        
        \\
        \raisebox{35pt}{\rotatebox[origin=c]{90}{male-4-casual}}&
        \includegraphics[height=3cm]{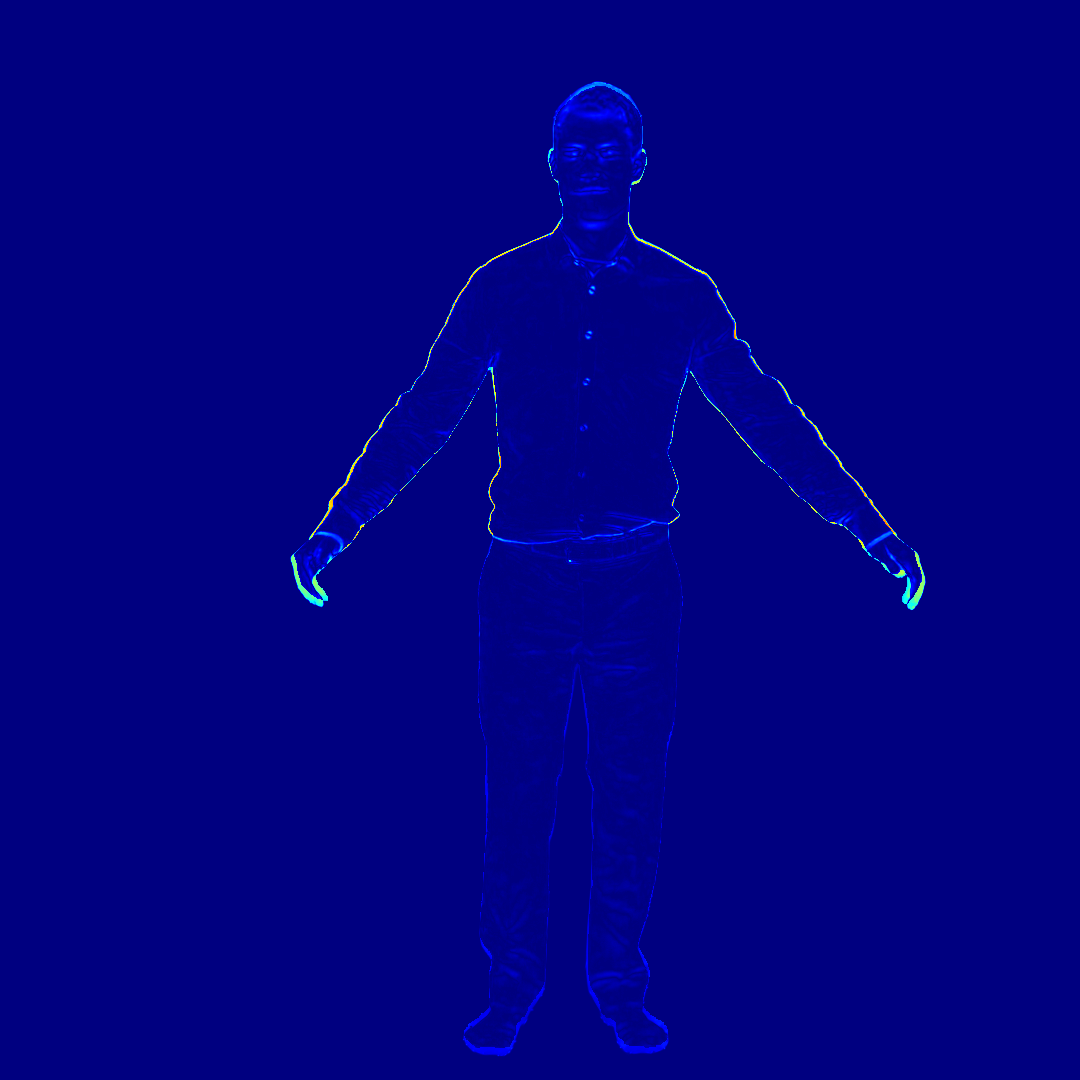} &
        \includegraphics[height=3cm]{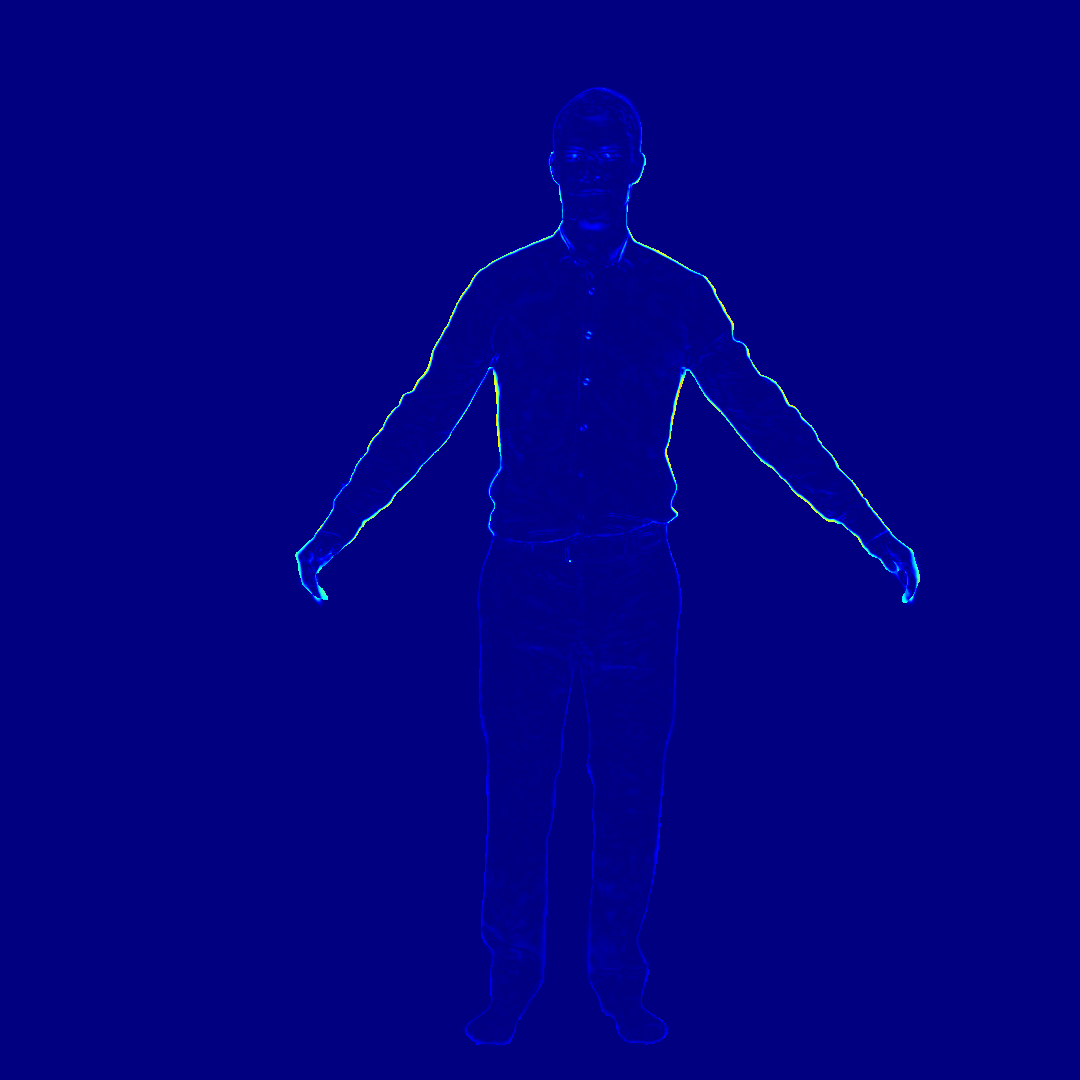} &
        \includegraphics[height=3cm]{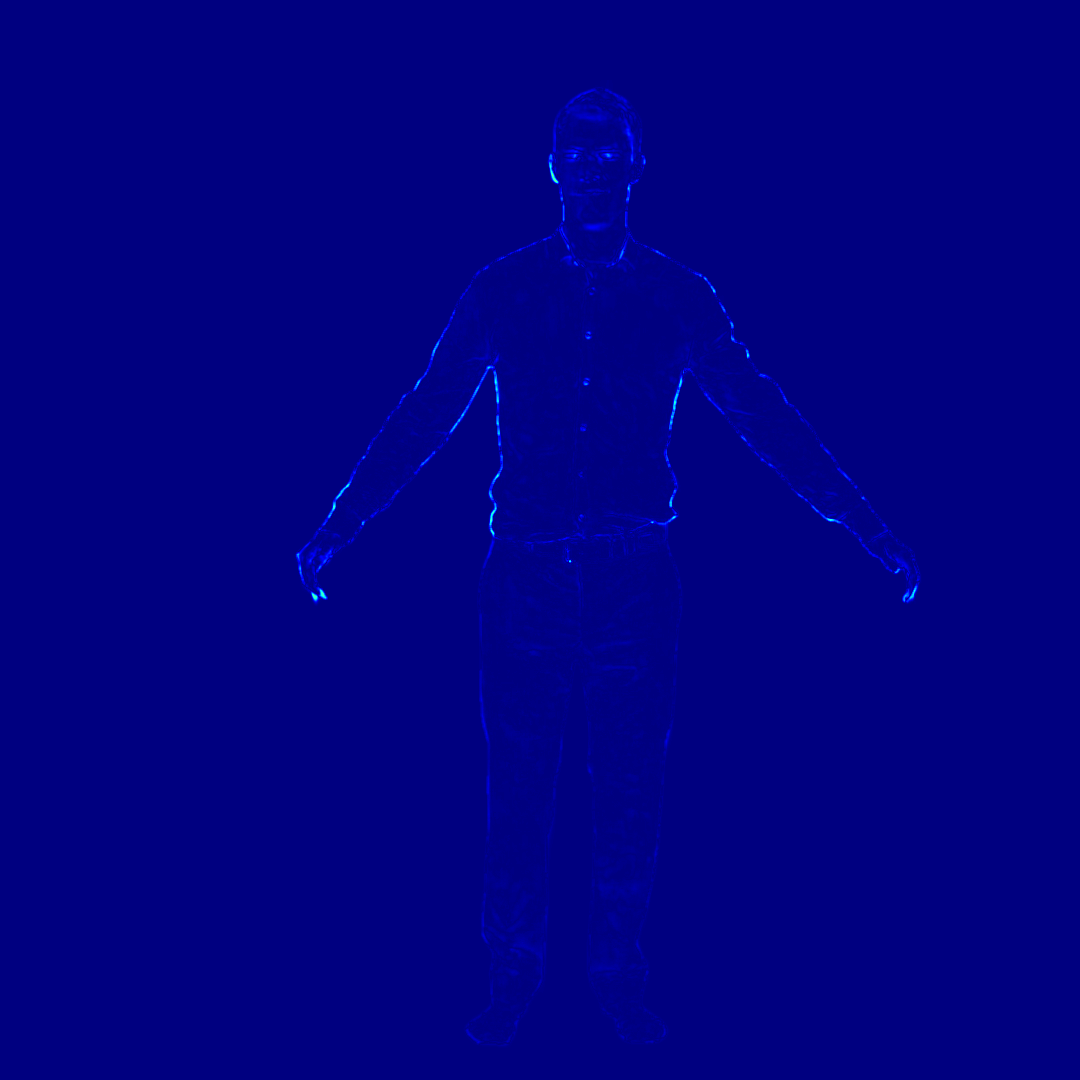} &
        \includegraphics[height=3cm]{images/heatmap/heatmap.png}

        \end{tabular}
    }
\caption{\textbf{Visualization of L1 Loss (between rendered and GT image) Qualitative Results on PeopleSnapshot.} We compare MaGS with 3DGS-Avatar and SplattingAvatar. All images are processed using the same pseudo-color conversion algorithm (CV2's COLORMAP-JET).}
\label{fig:add_peoplesnapshot_heatmap}
\end{figure*}

Similarly, Figure~\ref{fig:add_peoplesnapshot_heatmap} illustrates an L1 loss visualization of the results on the PeopleSnapshot dataset, providing insights into rendering accuracy in real-world scenarios. The visualization confirms that our method is highly accurate, further validating its suitability for human pose tasks.

\section{Additional Visualizations of Simulation}
\label{sec:simulations}
\begin{figure*}
    \centering
    \addtolength{\tabcolsep}{-6.5pt}
    \footnotesize{
        \setlength{\tabcolsep}{1pt} %
        \begin{tabular}{cccc}
            View1 Mesh & View1 MaGS & View2 Mesh & View2 MaGS\\
            \includegraphics[width=0.24\linewidth]{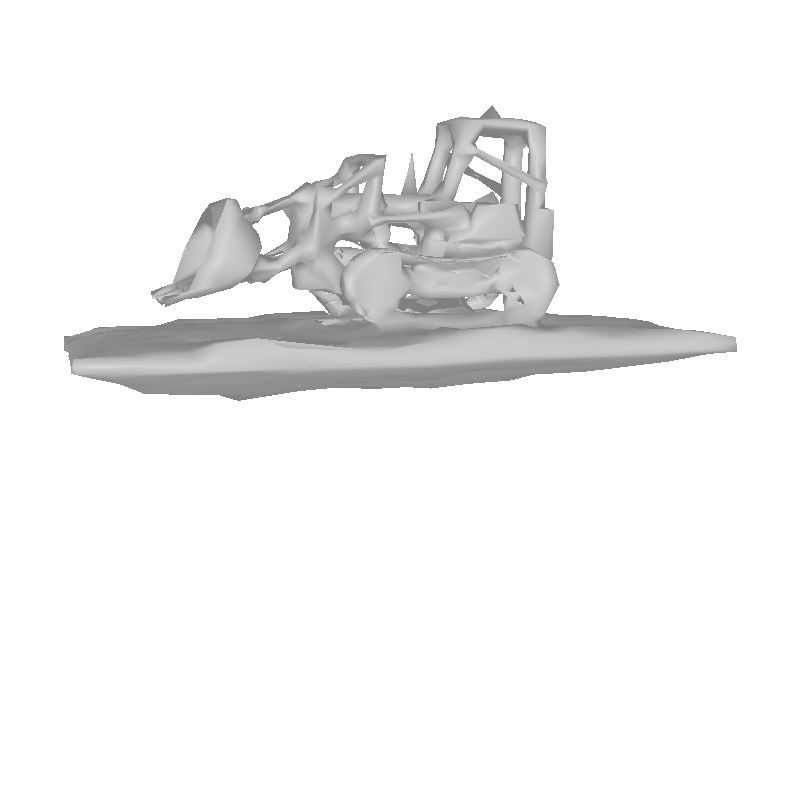}&
            \includegraphics[width=0.24\linewidth]{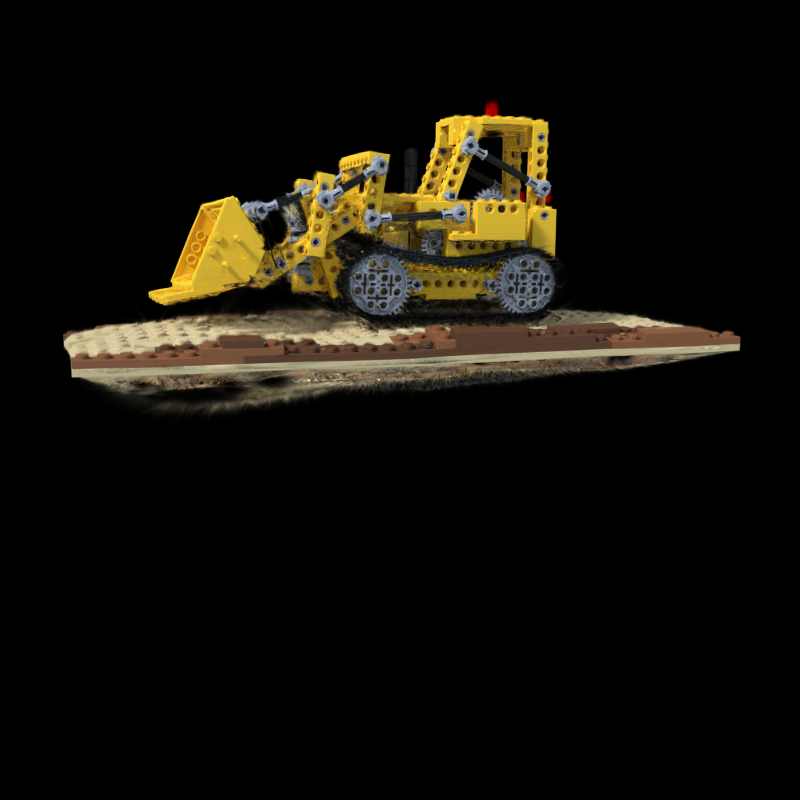}&
            \includegraphics[width=0.24\linewidth]{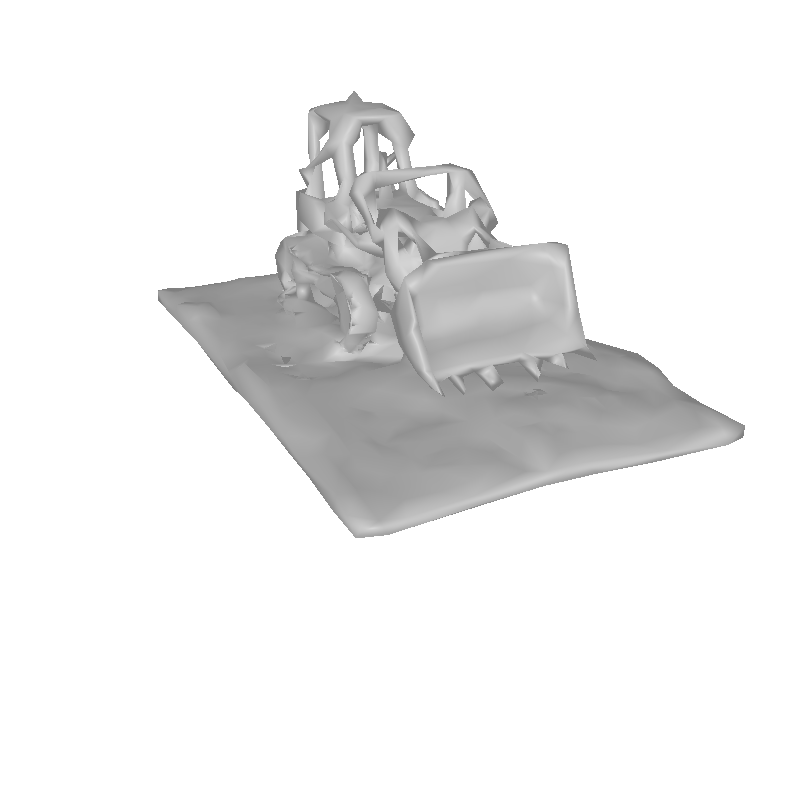}&
            \includegraphics[width=0.24\linewidth]{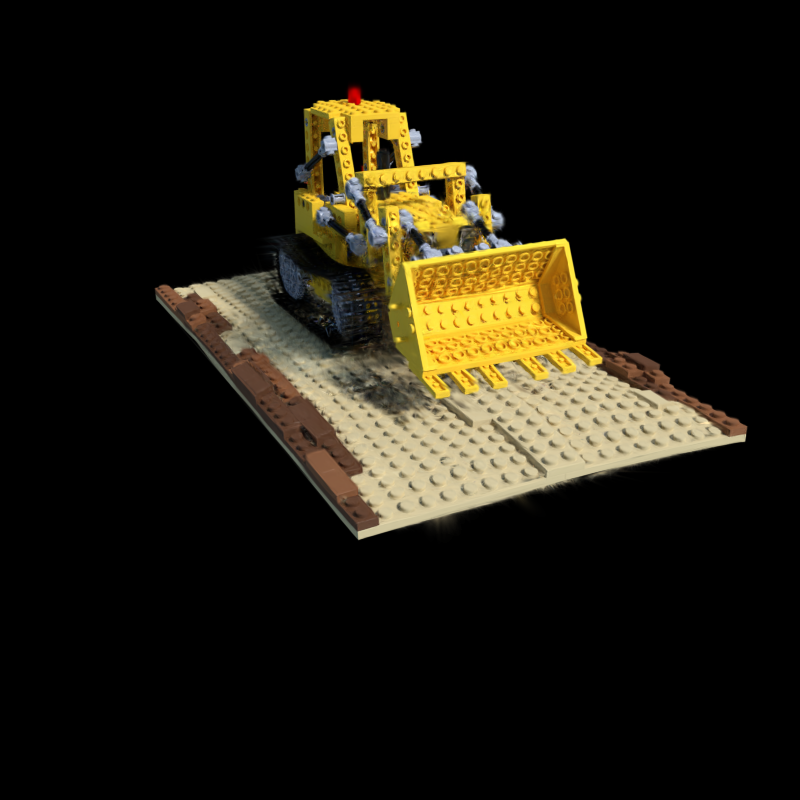}\\

            \includegraphics[width=0.24\linewidth]{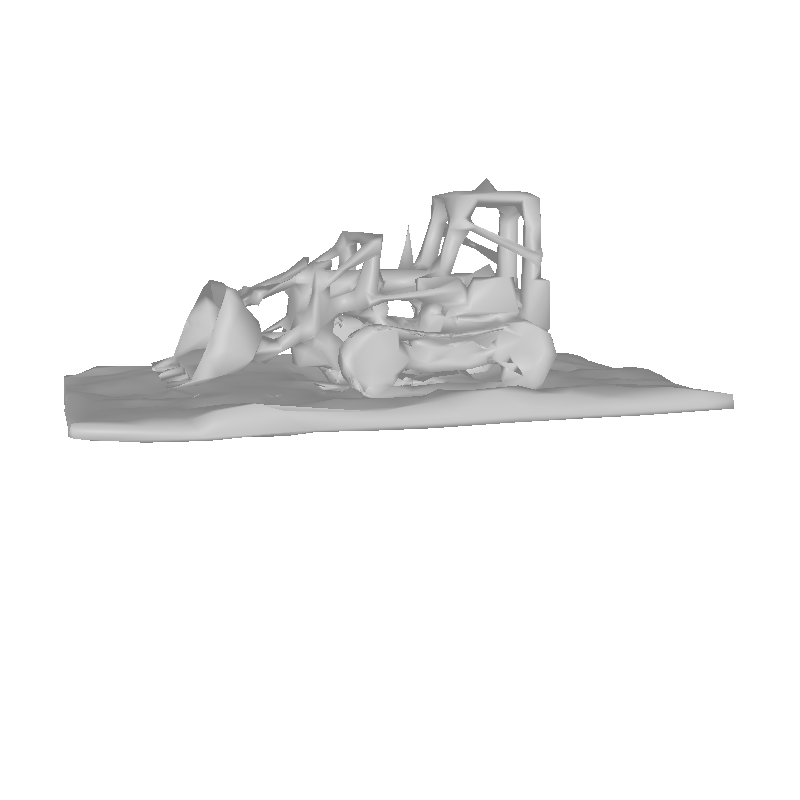}&
            \includegraphics[width=0.24\linewidth]{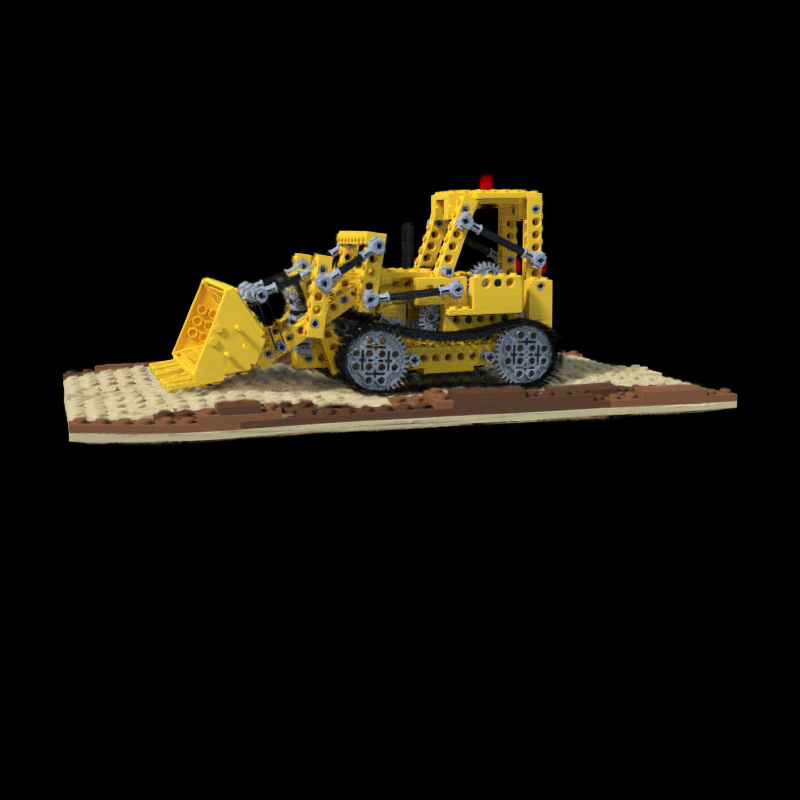}&
            \includegraphics[width=0.24\linewidth]{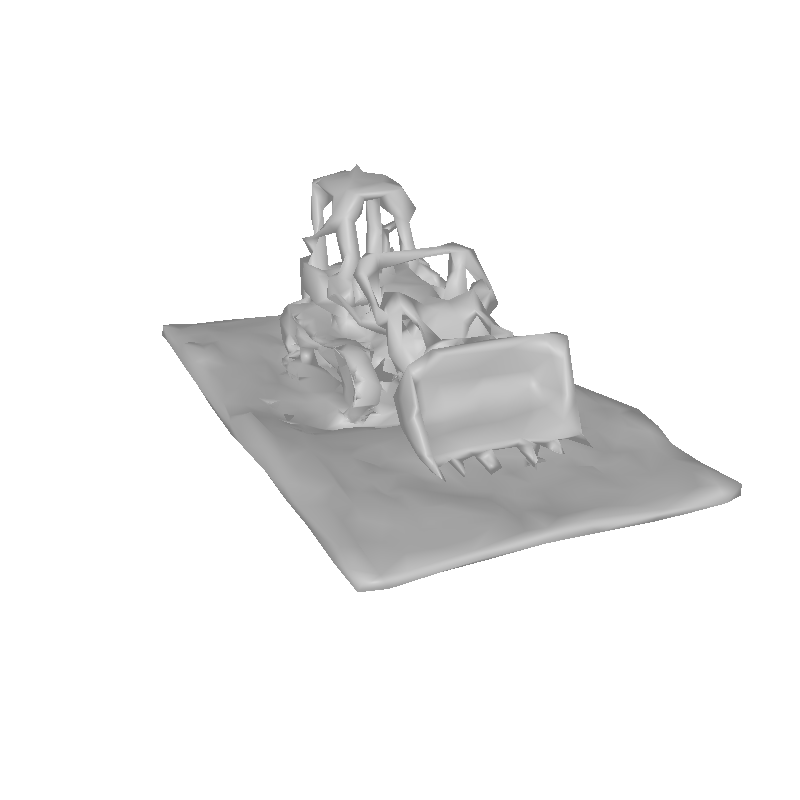}&
            \includegraphics[width=0.24\linewidth]{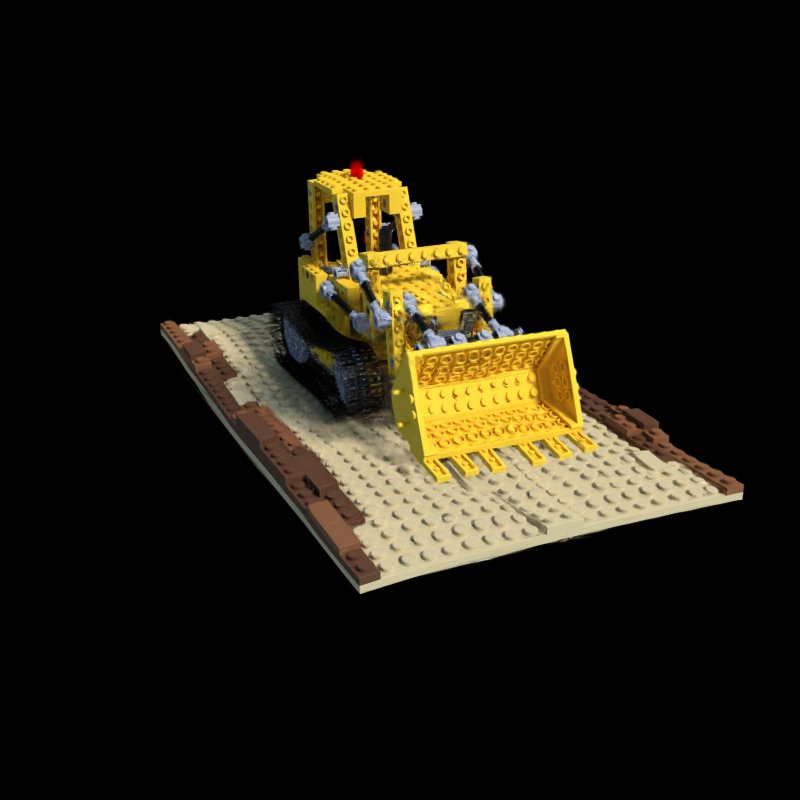}\\

            \includegraphics[width=0.24\linewidth]{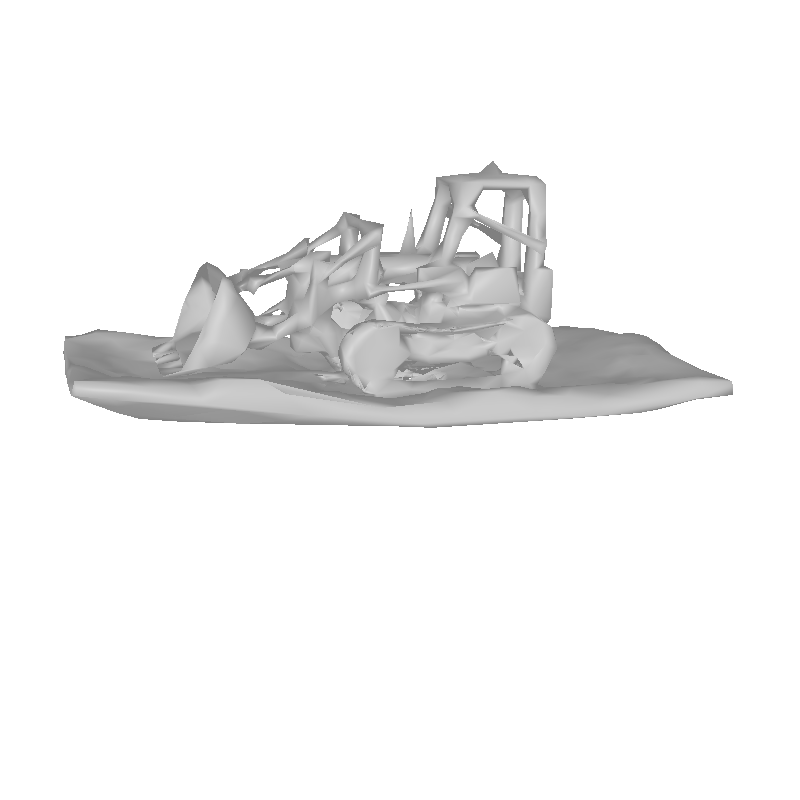}&
            \includegraphics[width=0.24\linewidth]{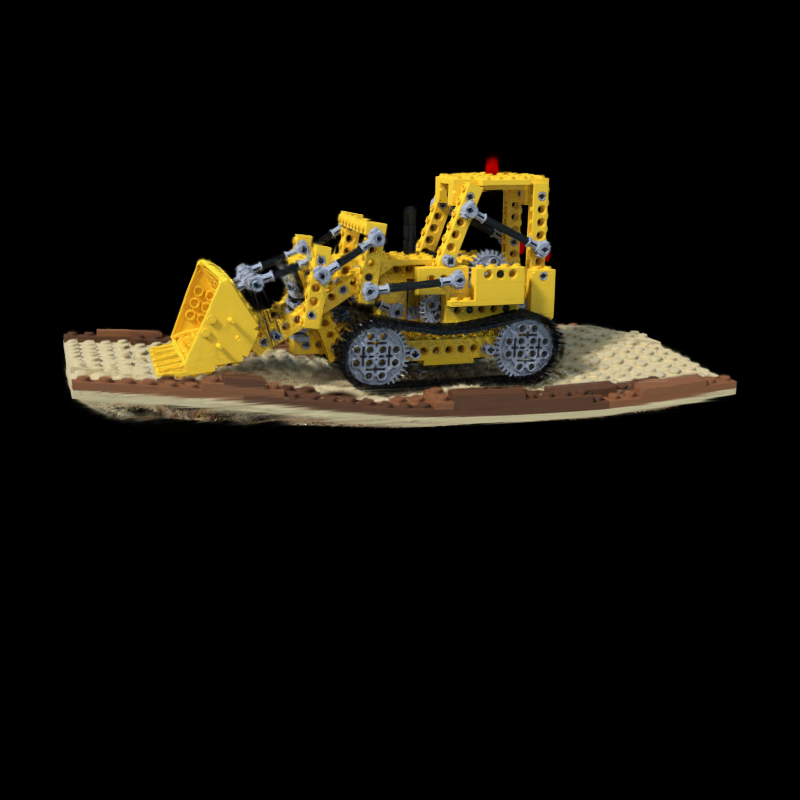}&
            \includegraphics[width=0.24\linewidth]{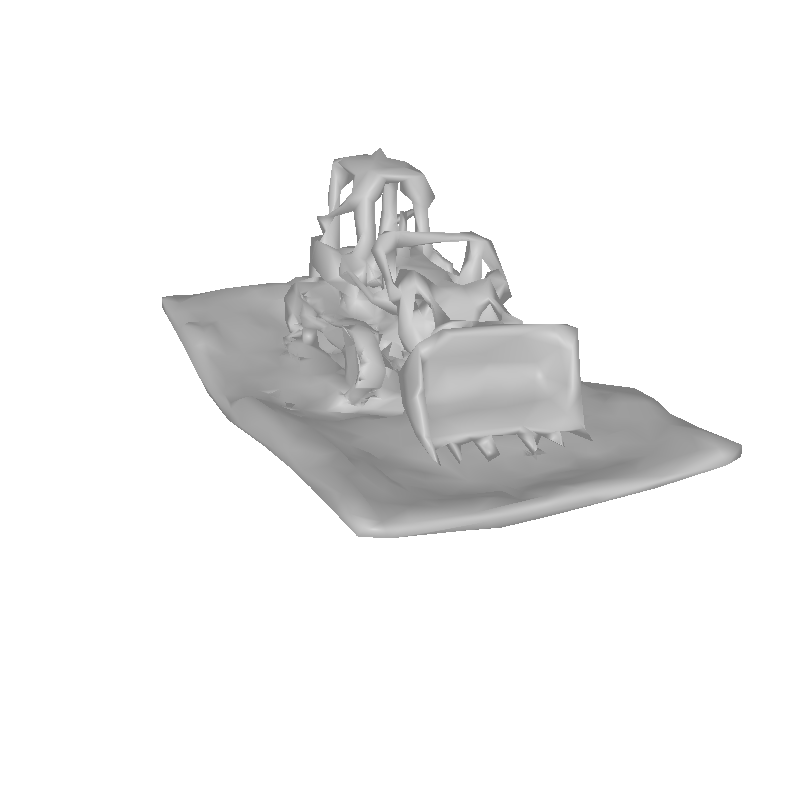}&
            \includegraphics[width=0.24\linewidth]{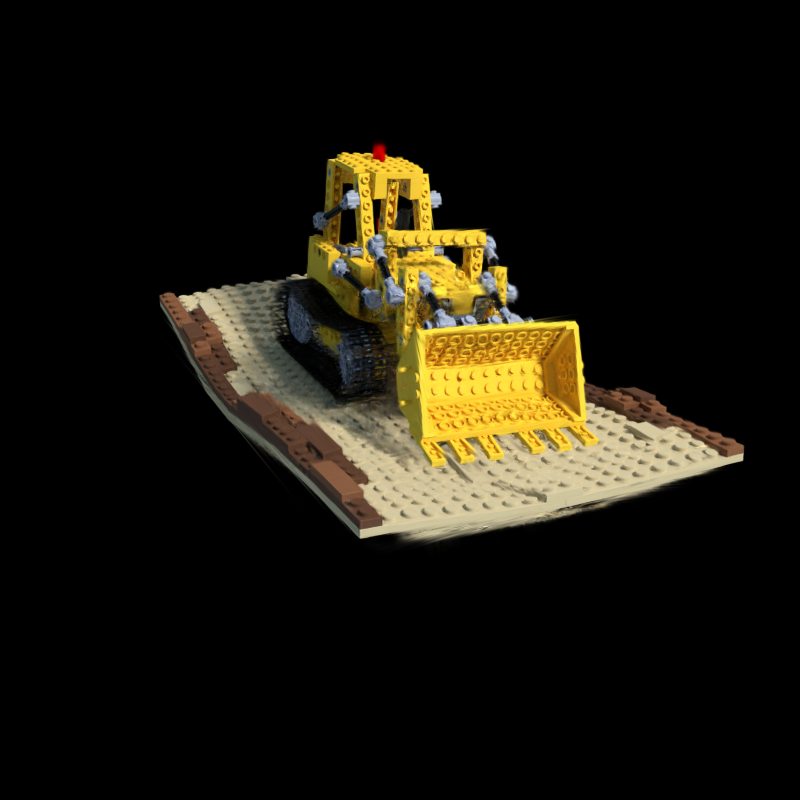}\\

            \includegraphics[width=0.24\linewidth]{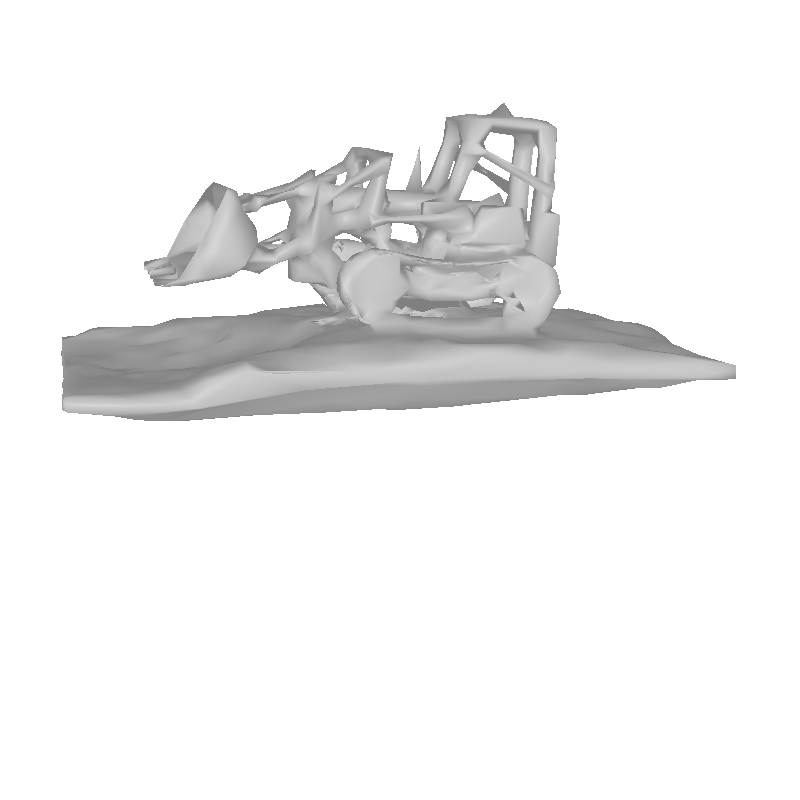}&
            \includegraphics[width=0.24\linewidth]{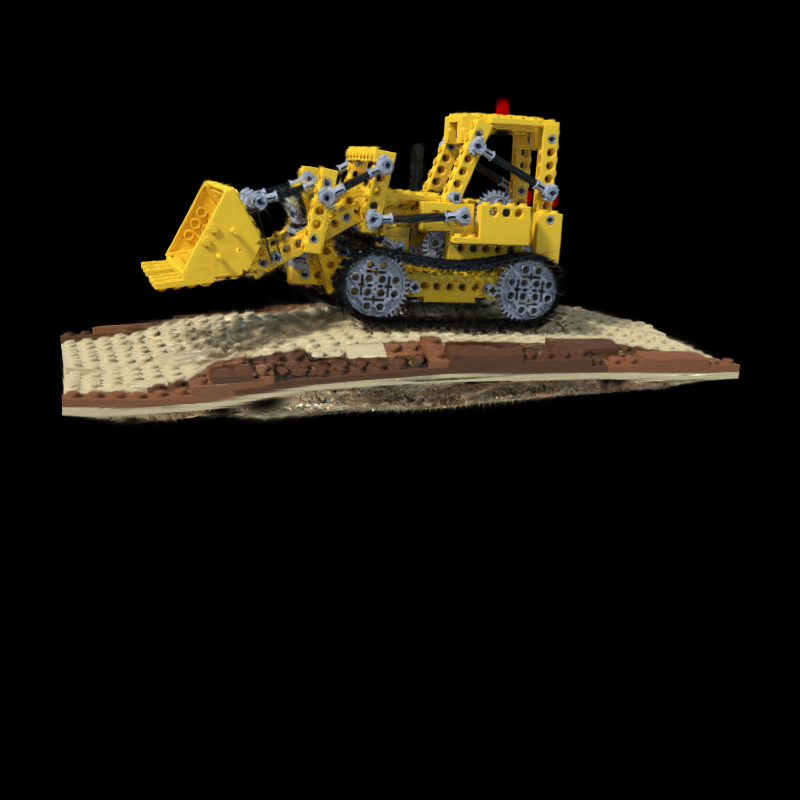}&
            \includegraphics[width=0.24\linewidth]{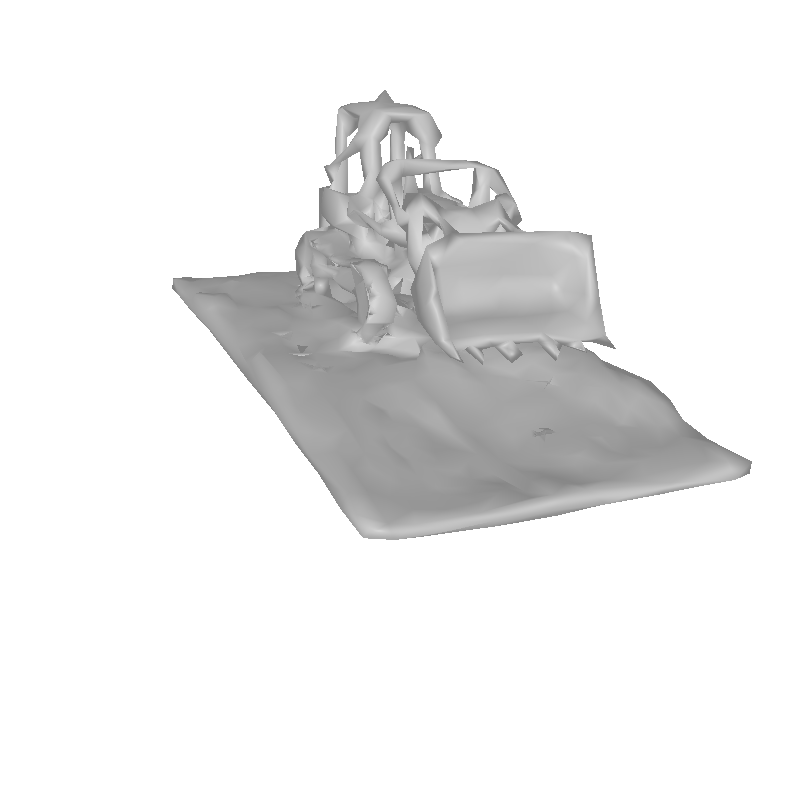}&
            \includegraphics[width=0.24\linewidth]{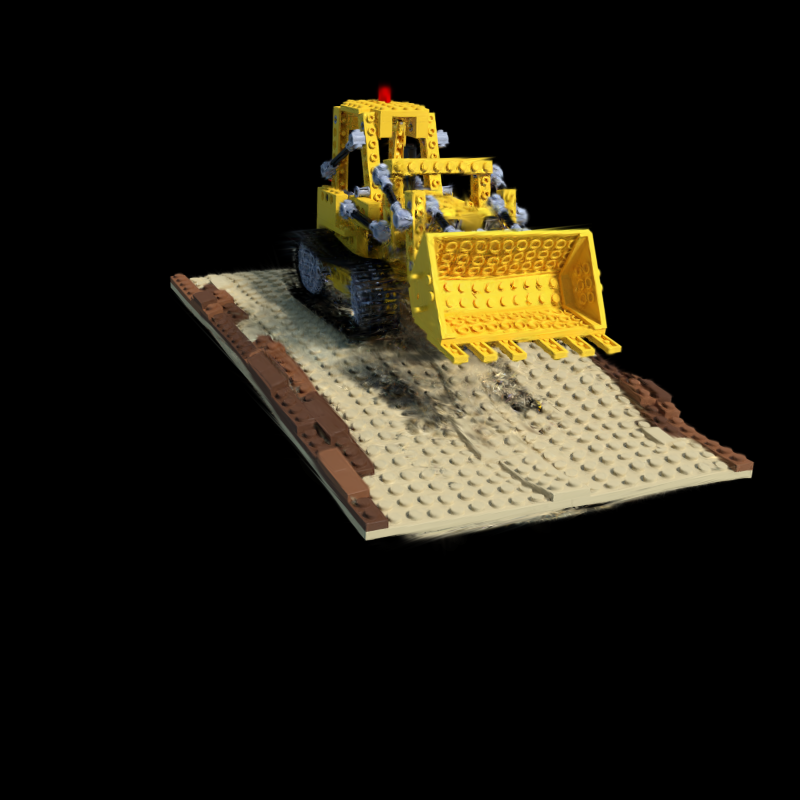}\\

            \includegraphics[width=0.24\linewidth]{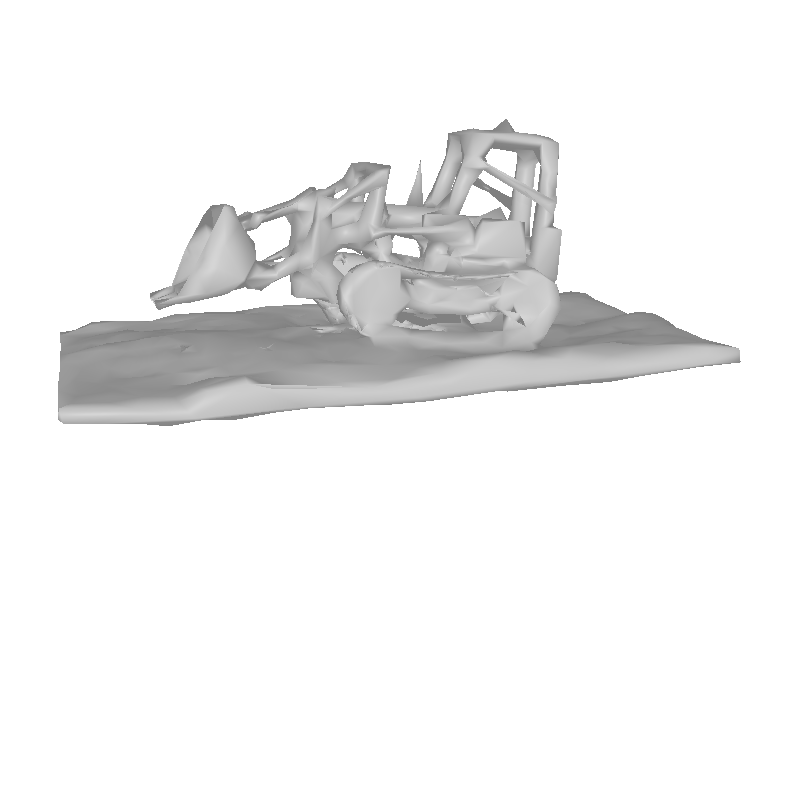}&
            \includegraphics[width=0.24\linewidth]{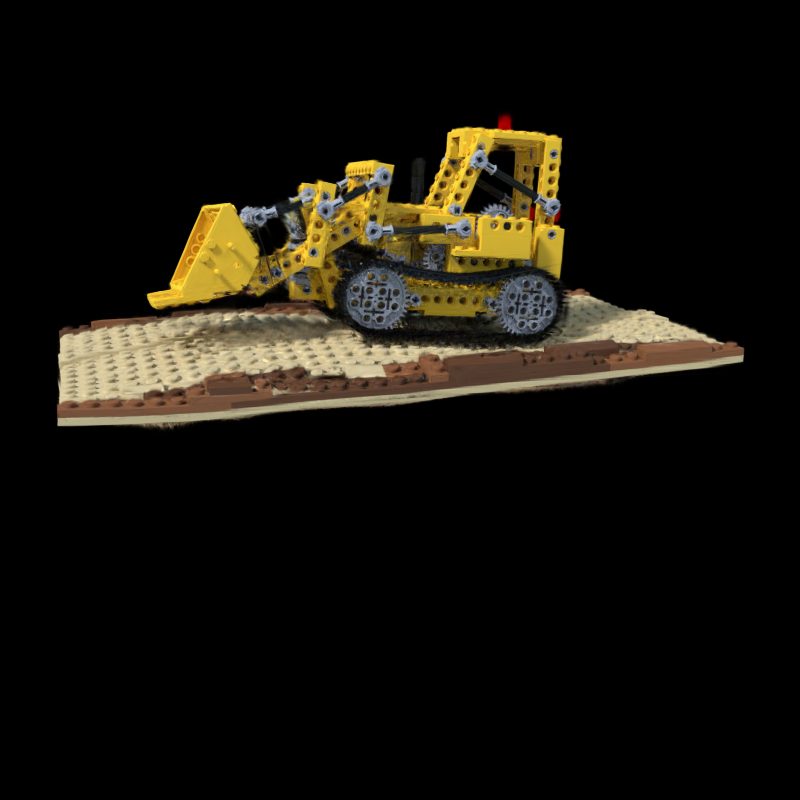}&
            \includegraphics[width=0.24\linewidth]{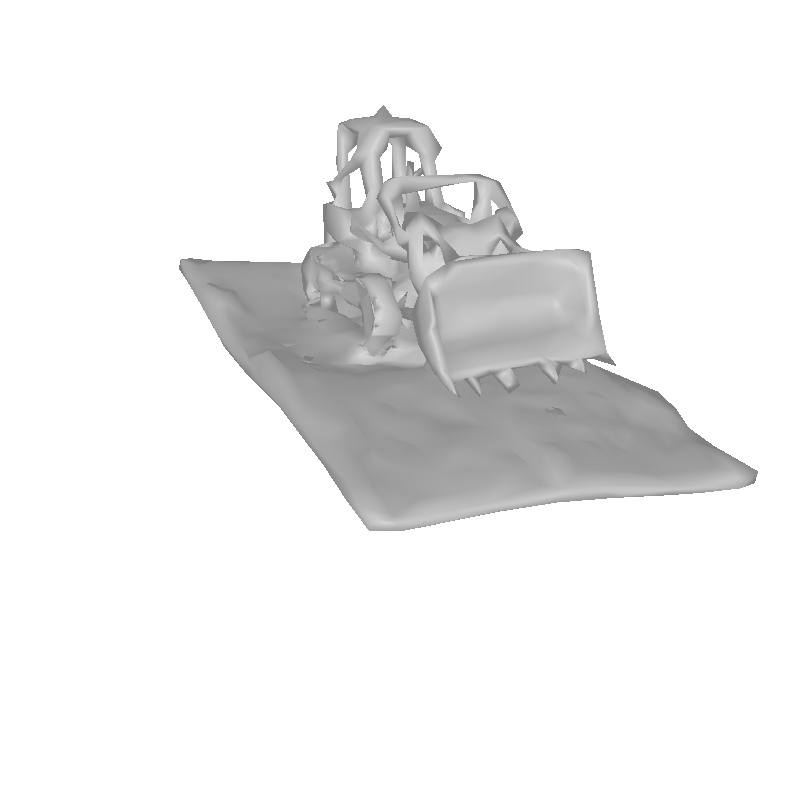}&
            \includegraphics[width=0.24\linewidth]{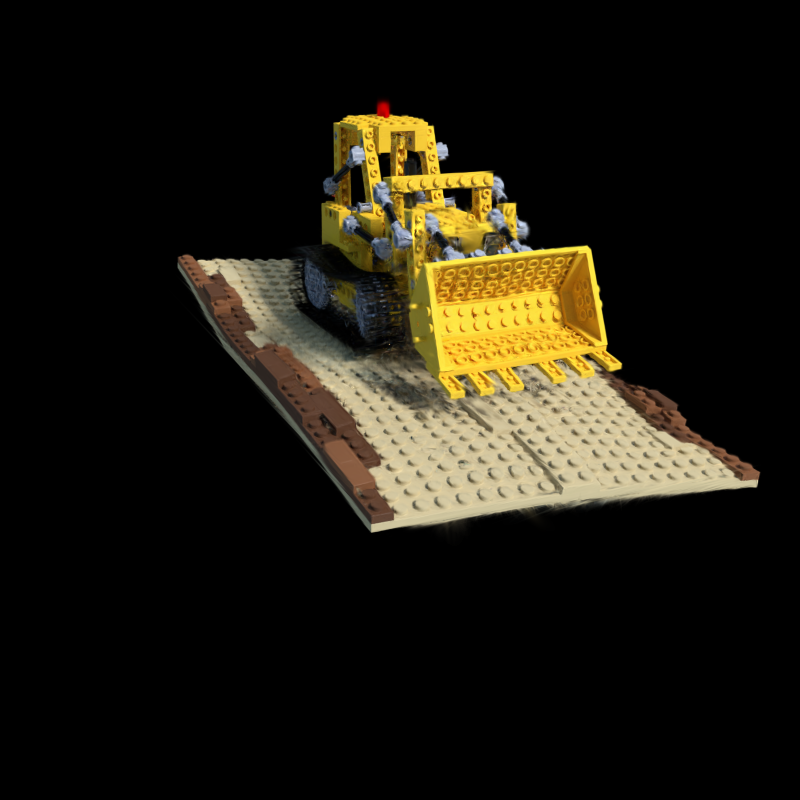}\\

        \end{tabular}
    }
    \caption{\textbf{Soft-body Drop Simulation on Lego.} MaGS effectively simulates the texture and deformation of objects during the collision process when they fall to the ground.}
    \label{fig:add_lego_simu}
\end{figure*}

Figure~\ref{fig:add_lego_simu} illustrates the results of simulations conducted on the Lego scene, demonstrating how our method effectively handles complex object shapes and dynamic motion in a simulated environment. These visualizations highlight the robustness of our approach to managing intricate geometries and movements, further validating its effectiveness in challenging scenarios.

\begin{figure*}
    \centering
    \addtolength{\tabcolsep}{-6.5pt}
    \footnotesize{
        \setlength{\tabcolsep}{1pt} %
        \begin{tabular}{cccc}
            View1 Mesh & View1 MaGS & View2 Mesh & View2 MaGS\\
            \includegraphics[width=0.24\linewidth]{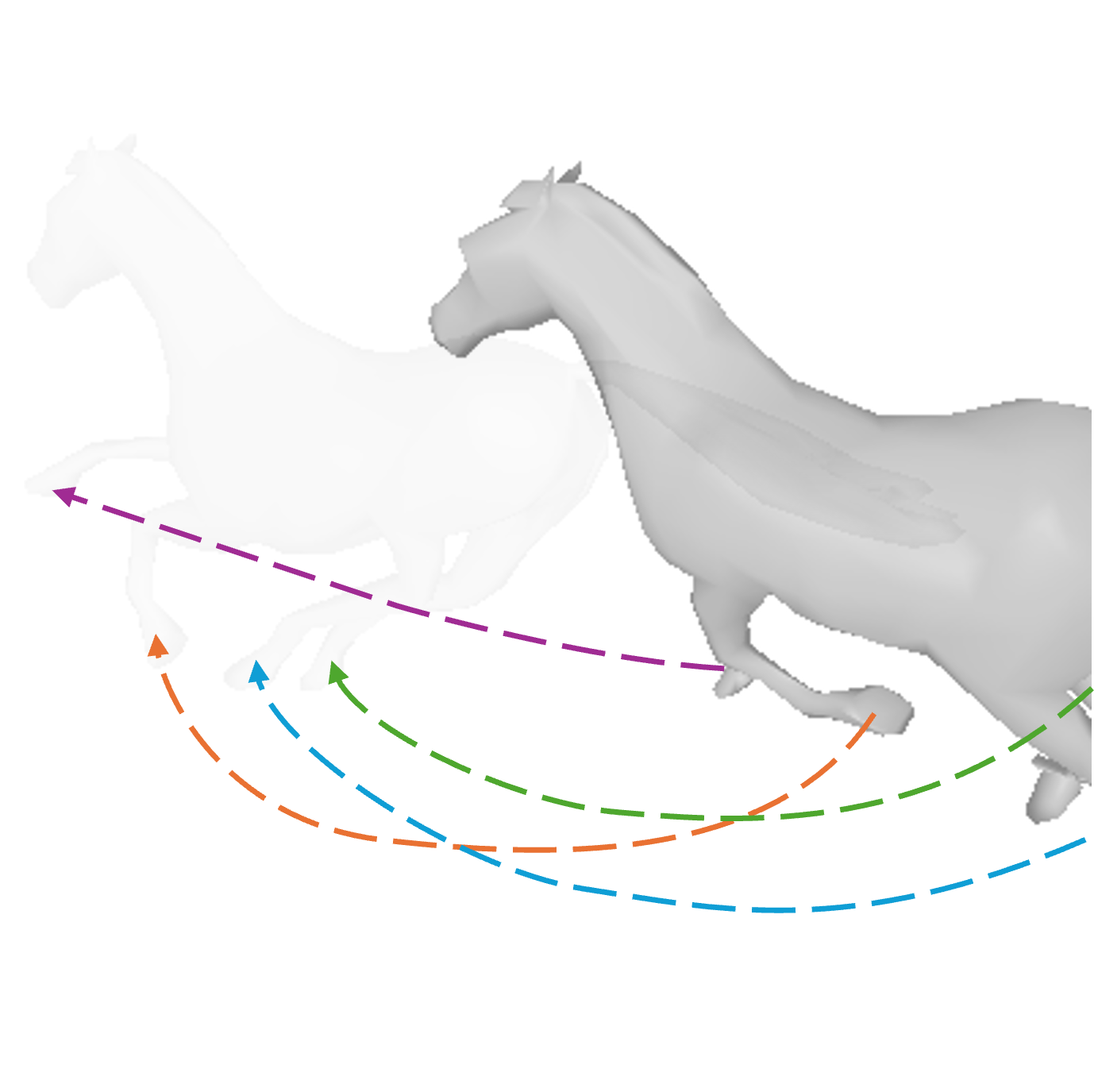}&
            \includegraphics[width=0.24\linewidth]{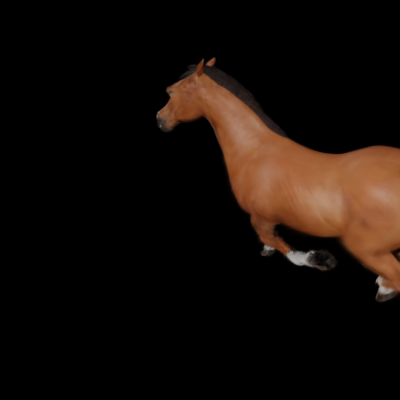}&
            \includegraphics[width=0.24\linewidth]{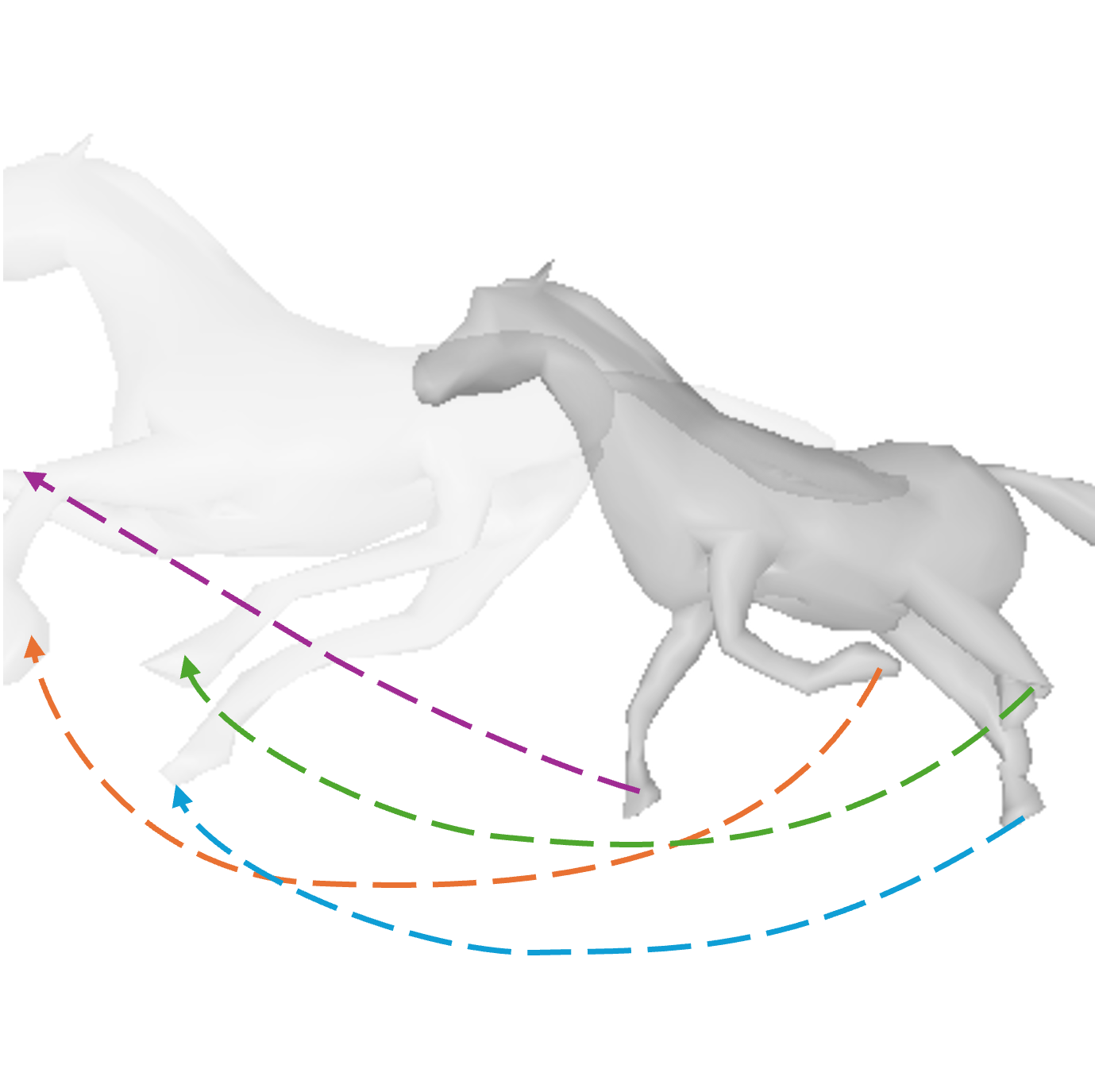}&
            \includegraphics[width=0.24\linewidth]{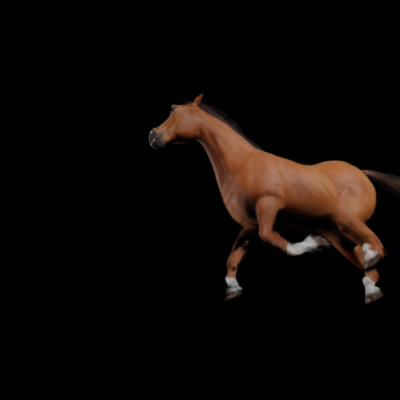}\\

            \includegraphics[width=0.24\linewidth]{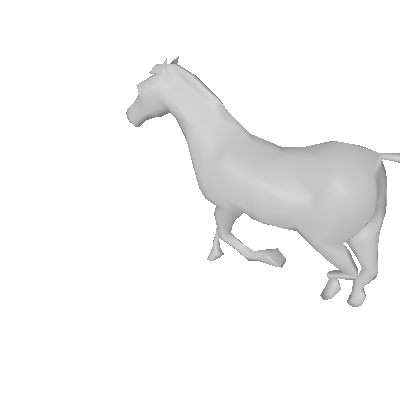}&
            \includegraphics[width=0.24\linewidth]{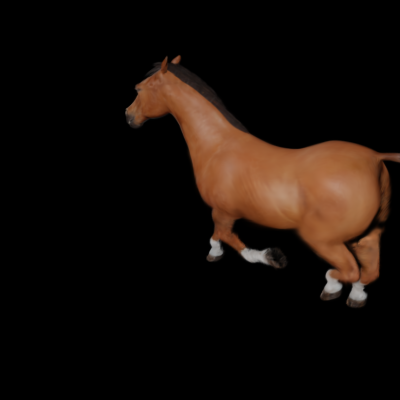}&
            \includegraphics[width=0.24\linewidth]{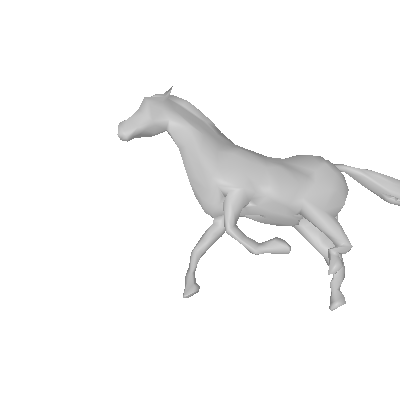}&
            \includegraphics[width=0.24\linewidth]{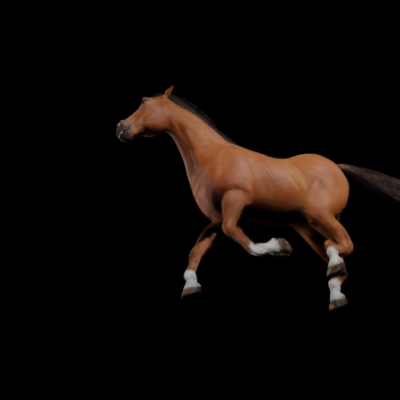}\\

            \includegraphics[width=0.24\linewidth]{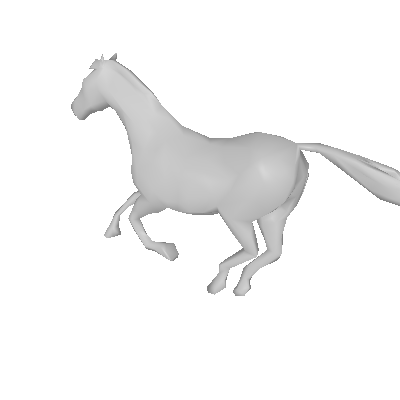}&
            \includegraphics[width=0.24\linewidth]{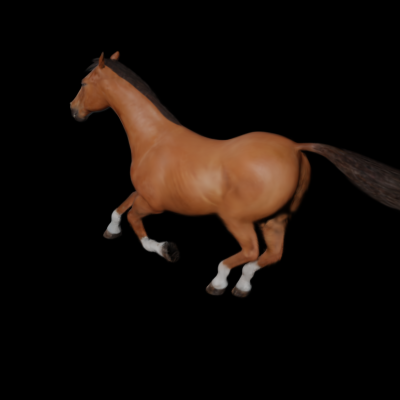}&
            \includegraphics[width=0.24\linewidth]{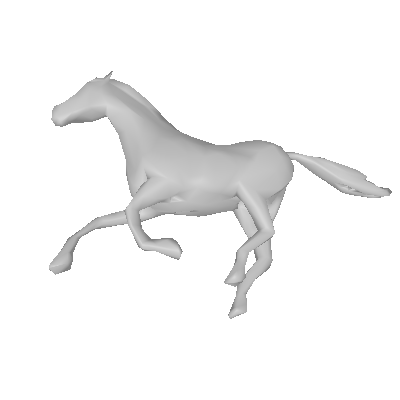}&
            \includegraphics[width=0.24\linewidth]{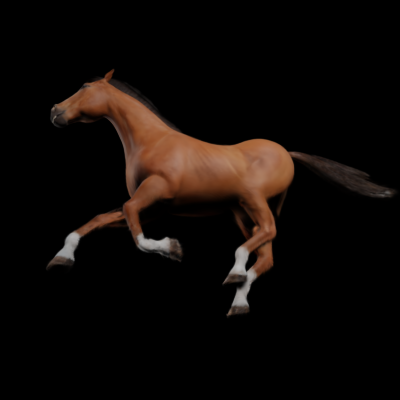}\\

            \includegraphics[width=0.24\linewidth]{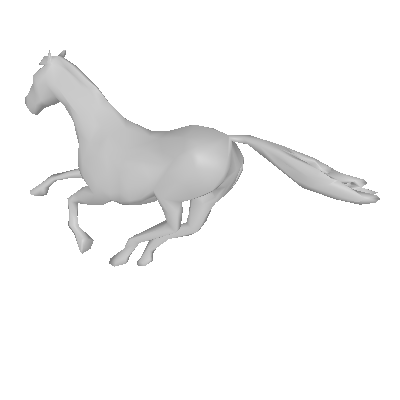}&
            \includegraphics[width=0.24\linewidth]{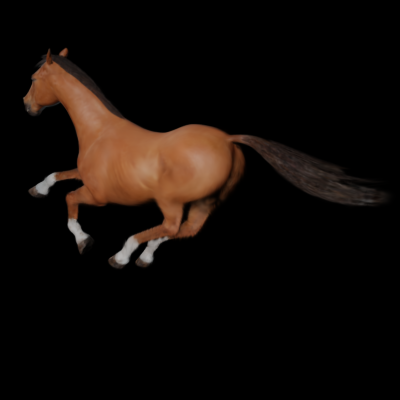}&
            \includegraphics[width=0.24\linewidth]{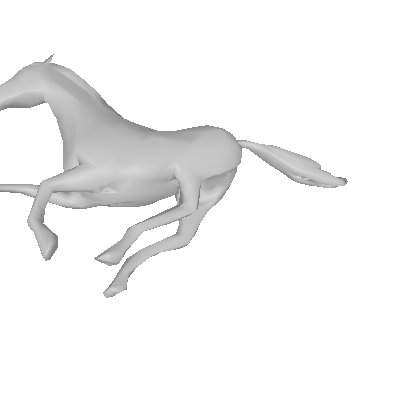}&
            \includegraphics[width=0.24\linewidth]{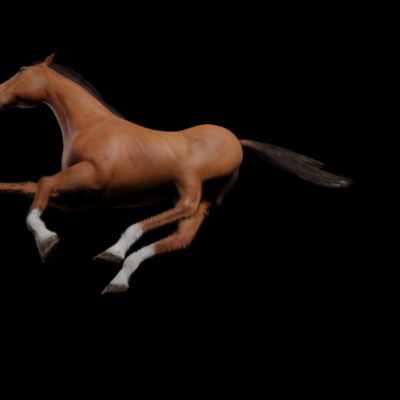}\\

            \includegraphics[width=0.24\linewidth]{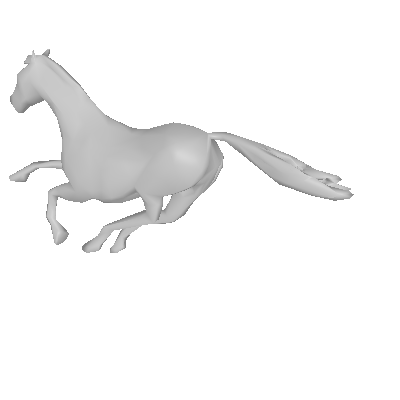}&
            \includegraphics[width=0.24\linewidth]{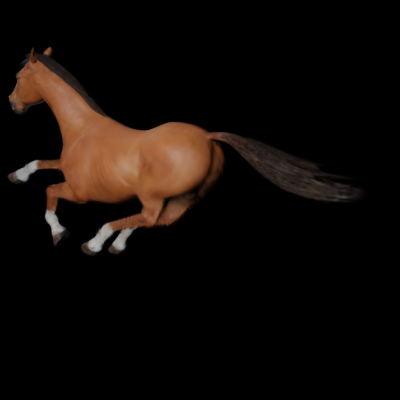}&
            \includegraphics[width=0.24\linewidth]{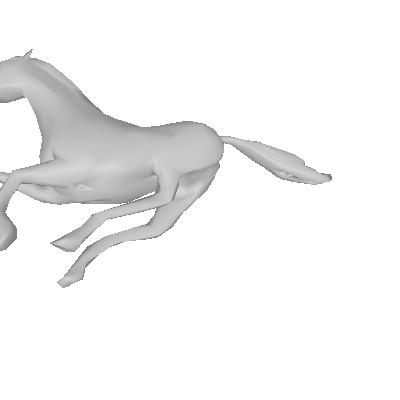}&
            \includegraphics[width=0.24\linewidth]{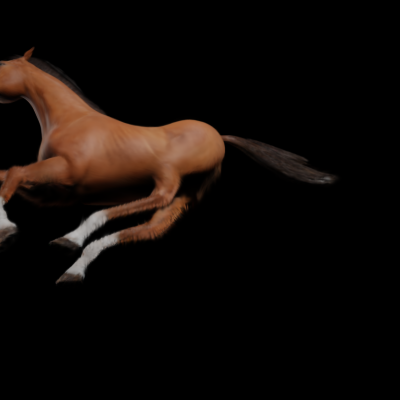}\\

        \end{tabular}
    }
    \caption{\textbf{Drag Editing Simulation on Horse.} The arrows indicate the vectors of the user's dragging forces. MaGS effectively preserves the geometric priors of objects during deformation.}
    \label{fig:add_horse_simu}
\end{figure*}

Similarly, we conducted simulations on the Horse scene with dragging-based editing. The horse's legs undergo significant deformations, with corresponding movements observed in the body. Despite these deformations, the textures in the detailed regions are preserved throughout the simulation, as seen in Figure~\ref{fig:add_horse_simu}.

\begin{figure*}
    \centering
    \addtolength{\tabcolsep}{-6.5pt}
    \footnotesize{
        \setlength{\tabcolsep}{1pt} %
        \begin{tabular}{cccc}
            View1 Frame 1-5 & View1 Frame 6-10 & View2 Frame 1-5 & View2 Frame 6-10 \\
            \includegraphics[width=0.24\linewidth]{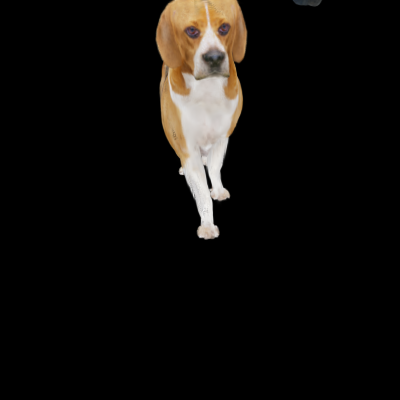}&
            \includegraphics[width=0.24\linewidth]{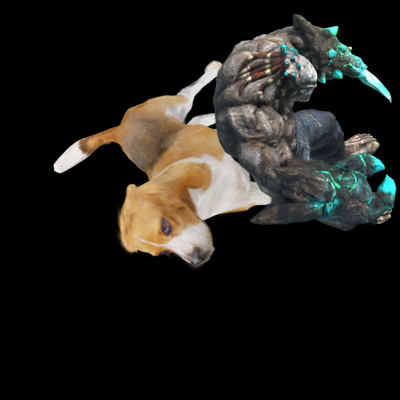}&
            \includegraphics[width=0.24\linewidth]{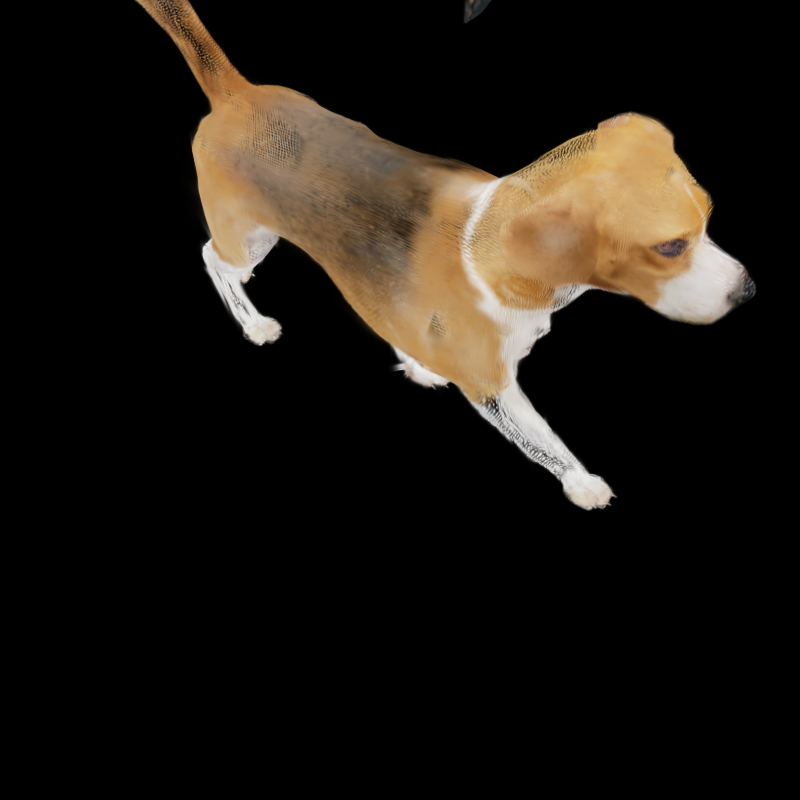}&
            \includegraphics[width=0.24\linewidth]{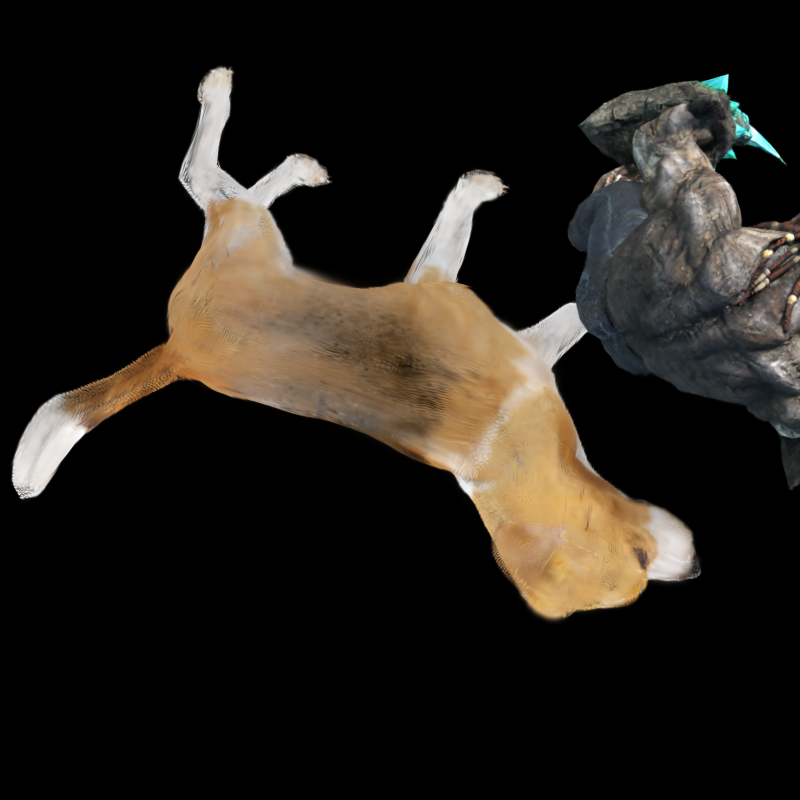}\\
            \includegraphics[width=0.24\linewidth]{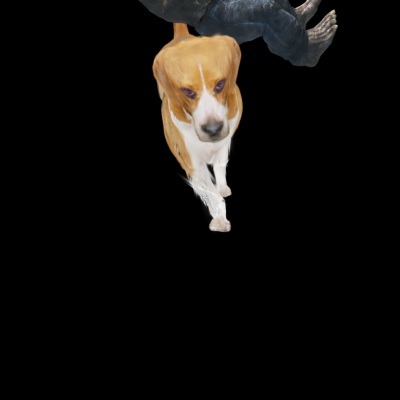}&
            \includegraphics[width=0.24\linewidth]{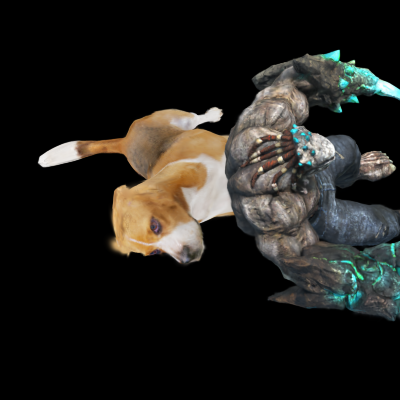}&
            \includegraphics[width=0.24\linewidth]{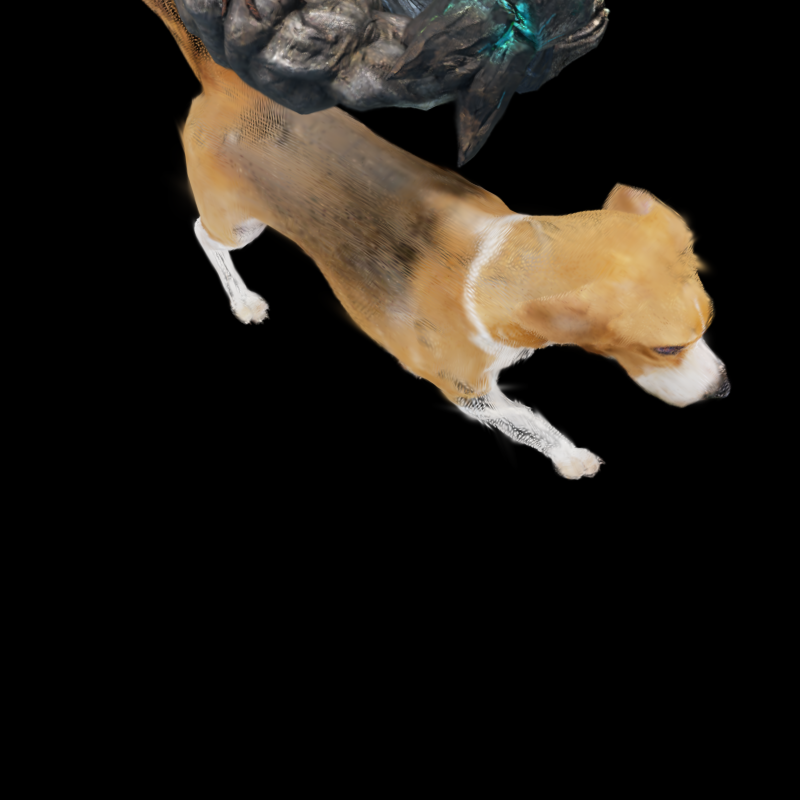}&
            \includegraphics[width=0.24\linewidth]{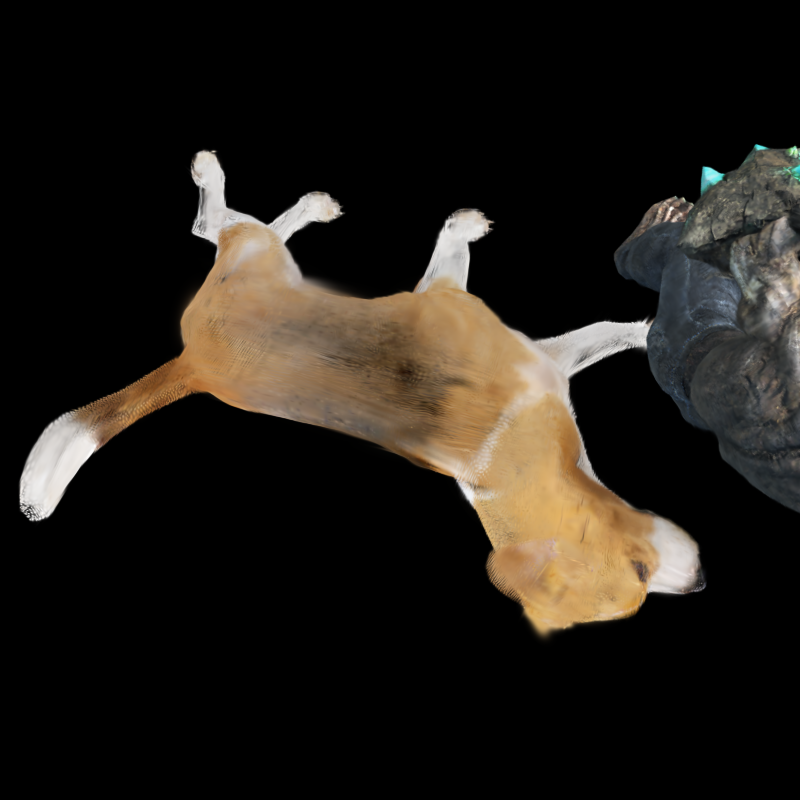}\\
            \includegraphics[width=0.24\linewidth]{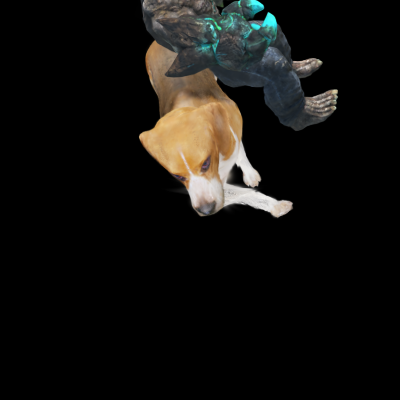}&
            \includegraphics[width=0.24\linewidth]{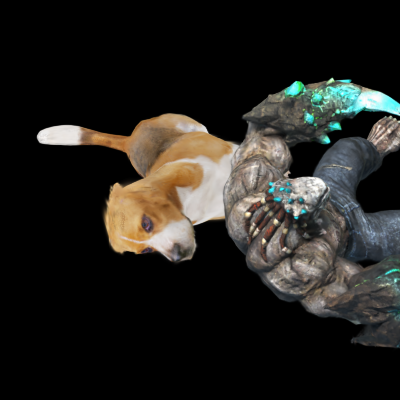}&
            \includegraphics[width=0.24\linewidth]{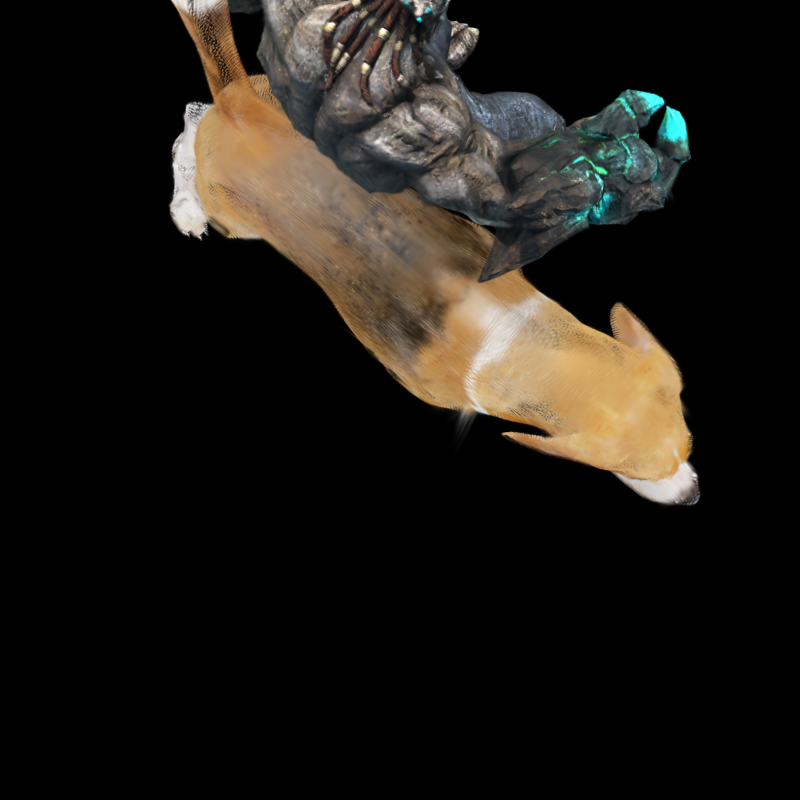}&
            \includegraphics[width=0.24\linewidth]{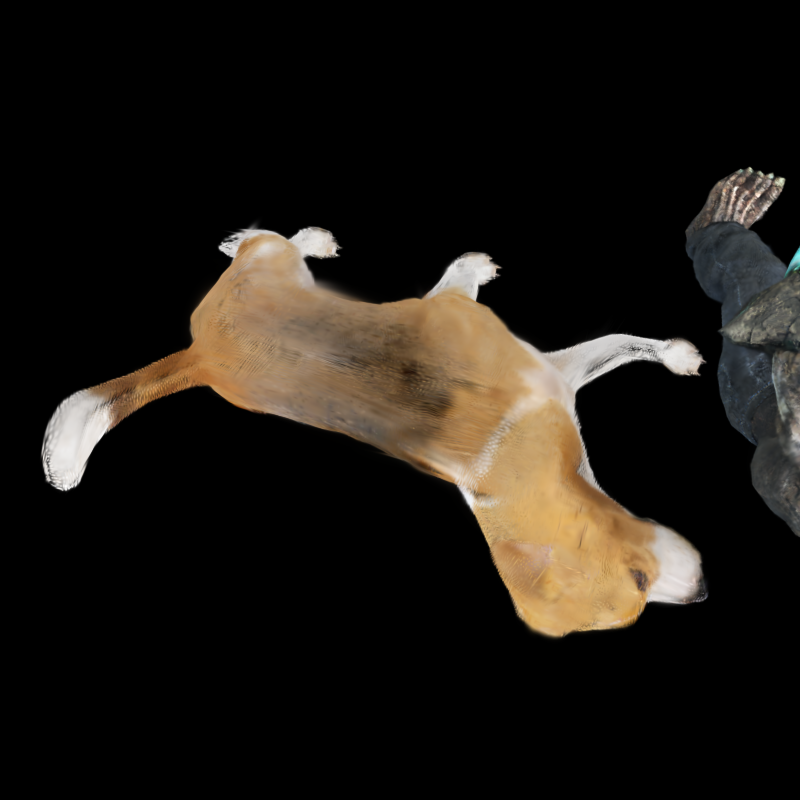}\\
            \includegraphics[width=0.24\linewidth]{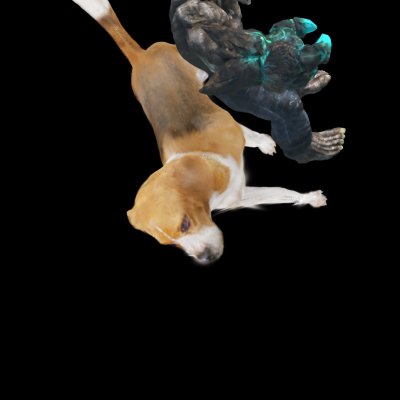}&
            \includegraphics[width=0.24\linewidth]{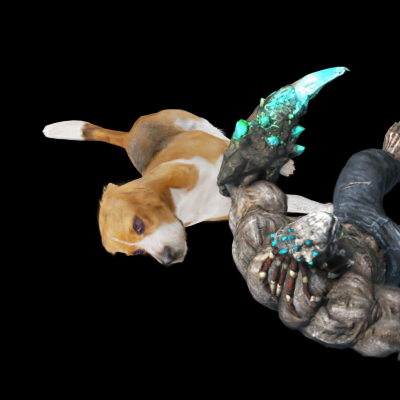}&
            \includegraphics[width=0.24\linewidth]{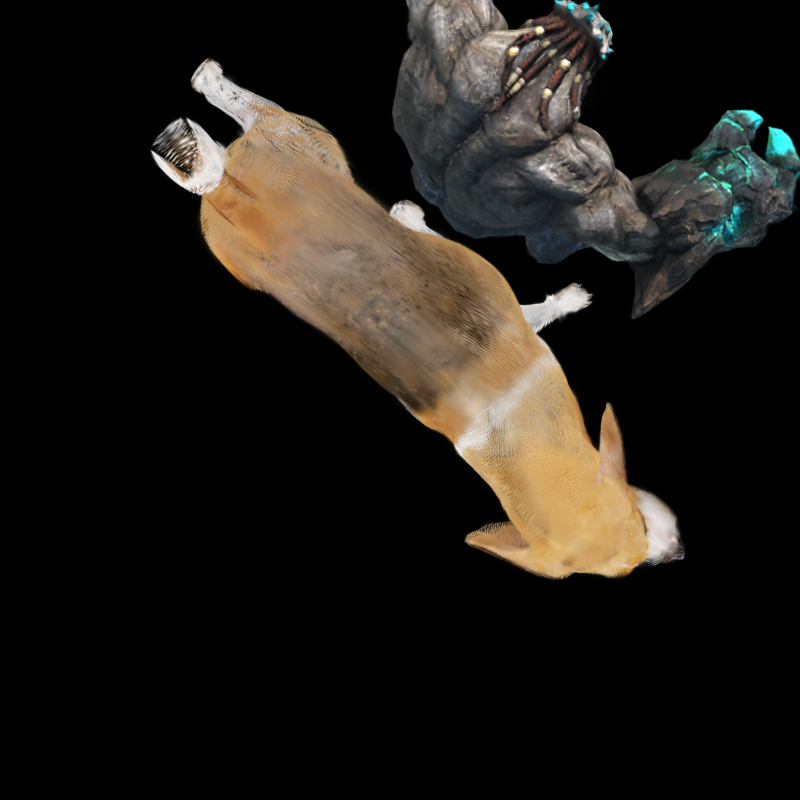}&
            \includegraphics[width=0.24\linewidth]{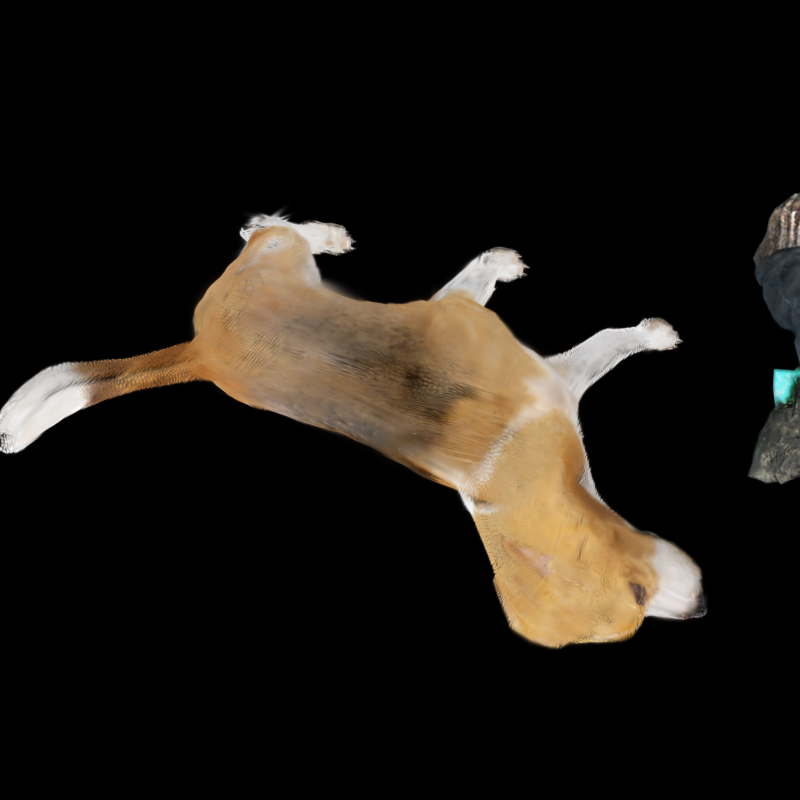}\\
            \includegraphics[width=0.24\linewidth]{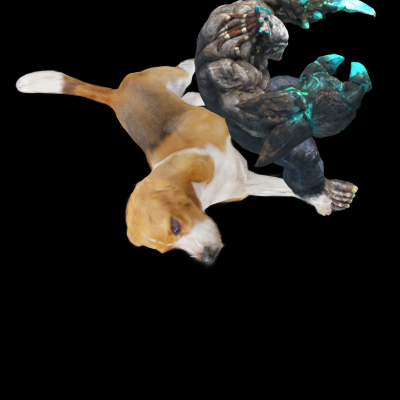}&
            \includegraphics[width=0.24\linewidth]{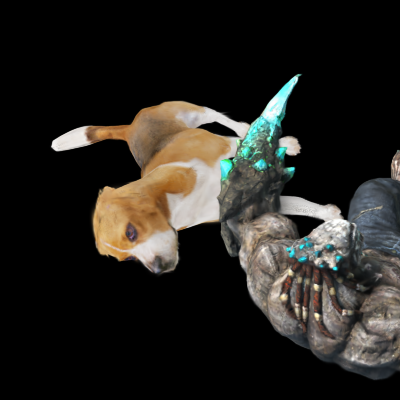}&
            \includegraphics[width=0.24\linewidth]{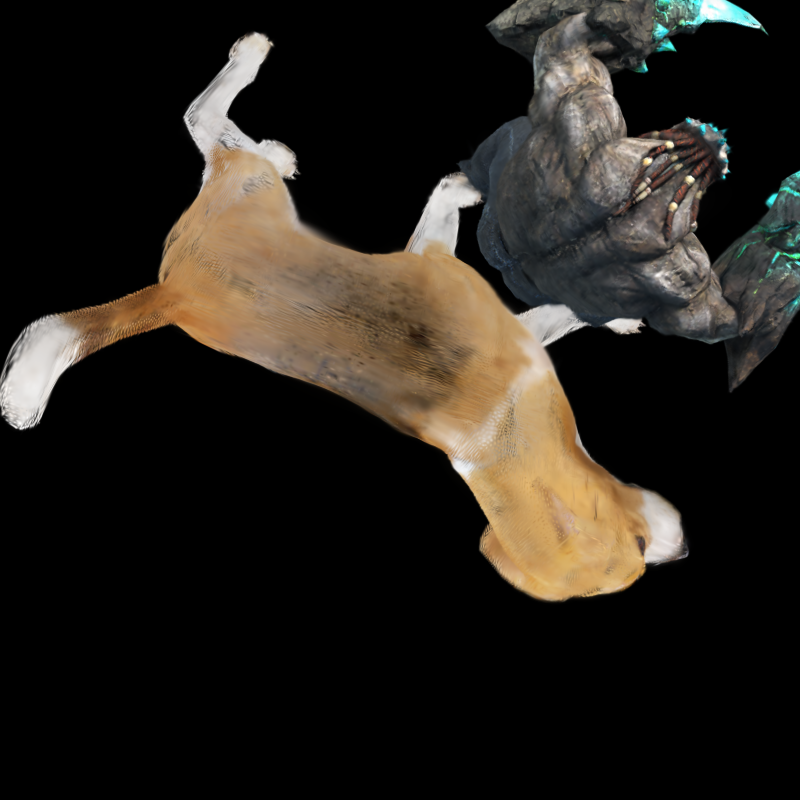}&
            \includegraphics[width=0.24\linewidth]{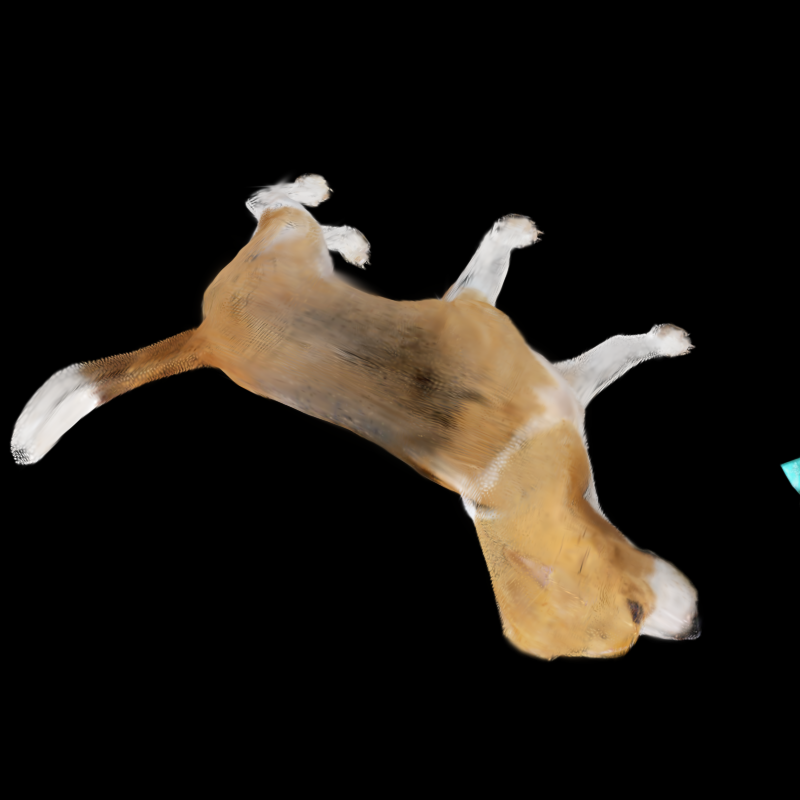}\\

        \end{tabular}
    }
	\caption{\textbf{
Gravity and Collision Simulation of Beagle and Mutant.} MaGS effectively simulates the physical-plausible deformations during the interaction of multiple objects.}
    \label{fig:add_multigs_simu}
\end{figure*}

Furthermore, we applied gravity and collision simulations to the Beagle from DG-Mesh and the Mutant from D-NeRF. After repositioning the objects, gravitational forces were applied. The results, visualized in Figure~\ref{fig:add_multigs_simu}, demonstrate the dynamics of falling, collision, and rebound behaviors. For clarity, we selected one frame every five frames for presentation, with the actual simulations exhibiting smoother and more coherent transitions.

\section{Detail Quantitative Comparisons} \label{sec:quantitative}

In this section, we present a comprehensive quantitative analysis of the performance of our proposed methods compared to prior approaches. Detailed experimental results from ablation studies and scene-specific evaluations are included to validate the robustness and effectiveness of our approach.

Table~\ref{tab:ablation_dnerf_full} summarizes the results of the ablation experiments conducted on the D-NeRF dataset, reporting metrics such as L1 loss, PSNR, and SSIM for each scene. These results demonstrate that the key components of our method (RMD-Net, RGD-Net, and Hovering) significantly improve the model’s performance.

Table~\ref{tab:dgmesh_dataset_full2} provides an analysis of the DGMesh dataset. The results indicate that our method consistently achieves higher PSNR and EMD scores across all scenes compared to existing methods, showcasing its adaptability and precision in handling complex mesh structures and dynamic motions. 

It is worth noting that the \textit{Torus2Sphere} scene involves objects undergoing significant topological changes. The provided ground-truth meshes for this scene do not guarantee consistent vertex and face correspondences across frames. This makes it unsuitable for evaluating methods that preserve inter-frame vertex and face correspondences. While our approach achieved higher EMD and PSNR scores for this scene, we have excluded it from the main quantitative tables due to these inconsistencies.

As mentioned in the main text, the test set for the Lego scene contains temporal and image inconsistencies. Yang \emph{et al.} provided a corrected version of the Lego, and we conducted tests on this corrected dataset. The results shown in Table~\ref{tab:add_lego} demonstrate that our method still outperforms other approaches on the Lego scene. 

\begin{table*}[]
\centering
\caption{\textbf{Ablation experiments on the D-NeRF dataset.}}
\label{tab:ablation_dnerf_full}
\resizebox{\textwidth}{!}{%
\begin{tabular}{@{}lcccccccccccc@{}}\toprule
 & \multicolumn{3}{c}{Bouncingballs} & \multicolumn{3}{c}{Hellwarrior} & \multicolumn{3}{c}{Hook} & \multicolumn{3}{c}{Jumpingjacks} \\
\multirow{-2}{*}{Method} & PSNR↑ & SSIM↑ & LPIPS↓ & PSNR↑ & SSIM↑ & LPIPS↓ & PSNR↑ & SSIM↑ & LPIPS↓ & PSNR↑ & SSIM↑ & LPIPS↓ \\ \midrule
MaGS w/o MDF and RDF & \cellcolor[HTML]{FFFF8F}40.31 & \cellcolor[HTML]{FFFF8F}0.9973 & \cellcolor[HTML]{FFFF8F}0.0065 & \cellcolor[HTML]{FFFF8F}42.38 & \cellcolor[HTML]{FFFF8F}0.9939 & \cellcolor[HTML]{FFFF8F}0.0137 & 37.67 & 0.9963 & 0.0098 & 40.30 & 0.9984 & 0.0051 \\
MaGS w/o Hover & 39.29 & 0.9966 & 0.0076 & 42.05 & 0.9937 & 0.0147 & \cellcolor[HTML]{FFFF8F}38.80 & \cellcolor[HTML]{FFFF8F}0.9972 & \cellcolor[HTML]{FFFF8F}0.0074 & \cellcolor[HTML]{FFFF8F}41.61 & \cellcolor[HTML]{FFFF8F}0.9988 & \cellcolor[HTML]{FFFF8F}0.0036 \\
MaGS w/o RDF & \cellcolor[HTML]{FAC791}40.80 & \cellcolor[HTML]{FAC791}0.9975 & \cellcolor[HTML]{FAC791}0.0058 & \cellcolor[HTML]{FAC791}42.86 & \cellcolor[HTML]{FAC791}0.9947 & \cellcolor[HTML]{FAC791}0.0125 & \cellcolor[HTML]{FAC791}40.28 & \cellcolor[HTML]{FAC791}0.9981 & \cellcolor[HTML]{FAC791}0.0055 & \cellcolor[HTML]{FAC791}42.30 & \cellcolor[HTML]{FAC791}0.9990 & \cellcolor[HTML]{FAC791}0.0031 \\
MaGS & \cellcolor[HTML]{F59194}41.97 & \cellcolor[HTML]{F59194}0.9976 & \cellcolor[HTML]{F59194}0.0055 & \cellcolor[HTML]{F59194}43.69 & \cellcolor[HTML]{F59194}0.9957 & \cellcolor[HTML]{F59194}0.0098 & \cellcolor[HTML]{F59194}41.23 & \cellcolor[HTML]{F59194}0.9984 & \cellcolor[HTML]{F59194}0.0049 & \cellcolor[HTML]{F59194}44.29 & \cellcolor[HTML]{F59194}0.9993 & \cellcolor[HTML]{F59194}0.0022 \\ \midrule
 & \multicolumn{3}{c}{Mutant} & \multicolumn{3}{c}{Standup} & \multicolumn{3}{c}{Trex} & \multicolumn{3}{c}{Average} \\
\multirow{-2}{*}{Method} & PSNR↑ & SSIM↑ & LPIPS↓ & PSNR↑ & SSIM↑ & LPIPS↓ & PSNR↑ & SSIM↑ & LPIPS↓ & PSNR↑ & SSIM↑ & LPIPS↓ \\ \midrule
MaGS w/o MDF and RDF & \cellcolor[HTML]{FFFF8F}44.31 & \cellcolor[HTML]{FFFF8F}0.9990 & \cellcolor[HTML]{FFFF8F}0.0025 & 44.97 & 0.9991 & 0.0026 & 38.07 & 0.9976 & 0.0047 & 41.14 & 0.9974 & 0.0064 \\
MaGS w/o Hover & 43.49 & 0.9989 & 0.0029 & \cellcolor[HTML]{FFFF8F}47.98 & \cellcolor[HTML]{FFFF8F}0.9996 & \cellcolor[HTML]{FFFF8F}0.0014 & \cellcolor[HTML]{FFFF8F}39.85 & \cellcolor[HTML]{FFFF8F}0.9989 & \cellcolor[HTML]{FFFF8F}0.0036 & \cellcolor[HTML]{FFFF8F}41.87 & \cellcolor[HTML]{FFFF8F}0.9977 & \cellcolor[HTML]{FFFF8F}0.0059 \\
MaGS w/o RDF & \cellcolor[HTML]{FAC791}44.70 & \cellcolor[HTML]{FAC791}0.9991 & \cellcolor[HTML]{FAC791}0.0023 & \cellcolor[HTML]{FAC791}48.66 & \cellcolor[HTML]{F59194}0.9997 & \cellcolor[HTML]{FAC791}0.0011 & \cellcolor[HTML]{FAC791}41.28 & \cellcolor[HTML]{FAC791}0.9992 & \cellcolor[HTML]{FAC791}0.0027 & \cellcolor[HTML]{FAC791}42.98 & \cellcolor[HTML]{FAC791}0.9982 & \cellcolor[HTML]{FAC791}0.0047 \\
MaGS & \cellcolor[HTML]{F59194}46.42 & \cellcolor[HTML]{F59194}0.9996 & \cellcolor[HTML]{F59194}0.0019 & \cellcolor[HTML]{F59194}49.16 & \cellcolor[HTML]{F59194}0.9997 & \cellcolor[HTML]{F59194}0.0010 & \cellcolor[HTML]{F59194}41.65 & \cellcolor[HTML]{F59194}0.9993 & \cellcolor[HTML]{F59194}0.0025 & \cellcolor[HTML]{F59194}44.06 & \cellcolor[HTML]{F59194}0.9985 & \cellcolor[HTML]{F59194}0.0040 \\ \bottomrule
\end{tabular}%
}
\end{table*}

\begin{table*}[]
\centering
\caption{\textbf{Quantitative Results on DG-Mesh.}}
\label{tab:dgmesh_dataset_full2}
\begin{tabular}{@{}lccccccccc@{}}
\toprule
\multicolumn{1}{c}{} & \multicolumn{3}{c}{Beagle} & \multicolumn{3}{c}{Girlwalk} & \multicolumn{3}{c}{Duck} \\
\multicolumn{1}{c}{\multirow{-2}{*}{Methods}} & CD↓ & EMD↓ & PSNR↑ & CD↓ & EMD↓ & PSNR↑ & CD↓ & EMD↓ & PSNR↑ \\ \midrule
D-NeRF & 1.0010 & 0.1490 & 34.47 & 0.6010 & 0.1900 & 28.63 & 0.9340 & 0.0730 & 29.79 \\
K-Plane & 0.8100 & 0.1220 & 38.33 & \cellcolor[HTML]{FFFF8F}0.4950 & 0.1730 & 32.12 & 1.0850 & \cellcolor[HTML]{FFFF8F}0.0550 & 33.36 \\
HexPlane & 0.8700 & \cellcolor[HTML]{FFFF8F}0.1150 & 38.03 & 0.5970 & 0.1550 & 31.77 & 2.1610 & 0.0900 & 32.11 \\
TiNeuVox-B & 0.8740 & 0.1290 & 38.97 & 0.5680 & 0.1840 & 32.81 & 0.9690 & 0.0590 & 34.33 \\
DG-Mesh & \cellcolor[HTML]{FFFF8F}0.6390 & 0.1170 & 33.41 & 0.7260 & 0.1360 & 32.91 & \cellcolor[HTML]{F59194}0.7900 & \cellcolor[HTML]{F59194}0.0470 & 32.26 \\
DynaSurfGS & \cellcolor[HTML]{FAC791}0.6090 & \cellcolor[HTML]{F59194}0.1100 & \cellcolor[HTML]{FFFF8F}40.74 & \cellcolor[HTML]{FAC791}0.4430 & \cellcolor[HTML]{F59194}0.1280 & \cellcolor[HTML]{FFFF8F}33.31 & \cellcolor[HTML]{FAC791}0.8060 & \cellcolor[HTML]{F59194}0.0470 & \cellcolor[HTML]{FFFF8F}36.31 \\
Dynamic 2D Gaussians & \cellcolor[HTML]{F59194}0.5440 & 0.1220 & \cellcolor[HTML]{FAC791}41.94 & \cellcolor[HTML]{F59194}0.3240 & \cellcolor[HTML]{FAC791}0.1290 & \cellcolor[HTML]{FAC791}41.17 & 1.0400 & 0.0920 & \cellcolor[HTML]{FAC791}38.95 \\
Ours & 0.8252 & \cellcolor[HTML]{FAC791}0.1115 & \cellcolor[HTML]{F59194}43.11 & 0.7216 & \cellcolor[HTML]{FFFF8F}0.1307 & \cellcolor[HTML]{F59194}44.78 & \cellcolor[HTML]{FFFF8F}0.8070 & 0.0681 & \cellcolor[HTML]{F59194}42.03 \\ \midrule
\multicolumn{1}{c}{} & \multicolumn{3}{c}{Horse} & \multicolumn{3}{c}{Bird} & \multicolumn{3}{c}{Torus2sphere} \\
\multicolumn{1}{c}{\multirow{-2}{*}{Methods}} & CD↓ & EMD↓ & PSNR↑ & CD↓ & EMD↓ & PSNR↑ & CD↓ & EMD↓ & PSNR↑ \\ \midrule
D-NeRF & 1.6850 & 0.2800 & 25.47 & 1.5320 & 0.1630 & 23.85 & \cellcolor[HTML]{FFFF8F}1.7600 & 0.2500 & 24.23 \\
K-Plane & 1.4800 & 0.2390 & 28.11 & 0.7420 & 0.1310 & 23.72 & 1.7930 & \cellcolor[HTML]{FAC791}0.1610 & \cellcolor[HTML]{FAC791}31.21 \\
HexPlane & 1.7500 & 0.1990 & 26.80 & 4.1580 & 0.1780 & 22.19 & 2.1900 & 0.1900 & 29.71 \\
TiNeuVox-B & 1.9180 & 0.2460 & 28.16 & 8.2640 & 0.2150 & 25.55 & 2.1150 & 0.2030 & 28.76 \\
DG-Mesh & \cellcolor[HTML]{FFFF8F}0.2990 & \cellcolor[HTML]{FFFF8F}0.1680 & \cellcolor[HTML]{FFFF8F}30.64 & \cellcolor[HTML]{FAC791}0.5570 & \cellcolor[HTML]{FFFF8F}0.1280 & \cellcolor[HTML]{FFFF8F}27.91 & \cellcolor[HTML]{F59194}1.6070 & 0.1720 & 11.84 \\
DynaSurfGS & \cellcolor[HTML]{FAC791}0.2960 & \cellcolor[HTML]{F59194}0.1450 & 28.68 & 1.6310 & 0.1380 & 26.88 & \cellcolor[HTML]{FAC791}1.6750 & 0.1710 & 29.13 \\
Dynamic 2D Gaussians & 0.3910 & 0.1770 & \cellcolor[HTML]{FAC791}31.92 & \cellcolor[HTML]{F59194}0.3280 & \cellcolor[HTML]{FAC791}0.1100 & \cellcolor[HTML]{FAC791}28.03 & 2.4790 & \cellcolor[HTML]{FFFF8F}0.1640 & \cellcolor[HTML]{FFFF8F}30.17 \\
Ours & \cellcolor[HTML]{F59194}0.2510 & \cellcolor[HTML]{FAC791}0.1525 & \cellcolor[HTML]{F59194}39.09 & \cellcolor[HTML]{FFFF8F}0.7263 & \cellcolor[HTML]{F59194}0.0900 & \cellcolor[HTML]{F59194}34.79 & 3.3167 & \cellcolor[HTML]{F59194}0.1227 & \cellcolor[HTML]{F59194}34.34 \\ \bottomrule
\end{tabular}%
\end{table*}

\begin{table*}[]
\centering
\caption{\textbf{Quantitative Results on Lego}}
\label{tab:add_lego}
\resizebox{1.2\columnwidth}{!}{%
\begin{tabular}{@{}lccc@{}}
\toprule
Methods & PSNR↑ & MS-SSIM↑ & VGG-LPIPS↓ \\ \midrule
4D-GS & 28.72 & 0.9822 & 0.0368 \\
D-MiSo & 28.43 & 0.9810 & 0.0461 \\
SP-GS & 30.83 & 0.9864 & 0.0221 \\
Deformable-GS & 33.07 & 0.9794 & 0.0183 \\
SC-GS & \cellcolor[HTML]{FFFF8F}33.11 & \cellcolor[HTML]{FFFF8F}0.9886 & \cellcolor[HTML]{FFFF8F}0.0178 \\
Grid4D & \cellcolor[HTML]{FAC791}33.24 & \cellcolor[HTML]{FAC791}0.9938 & \cellcolor[HTML]{FAC791}0.0132 \\
Ours & \cellcolor[HTML]{F59194}34.56 & \cellcolor[HTML]{F59194}0.9945 & \cellcolor[HTML]{F59194}0.0114
 \\ \bottomrule
\end{tabular}%
}
\end{table*}

\section{Performance Benchmark}
\label{sec:performance}
Table \ref{tab:add_performance_benchmark} presents the performance evaluation of the FPS (frames per second) in relation to the number of 3D Gaussians used across various scenes in the D-NeRF dataset at a resolution of \(800 \times 800\). The results demonstrate variability in FPS depending on the scene and the number of Gaussians (denoted as "Num (k)"). On average, MaGS achieves an FPS of 70.18 with 151k Gaussians across all scenes.
\begin{table*}[]
\centering
\caption{\textbf{FPS Experiments with Respect to the Number of 3D Gaussians.}}
\label{tab:add_performance_benchmark}
\resizebox{\columnwidth}{!}{%
\begin{tabular}{@{}lcc@{}}
\toprule
\multicolumn{3}{l}{D-NeRF   Dataset (800x800)} \\ \midrule
Scene & FPS & Num (k) \\ \midrule
Jumpingjacks    & 59.66        & 175             \\
Bouncing Balls  & 57.30        & 194             \\
Hell Warrior    & 111.03       & 79              \\
Hook            & 93.04        & 100             \\
Standup         & 67.48        & 134             \\
Trex            & 61.48        & 168             \\
Lego            & 59.28        & 172             \\
Mutant          & 52.18        & 191             \\ \midrule
\textbf{Average} & \textbf{70.18} & \textbf{151}    \\ \bottomrule
\end{tabular}%
}
\end{table*}

\end{document}